\documentclass[preprint,12pt]{elsarticle}

\usepackage{amssymb}

\usepackage{amsmath}
\usepackage{lipsum}
\usepackage{amsfonts}
\usepackage{graphicx}
\usepackage{epstopdf}
\usepackage{algorithmic}
\usepackage[ruled]{algorithm2e}
\usepackage{mathtools}
\usepackage{url}
\usepackage{booktabs}       %
\usepackage{amsthm}
\ifpdf
  \DeclareGraphicsExtensions{.eps,.pdf,.png,.jpg}
\else
  \DeclareGraphicsExtensions{.eps}
\fi

\usepackage[colorlinks=true, citecolor=blue, filecolor=black, linkcolor=red, urlcolor=blue]{hyperref}

\newtheorem{theorem}{Theorem}
\theoremstyle{definition}

\theoremstyle{definition}
\newtheorem{remark}{Remark}

\usepackage{enumitem}
\setlist[enumerate]{leftmargin=.5in}
\setlist[itemize]{leftmargin=.5in}

\usepackage{amsopn}

\usepackage[textsize=tiny]{todonotes}

\usepackage{cleveref}
\usepackage{geometry}
\usepackage{graphicx, epstopdf}
\usepackage[caption=false]{subfig}
\usepackage{bm}
\usepackage{wrapfig}

\definecolor{darkred}{rgb}{.7,0,0}

\definecolor{darkgreen}{rgb}{.15,.55,0}

\definecolor{darkblue}{rgb}{0,0,0.7}

\newcommand{\design}{\xi}
\newcommand{\designset}{\mathcal{A}}
\newcommand{\pdf}{p}
\newcommand{\policy}{\pi}
\newcommand{\param}{\theta}
\newcommand{\Param}{\Theta}
\newcommand{\paramset}{\bm{\Theta}}
\newcommand{\nuis}{\psi}
\newcommand{\Nuis}{\Psi}
\newcommand{\nuisset}{\bm{\Psi}}
\newcommand{\info}{{\sf i}}
\newcommand{\Info}{I}

\renewcommand{\[}{\left[}
\renewcommand{\]}{\right]}
\renewcommand{\(}{\left(}
\renewcommand{\)}{\right)}

\newcommand{\abs}[1]{\left|#1\right|}

\newcommand{\norm}[2]{\left\|\, #1 \,\right\|_{#2}}

\newcommand{\vvvert}{|\kern-1pt|\kern-1pt|}

\newcommand{\EE}{\mathbb{E}}
\newcommand{\II}{\mathbb{I}}

\newcommand{\RR}{\mathbb{R}}

\newcommand{\CA}{\mathcal{A}}

\newcommand{\CF}{\mathcal{F}}

\newcommand{\CL}{\mathcal{L}}
\newcommand{\CM}{\mathcal{M}}
\newcommand{\CN}{\mathcal{N}}

\newcommand{\CS}{\mathcal{S}}

\newcommand{\CU}{\mathcal{U}}

\newcommand{\argmax}{\operatornamewithlimits{arg\,max}}
\newcommand{\argmin}{\operatornamewithlimits{arg\,min}}
\newcommand{\DKL}{D_{\mathrm{KL}}}

\def\argmin{\mathop{\mathrm{argmin}}}

\crefname{section}{Section}{Sections}
\Crefname{section}{Section}{Sections}
\crefname{subsection}{Section}{Sections}
\Crefname{subsection}{Section}{Sections}
\crefname{figure}{Fig.}{Fig.}
\Crefname{figure}{Figure}{Figures}
\crefname{equation}{Eq.}{Eqs.}
\Crefname{equation}{Equation}{Equations}
\crefname{table}{Table}{Tables}
\Crefname{table}{Table}{Tables}
\crefname{algorithm}{Algorithm}{Algorithms}
\Crefname{algorithm}{Algorithm}{Algorithms}
\crefname{theorem}{Theorem}{Theorems}
\Crefname{theorem}{Theorem}{Theorems}
\crefname{remark}{Remark}{Remarks}
\Crefname{remark}{Remark}{Remarks}
\crefname{appendix}{}{}
\Crefname{appendix}{}{}

\crefdefaultlabelformat{#2\textup{#1}#3}

\usepackage{pifont}
\newcommand{\cmark}{\ding{51}}%
\newcommand{\xmark}{\ding{55}}%

\usepackage{changes} %
\usepackage{soul} %
\usepackage{ulem} %

\begin{document}

\begin{frontmatter}

\title{Variational Sequential Optimal Experimental Design using Reinforcement Learning}

\author[inst1]{Wanggang Shen}

\affiliation[inst1]{organization={Department of Mechanical Engineering, University of Michigan},%
            city={Ann Arbor},
            state={Michigan},
            country={United States}}

\author[inst1]{Jiayuan Dong}
\author[inst1]{Xun Huan}

\begin{abstract}
We present variational sequential optimal experimental design (vsOED), a novel method for optimally designing a finite sequence of experiments within a Bayesian framework with information-theoretic criteria. 
vsOED employs a one-point reward formulation with variational posterior approximations,
providing a provable lower bound to the expected information gain. 
Numerical methods are developed following an actor-critic reinforcement learning approach, including derivation and estimation of variational and policy gradients to optimize the design policy,
and posterior approximation using Gaussian mixture models and normalizing flows. 
vsOED accommodates nuisance parameters, implicit likelihoods, and multiple candidate models, while supporting flexible design criteria that can target designs for model discrimination, parameter inference, goal-oriented  prediction, and their weighted combinations.
We demonstrate vsOED across various engineering and science applications,
illustrating its superior sample efficiency 
compared to existing sequential experimental design algorithms.
\end{abstract}


\end{frontmatter}

\section{Introduction}
\label{sec:introduction}

Engineering and science revolve around the interplay between data and models: leveraging data to develop, calibrate, and improve models, and using models to predict outcomes, control processes, and guide decision-making. When data is expensive to acquire, carefully designing experiments becomes crucial. 
\emph{Optimal experimental design (OED)} (see~\cite{Huan2024} for a recent review) is a field dedicated to systematically quantifying the value of experiments and identifying the best conditions to conduct them. 
{Bayesian OED}~\cite{Chaloner1995,Ryan2016,Alexanderian2021,Rainforth2023,Strutz2024} further incorporates the effects of uncertainty, and optimizes experiments so that their data can maximize uncertainty reduction, often measured as the \emph{expected information gain (EIG)} or mutual information~\cite{Lindley1956}.

Sequential experimental design involves planning multiple experiments conducted in sequence, where the results of earlier experiments can inform the
design of subsequent ones---i.e., it involves `feedback'. 
A straightforward approach is to design one experiment (or a subset) at a time: plan the next experiment, perform it, use the resulting data to update the model and its uncertainty, and repeat. This approach is known as \emph{greedy} or \emph{myopic} design since it focuses on the immediate next experiment without considering future ones; i.e., it lacks `lookahead'.  
Greedy design is advantageous for its simplicity and flexibility in accommodating an unknown total number of experiments, and it has been studied extensively~\cite{Box1992, 
Dror2008, Cavagnaro2010, Solonen2012, Drovandi2013, Drovandi2014, Kim2014,Hainy2016,Kleinegesse2021}.

In this paper, we follow a sequential experimental design formulation (see, e.g., \cite{Muller2007}, \cite[VII.G]{vonToussaint2011}, \cite[Chapter~3]{Huan2015})
that incorporates both (i) lookahead, where each design
is selected while accounting for all remaining
experiments, and (ii) feedback, where designs are given by adaptive policies that determine the next experiment based on the current state of information. 
This formulation, referred to as \emph{sequential OED (sOED)} in~\cite{Huan2015, Huan2016, Shen2023}, is general but computationally challenging. Previous approaches~\cite{Carlin1998,Gautier2000,Pronzato2002, Brockwell2003, Christen2003, Murphy2003,
Wathen2006,Muller2022,Tec2023} have largely been limited to discrete settings or have not employed a Bayesian framework with information-theoretic design criteria. Only recently have more efficient computational techniques---often leveraging reinforcement learning---been developed to tackle sOED in more general settings~\cite{Huan2015, Huan2016, shen2021bayesian, foster2021deep, ivanova2021implicit, blau2022optimizing}.

Huan and Marzouk~\cite{Huan2015, Huan2016} presented sOED as a finite-horizon Markov decision process and proposed an approximate dynamic programming algorithm to solve it. However, their approach relies on an implicit policy representation, where each policy evaluation involves solving a stochastic optimization problem, making it computationally expensive for online applications. 
To address this, Shen and Huan~\cite{shen2021bayesian,Shen2023} introduced policy-gradient sOED (PG-sOED), an actor-critic method that trains an explicit parameterized policy network, significantly reducing computational costs.
Around the same time, Foster \textit{et al.}~\cite{foster2021deep} proposed Deep Adaptive Design ({DAD}), which trains a policy network to maximize a lower bound estimator of the EIG, known as the sequential prior contrastive estimator (PCE), using its gradients. Unlike PG-sOED, DAD does not rely on an actor-critic framework but requires access to the derivatives of the observation model (i.e., parameter-to-observable map). 
Blau \textit{et al.}~\cite{blau2022optimizing} further proposed learning a stochastic policy to maximize the sequential PCE via randomized ensembled double Q-learning~\cite{Chen2021}, which sidestepped the need for model derivatives. 
Building upon DAD, Ivanova \textit{et al.}~\cite{ivanova2021implicit} developed implicit DAD ({iDAD}) to handle problems with implicit likelihood, where the likelihood is intractable but data samples can still be generated.
iDAD constructs variational lower bounds of the EIG and tightens these bounds by simultaneously optimizing the policy and variational parameters, requiring only sample-based updates. However, similar to DAD, iDAD also requires model derivatives.

EIG bounds~\cite{poole2019variational} have proven to be effective for OED, as demonstrated by iDAD and related works.
For example, the tractable unnormalized Barber--Agakov (TUBA) lower bound~\cite{poole2019variational} incorporates the tuning of a `critic' function (different from the `critic' in `actor-critic'). 
The Nguyen--Wainwright--Jordan (NWJ) bound \cite{nguyen2010}, also known as the mutual information neural estimation f-divergence (MINE-f) bound \cite{belghazi2018}, is a special case of TUBA and has been applied to OED by Kleinegesse and Gutmann~\cite{kleinegesse2020}. 
The information noise-contrastive estimation (InfoNCE) bound \cite{Oord_18_InfoNCE} mirrors the PCE, but replaces the likelihood with an exponentiated critic function.
Notably, all these bounds support implicit likelihoods.
The Barber--Agakov (BA) lower bound~\cite{barber2004algorithm}, in particular, approximates the posterior density directly. 
Its advantage lies in enabling both density evaluation and posterior sampling.
Foster \textit{et al.} \cite{Foster2019} first applied the BA bound to OED with simple variational distributions (e.g., Gaussian, Bernoulli, Gamma). 
Dong \textit{et al.}~\cite{Dong_25} later proposed variational OED (vOED) to expand the capacity of the approximate posteriors by representing them with normalizing flows~\cite{Papamakarios_21_NFsReview, Kobyzev_20_NFsReview}. 

The above OED efforts primarily focused on maximizing the EIG of unknown model parameters. However, the ultimate goals of the experiments often extend beyond parameter inference to predicting other \emph{goal} quantities that depend on the model parameters but differ from the observation quantities. For example, while observations such as temperature and pressure facilitate the inference of a weather model's parameters, the goal might not be to learn these parameters, but to forecast future precipitation using the resulting parameter distributions. 
A \emph{goal-oriented} OED framework therefore directly targets the EIG for such goal quantities. 
This concept builds on classical I-, V-, and G-optimal designs that seek to minimize predictive variance in linear regression models~\cite{Atkinson2007}. 
Attia \textit{et al.}~\cite{Attia2018} 
demonstrated gradient-based optimization techniques to solve linear Bayesian D$_A$- and L-optimal design problems, while Wu \textit{et al.}~\cite{wu2023offline} developed scalable offline-online decompositions and low-rank approximations for high-dimensional linear settings. 
The theoretical formulation for \emph{nonlinear} goal-oriented OED was presented in~\cite{Bernardo1979}. Computationally, Butler \textit{et al.}~\cite{Butler2020} proposed OED for prediction (OED4P), a non-Bayesian approach based on matching push-forward distributions to observed data distributions~\cite{Butler2018,Butler2018a}. Later, Smith \textit{et al.}~\cite{smith2023prediction} introduced expected predictive information gain (EPIG), a nested Monte Carlo estimator for predictive EIG
under a greedy sequential (i.e., active learning) setting, but 
requiring the goal quantities to be the same as observation quantities except at new designs.
More recently, Zhong \emph{et al.}~\cite{Zhong2024} combined Markov chain Monte Carlo with kernel density estimation to estimate the probability densities of goal quantities directly. 
These studies highlight the challenge of intractable predictive densities for the goal quantities, motivating the potential effectiveness of likelihood-free EIG bounds. 
Kleinegesse and Gutmann~\cite{kleinegesse2021gradient} explored such bounds for various goal-oriented OED settings. 
However, existing goal-oriented efforts all focused on batch (i.e., non-sequential) or greedy OED, and integrating goal-oriented objectives into sOED has yet to be developed.

Building on these advances, we propose \emph{variational sequential optimal experimental design (vsOED)}, a method to solve the sOED problem under an actor-critic reinforcement learning framework powered by
policy gradient and variational BA bound to the EIG.
The key novelty and contributions of this work are summarized as follows.
\begin{itemize}
\item Comprehensive formulation: vsOED accommodates nuisance parameters, implicit likelihoods, and multiple candidate models, while offering flexible design criterion that can target designs for model discrimination, parameter inference, 
goal-oriented  prediction,
or weighted combinations thereof.

\item Theoretical guarantees: We prove the lower bound property of vsOED, and the equivalence of various reward formulations for vsOED.

\item Numerical methods: We develop techniques for solving vsOED, highlighted by estimating the actor-critic policy gradient and approximating posteriors using Gaussian mixture models and normalizing flows.

\item Empirical demonstration: We illustrate the effectiveness of vsOED, particularly under limited sample budgets, through comparisons with existing algorithms across diverse numerical experiments.
\end{itemize}
The vsOED code is available at \url{https://github.com/wgshen/vsOED}.

The paper is structured as follows. \Cref{sec:formulation} formulates the sOED and vsOED problem statements, including key theorems on their relationships. 
\Cref{sec:method} details the numerical methods for solving vsOED, centering around deriving and computing the policy gradient under an actor-critic framework. 
\Cref{sec:results} evaluates vsOED through numerical experiments and benchmarks against existing algorithms. Finally, \cref{sec:conclusions} concludes with summarizing remarks and a discussion of future work.

\section{Problem formulation}
\label{sec:formulation}

\subsection{Preliminaries}
\label{sec:background}

We focus on the design of a finite sequence of $N$ experiments indexed by $k \in \{0,\ldots,N-1\}$. We assume $N$ is known and fixed. Each experiment is associated with a real-valued design vector $\design_k \in \designset_k \subseteq \RR^{N_{\design}}$ 
and observation vector $Y_k \in \mathcal{Y}_k \subseteq \RR^{N_y}$. 
We adopt the convention to use upper case for denoting a random variable or vector such as $Y_k$, and lower case for its realized value such as $y_k$.
For simplicity, we assume $N_{\design}$ and $N_y$ to be constant across $k$. 
At the $k$th experiment, the information sequence of all completed experiments' designs and observations is denoted by $\Info_k=\{ \design_0,Y_0,\dots,\design_{k-1},Y_{k-1} \}$, with $\Info_0=\emptyset$.
Furthermore, let $\CM_m$ be a countable set of models indexed by $m \in \{1,2,\dots\}$ where each model has its own parameters of interest (PoIs) $\Param_m \in \paramset_m \subseteq \RR^{N_{\param_m}}$, 
nuisance parameters $\Nuis_m \in \nuisset_m \subseteq\RR^{N_{\nuis_m}}$, 
and predictive quantities of interest (QoIs) $Z_m \in \mathcal{Z}_m \subseteq \RR^{N_{z_m}}$. 
The experiment observation is related to the parameters via an observation model, for example often in the form
\begin{align}
    Y_k = G_k(\Param_m, \Nuis_m, \design_k; m, \Info_k) + \mathcal{E}_k,
    \label{eqn:observation}
\end{align}
where $G_k$ is a nonlinear observation forward operator and $\mathcal{E}_k$ is the observation noise. The QoIs are related to the parameters via a predictive model 
\begin{align}
    Z_m = H(\Param_m, \Nuis_m, \mathcal{E}_{Z}; m),
    \label{eqn:predictive}
\end{align}
where $H$ is a nonlinear predictive stochastic forward operator with $\mathcal{E}_{Z}$ representing the predictive stochastic source. If the predictive model is deterministic, we can write $Z_m = H(\Param_m, \Nuis_m; m)$. 
$G_k$ and $H$ often involve the most expensive computations (e.g., solving various differential equations), hence the number of these forward solves is often used as the unit for computational cost.

Under a Bayesian approach, after the $k$th experiment, the joint probability density on $M$, $\Param_m$, and $\Nuis_m$
updates following Bayes' rule:
\begin{align}
    \label{eq:bayes_rule}
    p(m,\param_m,\nuis_m|\info_{k+1})=p(m,\param_m,\nuis_m|\design_k,y_k,\info_k) = \frac{p(y_k|m,\param_m,\nuis_m, \design_k,\info_k) \, p(m,\param_m,\nuis_m|\info_k)}{p(y_k|\design_k,\info_k)}, 
\end{align}
where $p(m,\param_m,\nuis_m|\design_k,y_k,\info_k)$ is the posterior which can also be written as $p(m,\param_m,\nuis_m|\info_{k+1})$, 
$p(m,\param_m,\nuis_m|\info_k)$ is the prior, $p(y_k|m,\param_m,\nuis_m,\design_k,\info_k)$ is the likelihood, and
$p(y_k|\design_k,\info_k)$ is the marginal likelihood. 
In order to simplify notation in the remainder of this paper, we adopt the
convention
where when $m$ is not explicitly mentioned, conditioning on $m$ is implied through other variables' subscripts,
e.g., $p(\param_m,\nuis_m|\info_{k})=p(\param_m,\nuis_m|m,\info_{k})$.
The posterior after the $k$th experiment $p(m,\param_m,\nuis_m|\info_{k+1})$ then serves as the prior for the $(k+1)$th experiment and is again applied to \cref{eq:bayes_rule}.
Bayes' rule can be recursively applied for sequential experiments.

Upon propagating the parameter distributions through 
$H$, the prior- and posterior-predictive densities for $Z_m$ can be respectively written as:
\begin{align}
    p(z_m) &= %
    \iint p(\param_m,\nuis_m) \, p(z_m|\param_m,\nuis_m) \,\mathrm{d}\param_m\,\mathrm{d}\nuis_m, \label{eq:prior_predictive}\\
    p(z_m|\info_{k+1}) &= %
    \iint p(\param_m,\nuis_m|\info_{k+1}) \, p(z_m|\param_m,\nuis_m) \,\mathrm{d}\param_m\,\mathrm{d}\nuis_m.
    \label{eq:post_predictive}
\end{align}
If $H$ is deterministic, $p(z_m|\param_m,\nuis_m)$ collapses to a Dirac delta centered at $z_m=H(\param_m,\nuis_m;m)$.

\subsection{Sequential optimal experimental design (sOED)}
\label{sec:sOED}

We present sOED following~\cite{Shen2023} and earlier works~\cite{Huan2015,Huan2016} that modeled it through a Markov decision process (MDP). The MDP is defined by a tuple 
$\bigl ( \CS, \{ \CA_k \}_k, s_0, \{ r_k(\cdot) \}_k, \{T_k(\cdot) \}_k \bigr )$
consisting of a state space $\CS$ for the state variable $S_k$
that can take values $s_k \in \CS$,
action spaces $\CA_k$ comprising possible actions (which here are designs) $\design_k\in \CA_k$,
an initial state $s_0$, scalar-valued reward functions $r_k(s_k,\design_k,y_k)$
that evaluate the instantaneous reward when taking action $\design_k$ and observing $y_k$ at
state $s_k$, and state transition kernels
$T_k(\mathcal{S}_{k+1} \vert
  s_k, \design_k)$
that evaluate the probability of transitioning to any set of states
$\mathcal{S}_{k+1} \subseteq \mathcal{S}$ at stage $k+1$ having taken action $\design_k$ at state $s_k$.
In the context of experimental design, the action being taken is the selection of a
design; thus we use the terms `action' and `design'
interchangeably.

\paragraph{State} The state before the $k$th experiment is described by $S_k=\{S_{k}^{b},S_{k}^{p}\}$, a quantity that summarizes all information deemed relevant to future
design decisions.
We split {$S_k$} into a `belief state' {$S_{k}^{b}$}
representing the state of knowledge/uncertainty in $M$, $\Param_m$, $\Nuis_m$, and $Z_m$, and a
`physical state' {$S_{k}^{p}$} comprising any other deterministic
variables that may be relevant to the design process. 
A realization of the belief state, $s_{k}^{b}$, is simply the joint posterior distribution of $M$, $\Param_m$, $\Nuis_m$, and $Z_m$
given all past experimental designs and realized observations,
$\info_k$. 
Numerically, it can be represented by, for example, a
  density or distribution function approximation, or an ensemble of particles, or
  by tracking $\info_k$ directly.
Tracking $\info_k$ is easiest to implement since it does not require
additional calculations translating  $(\design_i)_{i < k}$ and
  $(y_i)_{i < k}$ to another representation, but the dimension of $\info_k$
grows with $k$ albiet 
bounded for finite $N$.

Maintaining only a belief state, in the form of the posterior, does
  not suffice to preserve the Markov property of the system if the likelihood 
  depends on the history of past experiments $\info_k$, as 
  in \cref{eq:bayes_rule}. This can be fixed by introducing a
  physical state.
With regard to the $\info_k$-dependence in $\pdf(y_k|m,\param,\nuis_k,\design_k,\info_k)$, the
physical state essentially extracts and tracks relevant features
from $\info_k$ that allow the likelihood to be evaluated or the
observations $Y_k$ simulated.  
If $\Info_k$ is adopted as the belief state, then information about the physical state is already
contained in $\Info_k$, even if only implicitly.
We adopt $S_k=\Info_k$ in this work and will use them interchangeably.

\paragraph{Action (design) and policy} 
Sequential experimental design is adaptive in nature, and looks for a strategy,
called a \emph{policy} (or \emph{actor}), describing \emph{how to choose} the design
depending on the current state.
The policy is a collection of functions
$\policy = \{\mu_k : \CS \to \CA_k, k=0,\ldots,N-1\}$, where the policy
function $\mu_k$ returns the design for the $k$th experiment given the
current state, $\design_k=\mu_k(s_k)$.

\paragraph{State transition dynamics} 
When an experiment is performed,
the state changes according to a transition kernel
$T_k(\mathcal{S}_{k+1} \vert
  s_k, \design_k)$
describing the probability of
transitioning from the current state $s_k$, having chosen design
$\design_k$ and observed the outcome of resulting experiment, to
any set of states
  at stage $k+1$, $\mathcal{S}_{k+1} \subseteq \mathcal{S}$.
This kernel is generally intractable to evaluate, but can
  be simulated by first sampling $Y_k$ given the design $\design_k$, and then applying
  Bayes' rule \cref{eq:bayes_rule}. We denote the latter transition dynamics by
$s_{k+1}=\CF_k(s_k,\design_k,y_k)$; the function $\CF_k$ encapsulates
the transition from prior to posterior, given values of $\design_k$
and the realized data $y_k$, following \cref{eq:bayes_rule}.
If adopting $\Info_k$ as the state, then the 
transition is a simple concatenation
$\info_{k+1} = \{\info_k, \design_k, y_k\}$.

\paragraph{Reward (utility)} Here $r_k(s_k,\design_k,y_k) \in \RR$ denotes the reward immediately obtained from the $k$th experiment, and $r_N(s_N) \in \RR$ is the terminal reward that can be computed only after all experiments are completed. Examples of information-theoretic rewards will be provided in \cref{sec:utility,sec:one_point_utility}.

\paragraph{sOED problem statement} The sOED problem seeks a design policy that maximizes the expected utility $U(\policy)$:
\begin{align}
    \label{eq:optimal_policy}
    \policy^\ast \in \argmax_{\policy=\{\mu_0,\ldots,\mu_{N-1}\}}& \qquad \left\{ U(\policy) := \EE_{Y_{0:N-1}|\policy,s_0}\[\sum_{k=0}^{N-1}r_k(S_k,\design_k,Y_k)+r_N(S_N)\]\right\}\\
    \text{subject to} \hspace{0.4em}& 
    \qquad \design_k = \mu_k(S_k) \in \designset_k, \nonumber\\
    &\qquad S_{k+1}=\CF_k(S_k,\design_k,Y_k),
    \quad \text{for}\,\, k=0,\dots,N-1.\nonumber
\end{align}
Here $Y_{0:N-1} = Y_0,Y_1,\ldots,Y_{N-1}$. The initial state $s_0$ is assumed known; if it is not known, another expectation can be taken over $S_0$. 
While we strive to explicitly write out the conditioning on $s_0$, it should be interpreted that all terms in this paper are implicitly conditioning on $s_0$ even if not written. 
When adopting $S_k=\Info_k$, the constraints can be written as 
$\design_k = \mu_k(\Info_k) \in \designset_k$ and $\Info_{k+1}=\{\Info_k, \design_k, Y_k\}$.
This sOED formulation generalizes both the batch and greedy designs~\cite[Section 2.3]{Shen2023}.

\subsection{Experimental design rewards}
\label{sec:utility}

We propose two information gain (IG)-based~\cite{Lindley1956,Ginebra2007} reward formulations incorporating multiple design objectives. 
Non-IG-based reward terms (i.e., those that do not depend on the posteriors, such as a cost term of the experiment that depends only on $\design_k$) may also be added without affecting the results below, but we omit them for simplicity of presentation.
We will largely use $\Info_k$ in place of $S_k$ from hereon.

\paragraph{(1) {Terminal-information-gain (TIG)}} TIG targets the overall IG (i.e., the Kullback--Leibler [KL] divergence from prior to posterior) from all $N$ experiments via the terminal reward:
\begin{align}
 &r_k(\info_k, \design_k, y_k) = 0, \quad k=0,\ldots,N-1, \label{eq:terminal1}\\
 &r_N(\info_N) 
 = \alpha_{M} \DKL\(\, P_{M|\info_N}\,||\,P_M \,\) \nonumber \\
 &\hspace{4em} + \EE_{M|\info_N} \[ \alpha_{\Param} \DKL\(\, p_{\Param_m|\info_N}\,||\,p_{\Param_m} \,\) + \alpha_Z \DKL\(\, p_{Z_m|\info_N}\,||\,p_{Z_m} \,\) \],
 \label{eq:terminal_info_gN}
\end{align}
where in the KL divergence the upper case $P$ denotes probability mass (for discrete random variables), subscript indicates the random variable that corresponds to the probability mass or density, and
$\alpha_{M}, \alpha_{\Param}, \alpha_Z \in [0,1]$ are the weights for IG respectively coming from the model indicator, PoIs, and QoIs. For example, setting $\alpha_{M}=1$ and $\alpha_{\Param}=\alpha_Z=0$ reduces to 
`OED for model indicator' (i.e., design for model discrimination);
$\alpha_{\Param}=1$ and $\alpha_{M}=\alpha_Z=0$ reduces to 
`OED for PoIs'
(i.e., design for parameter inference); 
$\alpha_Z=1$ and $\alpha_{\Param}=\alpha_{M}=0$ reduces to 
`OED for QoIs'
(i.e., design for goal-oriented prediction).  
In the special case when $\alpha_{M}=\alpha_{\Param}=1$ and $\alpha_Z=0$, 
the terminal reward $r_N$ becomes $\DKL\(\, p_{M,\Param_m|\info_N}\,||\,p_{M,\Param_m} \,\)$ (\cref{app:total_info_gain}); when $\alpha_{M}=\alpha_Z=1$ and $\alpha_{\Param}=0$, it similarly becomes $\DKL\(\, p_{M,Z_m|\info_N}\,||\,p_{M,Z_m} \,\)$ (\cref{app:total_info_gain}). 
When nuisance parameters $\Nuis_m$ are absent, one should not set both $\alpha_{\Param}$ and $\alpha_Z$ to 1 since the IG in $Z_m$ is already absorbed into the IG in $\Param_m$ (\cref{app:joint_info_gain}). 

\paragraph{(2) {Incremental-information-gain (IIG)}} IIG adopts incremental IG as the immediate rewards:
\begin{align}
    &r_k(\info_k, \design_k, y_k) = \alpha_{M} \DKL\(\, P_{M|\info_{k+1}}\,||\,P_{M|\info_k} \,\) 
 \nonumber\\
 &\hspace{6.2em}+ \EE_{M|\info_{k+1}} \Big[ \alpha_{\Param} \DKL\(\, p_{\Param_m|\info_{k+1}}\,||\,p_{\Param_m|\info_k} \,\) 
 \nonumber\\
 &\hspace{11em}
 + \alpha_Z \DKL\(\, p_{Z_m|\info_{k+1}}\,||\,p_{Z_m|\info_k} \,\) \Big], \quad k=0,\ldots,N-1,\label{eq:incremental1}\\
    &r_N(\info_N) = 0. \label{eq:incremental2}
\end{align}
Note that $\info_{k+1}$ can be evaluated in \cref{eq:incremental1} since $\info_{k+1}=\{\info_k, \design_k, y_k\}$.

Let $U_T(\policy)$ denote the resulting sOED expected utility in \cref{eq:optimal_policy} when adopting the TIG rewards in \cref{eq:terminal1,eq:terminal_info_gN}, and $U_I(\policy)$ denote that when adopting the IIG rewards in \cref{eq:incremental1,eq:incremental2}.

\begin{theorem}[Terminal-incremental equivalence]
\label{prop:terminal_incremental}
$U_T(\policy)=U_I(\policy)$ for any policy $\policy$. 
\end{theorem}
A proof is provided in \cref{app:terminal_incre}.
Hence, both formulations induce the same sOED problem.

\subsection{One-point formulation of rewards}
\label{sec:one_point_utility}

To compute the sOED objective in \cref{eq:optimal_policy} with the TIG or IIG
rewards,
an expectation needs to be taken over $Y_{0:N-1} \vert \policy,
s_0$. This requires sampling trajectories. For each trajectory, a
model indicator and corresponding PoIs and nuisance parameters are drawn from  
the priors, $m_0^{(i)} \sim P_M, \param_{m,0}^{(i)} \sim
p\big(\param_{m,0} \vert m_0^{(i)}\big), \nuis_{m,0}^{(i)} \sim
p\big(\nuis_{m,0} \vert m_0^{(i)}\big)$,
and a corresponding QoI sample is also drawn, $z_{m,0}^{(i)}
\sim p\big(z_{m,0}|\param_{m,0}^{(i)},\nuis_{m,0}^{(i)},m_0^{(i)}\big)$
(if $Z_m$ has a deterministic $H$, then simply $z_{m,0}^{(i)}
= H\big(\param_{m,0}^{(i)},\nuis_{m,0}^{(i)};m_0^{(i)}\big)$). 
Then, $m_0^{(i)}$, $\param_{m,0}^{(i)}$, and $\nuis_{m,0}^{(i)}$ generate a
trajectory $\info_N$. For any such trajectory, we substitute these `oracle' values of the model indicator and model parameters that generated $\info_N$ into the integrands of the KL divergence terms to produce a `one-point' approximation $\check{r}_k$ for each $r_k$.

\paragraph{(1) {One-point-TIG}} One-point-TIG entails the following new reward terms:
\begin{align}
 &\check{r}_k(\info_k, \design_k, y_k) = 0, \quad k=0,\ldots,N-1, \label{eq:one_point_terminal1}\\
 &\check{r}_N(\info_N) 
 = \alpha_{M} \log \frac{P(m_0|\info_N)} {P(m_0)}  + \alpha_{\Param} \log \frac{p(\theta_{m,0}|\info_N)}{p(\theta_{m,0})} + \alpha_Z \log \frac{p(z_{m,0}|\info_N)}{p(z_{m,0})}.
 \label{eq:one_point_terminal_info_gN}
\end{align}

\paragraph{(2) {One-point-IIG}} One-point-IIG entails the following new reward terms:
\begin{align}
   & \check{r}_k(\info_k, \design_k, y_k) = \alpha_{M} \log \frac{P(m_{0}|\info_{k+1})}{P(m_{0}|\info_k)} + \alpha_{\Param} \log  \frac{p(\param_{m,0}|\info_{k+1})}{p(\param_{m,0}|\info_k) } 
   \nonumber\\
&\hspace{6.2em}
 + \alpha_Z \log \frac{p(z_{m,0}|\info_{k+1})}{p(z_{m,0}|\info_k)}, 
 \quad k=0,\ldots,N-1,
 \label{eq:one_point_incremental1}
 \\ & \check{r}_N(\info_N) = 0. \label{eq:one_point_incremental2}
\end{align}
More precisely, the one-point reward terms are now functions of the trajectory-generating oracle model and parameter values, i.e., $\check{r}_k(\info_k, \design_k, y_k, m_0, \param_{m,0}, z_{m,0})$ and $\check{r}_N(\info_N, m_0, \param_{m,0}, z_{m,0})$, but with the constraint that the $m_0, \param_{m,0}, z_{m,0}$ arguments are those that generated the $\info_N$ argument; we omitted the oracle arguments in the $\check{r}_k$ expressions above for simplicity. 
Thus, we need to take expectation jointly over these inputs, to arrive at
the \emph{one-point expected utility}:
\begin{align}
    \label{eq:one_point_expected_utility}
    \check{U}(\policy) = \EE_{M_{0},\Param_{m,0},\Nuis_{m,0},Z_{m,0}}
    \[\EE_{\Info_N|\policy,s_0,M_{0},\Param_{m,0},\Nuis_{m,0}}\[\sum_{k=0}^{N-1}\check{r}_k(\Info_k,\design_k,Y_k)+\check{r}_N(\Info_N)\]\],
\end{align}
where we switch to the notation with $S_k=\Info_k$ to correspond to the $\Info_k$-based reward definitions, use the equivalence of $\EE_{Y_{0:N-1}|\policy,s_0,\ldots}$ and $\EE_{\Info_N|\policy,s_0,\ldots}$,
 and follow the policy $\design_k=\mu_{k}(\Info_k)$.
Moreover, the inner expectation's conditioning on $Z_{m,0}$ is dropped since the $\Info_N$ sequence (and the $Y_k$'s within $\Info_N$) does not depend on $Z_m$ when $\policy$ is given. Furthermore, the expectation over $Z_{m,0}$ can be ignored altogether for $Z_m$'s with deterministic $H$. 

Let $\check{U}_T(\policy)$ denote the resulting sOED expected utility from \cref{eq:one_point_expected_utility} when adopting the one-point-TIG rewards in \cref{eq:one_point_terminal1,eq:one_point_terminal_info_gN}, and $\check{U}_I(\policy)$ denote that when adopting the one-point-IIG rewards in \cref{eq:one_point_incremental1,eq:one_point_incremental2}.

\begin{theorem}[One-point equivalence]
\label{prop:one_point_expected_utility}
$U_T(\policy)=\check{U}_T(\policy)=\check{U}_I(\policy)=U_I(\policy)$ for any policy $\policy$.
\end{theorem}
A proof is provided in \cref{app:one_point_expected_utility}. Hence, both the original sOED and one-point formulations, using either TIG or IIG, induce the same sOED problem. 

\begin{remark}
\label{rm:prior_cancel}
For $\check{U}_I(\policy)$, under the summation, all intermediate posteriors cancel out and only the prior terms $p(\cdot)$ and the final posterior terms $p(\cdot|\Info_N)$ survive. However, working with intermediate posteriors in the incremental rewards can lead to denser rewards that improves numerical performance \cite{blau2022optimizing}. 
\end{remark}

\begin{remark}
For any expected utility form,
the prior terms $p(\cdot)$  may be omitted since they are independent of $\policy$ and would only shift the expected utility without affecting the maximizer (\cref{app:omit_prior}). Therefore, when the prior is difficult to compute (e.g., the prior-predictive $p(z_m)$ in \cref{eq:prior_predictive}
that needs to marginalize out $\Param_m$ and $\Nuis_m$),
we will drop that prior term when optimizing the policy.
\end{remark}

\subsection{Variational sequential optimal experimental design (vsOED)}
\label{sec:vsOED}

Evaluating the probability terms of the one-point formulations in \cref{sec:one_point_utility} remains highly challenging. 
To make the computation tractable, we replace all posterior terms $p(\cdot|\info_k)$ with variational posterior approximations $q(\cdot|\info_k;\phi_{(\cdot)})$ parameterized by $\phi_{(\cdot)}$.

\paragraph{(1) {Variational-one-point-TIG}} Variational-one-point-TIG reward terms update to:
\begin{align}
     &\tilde{r}_k(\info_k, \design_k, y_k; \phi) = 0, \quad k=0,\ldots,N-1, 
     \label{eq:variational_one_point_terminal1}\\
    &\tilde{r}_N(\info_N;\phi) = \alpha_{M} \log \frac{q(m_{0}|\info_N;\phi_M)}{P(m_{0})}  + \alpha_{\Param} \log \frac{q(\param_{m,0}|\info_N;\phi_{\Theta_m})}{p(\param_{m,0})} 
    + \alpha_Z \log \frac{q(z_{m,0}|\info_N;\phi_{Z_m})}{p(z_{m,0})},
    \label{eq:variational_one_point_terminal_info_gN}
\end{align}
where $\phi=\{\phi_M, \phi_{\Theta_m}, \phi_{Z_m}\}$ is the full set of variational parameters.
\paragraph{(2) {Variational-one-point-IIG}} Variational-one-point-IIG reward terms update to:
\begin{align}
    &\tilde{r}_k(\info_k, \design_k, y_k;\phi) = \alpha_{M} \log \frac{q(m_{0}|\info_{k+1};\phi_{M})}{q(m_{0}|\info_k;\phi_{M})} + \alpha_{\Param} \log  \frac{q(\param_{m,0}|\info_{k+1};\phi_{\Theta_m})}{q(\param_{m,0}|\info_k;\phi_{\Theta_m})} 
    \nonumber\\
    &\hspace{7.2em} + \alpha_Z \log \frac{q(z_{m,0}|\info_{k+1};\phi_{Z_m})}{q(z_{m,0}|\info_k;\phi_{Z_m})}, \quad  k=0,\ldots,N-1,
    \label{eq:variational_one_point_incremental1} 
    \\
    &\tilde{r}_N(\info_N;\phi) = 0,
    \label{eq:variational_one_point_incremental2}
\end{align}
with the understanding that here $q(\cdot|\info_0;\phi_{(\cdot)})$ is simply the prior $p(\cdot)$ without any approximation. 
Similar to $\check{r}_k$, the new $\tilde{r}_k$ expressions are also functions of the oracle variables, but we omitted them in the expressions above for simplicity.
Upon taking expectation over all random variables, the corresponding \emph{variational one-point expected utility} becomes:
\begin{align}
\label{eq:variational_one_point_expected_utility}
    \tilde{U}(\policy,\phi) = \EE_{M_{0},\Param_{m,0},\Nuis_{m,0},Z_{m,0}}
    \[\EE_{\Info_N|\policy,s_0,M_{0},\Param_{m,0},\Nuis_{m,0}}\[\sum_{k=0}^{N-1}\tilde{r}_k(\Info_k,\design_k,Y_k;\phi)+\tilde{r}_N(\Info_N;\phi)\]\],
\end{align}
where $\design_k=\mu_{k}(\Info_k)$, and again the outer expectation over $Z_{m,0}$ can be ignored for $Z_m$'s with deterministic $H$. 

Let $\tilde{U}_T(\policy)$ denote the resulting sOED expected utility from \cref{eq:variational_one_point_expected_utility} when adopting the variational-one-point-TIG rewards in \cref{eq:variational_one_point_terminal1,eq:variational_one_point_terminal_info_gN}, and $\tilde{U}_I(\policy)$ denote that when adopting the variational-one-point-IIG rewards in \cref{eq:variational_one_point_incremental1,eq:variational_one_point_incremental2}.

\begin{theorem}[Variational lower bound]
\label{prop:variational_one_point_expected_utility}
$\tilde{U}_I(\policy;\phi) = \tilde{U}_T(\policy;\phi) \leq \check{U}_T(\policy) = \check{U}_I(\policy) = U_T(\policy) = U_I(\policy)$
for any policy $\policy$ and variational parameters $\phi$. The bound is tight if and only if all final posteriors are perfectly approximated, i.e., $q(\cdot|\info_N;\phi_{(\cdot)}) = p(\cdot|\info_N)$ (except the trivial case when $\alpha_{M}= \alpha_{\Param}= \alpha_Z = 0$).
\end{theorem}

A proof is provided in \cref{app:variational_expected_utility_lower_bound}. 

\begin{remark}
In batch OED and when $\alpha_M = \alpha_Z = 0$ and $\alpha_{\Param}=1$, \cref{eq:variational_one_point_expected_utility} becomes the BA lower bound for mutual information between $\Param$ and $Y$~\cite{barber2004algorithm, poole2019variational,Foster2019}. 
\end{remark}

\begin{remark}
The results in all aforementioned theorems also hold when incorporating non-IG-based reward terms (i.e., those that do not depend on the posteriors, such as a cost term of the experiment that depends only on $\design_k$).
\end{remark}

\begin{remark}
The tightness of the bound does not depend on the quality of the intermediate variational posteriors (i.e., $q(\cdot|\info_k;\phi_{(\cdot)})$ for $k=1,\dots,N-1$) due to their cancellations, similar to \cref{rm:prior_cancel}. 
Thus, low-quality intermediate posterior approximations may be used without affecting the theoretical value of the bound (see the first part of \cref{app:variational_expected_utility_lower_bound}). 
In the special case where all intermediate posteriors are approximated by the prior, {variational-one-point-IIG} collapses to {variational-one-point-TIG}. As we will show through numerical examples, however, good intermediate posterior approximations can lead to better computational performance.
\end{remark}

\paragraph{Variational sOED problem statement} The \textbf{vsOED} problem seeks a design policy that maximizes the lower bound $\tilde{U}(\policy;\phi)$:
\begin{align}
\label{eq:vsOED}
\{\policy^\ast,\phi^\ast\} \in \argmax_{\policy=\{\mu_0,\ldots,\mu_{N-1}\},\phi}  & \qquad \tilde{U}(\policy;\phi)\\
\text{subject to} \hspace{0.7em}& 
    \qquad \design_k = \mu_k(\Info_k)
    \in \designset_k, \nonumber\\
    &\qquad \Info_{k+1}=\{\Info_k, \design_k, Y_k\},
    \quad \text{for}\,\, k=0,\dots,N-1,\nonumber
\end{align}
where we used the notation adopting $S_k=\Info_k$.

\section{Numerical methods for vsOED}
\label{sec:method}

Taking a similar approach as~\cite{Shen2023}, we propose an actor-critic policy gradient method to solve the vsOED problem in \cref{eq:vsOED}. 
The key to this method is to derive and compute gradients of the expected utility lower bound $\tilde{U}$ defined in \cref{eq:variational_one_point_expected_utility} with respect to the variational parameters and the policy, and use the gradients to improve the policy via, for example, gradient ascent.

To bring the policy (which entails functions) to a finite-dimensional space, we parameterize the policy $\policy$ by $w \in \RR^{N_w}$ and denote the parameterized policy as $\policy_w$. Forming the policy (i.e., the \emph{actor}) explicitly in such a manner
offers significantly faster online evaluation speeds~\cite{shen2021bayesian, foster2021deep, ivanova2021implicit, blau2022optimizing} compared to dynamic programming \cite{Huan2016} and greedy design that require solving optimization problems on the fly.
Upon replacing $\policy$ with $\policy_w$,
the vsOED problem from \cref{eq:vsOED} becomes:
\begin{align}
    \label{eq:variational_optimal_policy}
\{w^\ast,\phi^\ast\} \in \hspace{0.6em} \argmax_{w,\phi} \hspace{0.6em} & \qquad \tilde{U}(w;\phi)\\
\text{subject to}& 
    \qquad \design_k = \mu_{k,w}(\Info_k)
    \in \designset_k, \nonumber\\
    &\qquad \Info_{k+1}=\{\Info_k, \design_k, Y_k\},
    \quad \text{for}\,\, k=0,\dots,N-1.\nonumber
\end{align}
The gradient of $\tilde{U}$ with respect to $\phi$ can be obtained trivially by applying the Leibniz rule:
\begin{align}
    \nabla_\phi \tilde{U}(w;\phi) = \EE_{M_{0},\Param_{m,0},\Nuis_{m,0},Z_{m,0}}\[\EE_{\Info_N|w,s_0,M_{0},\Param_{m,0},\Nuis_{m,0}}\[\sum_{k=0}^{N-1}\nabla_\phi \tilde{r}_k(\Info_k,\design_k,Y_k;\phi) + \nabla_\phi \tilde{r}_N(\Info_N;\phi)\]\],
    \label{eq:variational_gradient}
\end{align}
where $\design_k=\mu_{k,w}(\Info_k)$.
The actor-critic policy gradient can be derived near-identically as the proof in Appendix B of~\cite{Shen2023} except that the expressions for vsOED now involve an additional outer expectation jointly over $M_{0},\Param_{m,0},\Nuis_{m,0},Z_{m,0}$; therefore we do not repeat the derivation in this paper. The final vsOED policy gradient expression is:
\begin{align}
    \nabla_w \tilde{U}(w;\phi) = \EE_{M_{0},\Param_{m,0},\Nuis_{m,0},Z_{m,0}} \[\sum_{k=0}^{N-1} \EE_{\Info_k|w,s_0,M_{0},\Param_{m,0},\Nuis_{m,0}}\Big[ \nabla_w \mu_{k,w}(\Info_k) \nabla_{\design_k} \tilde{Q}_k^{\policy_w}\(\Info_k,\design_k;\phi\) \Big]\],
    \label{eq:policy_gradient}
\end{align}
where 
$\design_k=\mu_{k,w}(\Info_k)$, and
$\tilde{Q}_k^{\policy_w}$ is the \emph{action-value function} (i.e., the \emph{critic}) {induced by the variational one-point reward terms $\tilde{r}_k$ and $\tilde{r}_N$} and defined as:
\begin{align}
\tilde{Q}_k^{\policy_w}(\info_k,\design_k;\phi)&= \EE_{M_{0},\Param_{m,0},\Nuis_{m,0},Z_{m,0}|\info_k}\Bigg[\EE_{Y_{k:N-1}|w,s_0,\info_k,\design_k,M_{0},\Param_{m,0},\Nuis_{m,0}}\bigg[\tilde{r}_k(\info_k,\design_k,{Y_k};\phi) 
\nonumber\\
&\hspace{13.5em}+ \sum_{t=k+1}^{N-1} \tilde{r}_t{(\Info_t,\mu_{t,w}(\Info_t),Y_t;\phi)} + \tilde{r}_N({\Info_N};\phi)\bigg]\Bigg]
\label{e:Q1}\\
&=\EE_{M_{0},\Param_{m,0},\Nuis_{m,0},Z_{m,0}|\info_k}\Bigg[\EE_{Y_{k}|w,s_0,\info_k,\design_k,M_{0},\Param_{m,0},\Nuis_{m,0}}\bigg[ \tilde{r}_k(\info_k,\design_k,{Y_k};\phi) \nonumber\\
&\hspace{18.em}+ \tilde{Q}^{\policy_w}_{k+1}({\Info_{k+1},\mu_{k+1,w}(\Info_{k+1})};\phi)\bigg]\Bigg],
\label{e:Q2}\\
\tilde{Q}_{N}^{\policy_w}(\info_N,\cdot;\phi) &= 
{\EE_{M_{0},\Param_{m,0},\Nuis_{m,0},Z_{m,0}|\info_N} \bigg[} 
\tilde{r}_N(\info_N;\phi) {\bigg]}
,\label{e:Q3}
\end{align}
for $k=0,\ldots,N-1$ and subject to
{$\Info_{k+1}=\{\Info_k, \design_k, Y_k\}$}.  {The value of the critic
  $\tilde{Q}_k^{\policy_w}(\info_k,\design_k;\phi)$} is the {expected remaining
  cumulative reward (i.e., the expected sum of all remaining rewards)} under variational parameters $\phi$,
for performing the $k$th experiment {at} design $\design_k$
{from} state $\info_k$ 
{(i.e., from the lastest posterior that is conditioned on $\info_k$, as indicated by the outer conditional expectation)} 
and thereafter following policy
$\policy_w$.
The critic can also be written in a recursive manner in \cref{e:Q2}. 
To facilitate computing the $\nabla_{\design_k} \tilde{Q}_k^{\policy_w}$ term in \cref{eq:policy_gradient}, we also parameterize the critic functions by $\nu\in \RR^{N_{\nu}}$ and denote the parameterized form as $\tilde{Q}^{\policy_w}_{k,\nu}$.

In summary, the two main steps involve: (a) forming variational posterior approximations $q$ and estimating $\nabla_\phi \tilde{U}$ in \cref{eq:variational_gradient}, and (b) forming approximate critic $\tilde{Q}^{\policy_w}_{k,\nu}$ and estimating $\nabla_w \tilde{U}$ in \cref{eq:policy_gradient}. We detail these two steps below.

\subsection{Variational gradient}
\label{ss:variational_gradient}

Adopting the variational-one-point-TIG rewards into \cref{eq:variational_gradient}, it is easy to verify that
\begin{align}
    \nabla_\phi \tilde{U}(w;\phi) = \EE_{M_{0},\Param_{m,0},
    \Nuis_{m,0},Z_{m,0}}\Bigg[\EE_{\Info_N|w,s_0,M_{0},\Param_{m,0},\Nuis_{m,0}}\Big[ &\alpha_{M} \nabla_{\phi_{M}} \log q(M_{0}|\Info_N;\phi_{M}) \nonumber \\
    &+ \alpha_{\Param} \nabla_{\phi_{\Param_m}} \log q(\Param_{m,0}|\Info_N;\phi_{\Param_m}) 
    \nonumber \\
    &+\alpha_{Z} \nabla_{\phi_{Z_m}} \log q(Z_{m,0}|\Info_N;\phi_{Z_m}) \Big]\Bigg],
    \label{e:variationa_grad_specific}
\end{align}
where the log-prior terms disappear under the gradient operation since they do not depend on $\phi$. 
The variational gradient for the variational-one-point IIG case is exactly the same, since the intermediate variational posteriors all cancel per \cref{app:variational_expected_utility_lower_bound} and the gradient to those terms' $\phi$'s will always be zero. 
In order to obtain good intermediate posterior approximations in the variational-one-point-IIG case, we first note that the variational `partial-length' expected utility up to stage $k$ remains a lower bound to its non-variational counterpart for all $k$---that is, the result in \cref{app:variational_expected_utility_lower_bound} remains true if we replace $\Info_N$ with $\Info_k$, $\forall k$. Then, we can optimize the intermediate posteriors' $\phi$'s by successively maximizing these partial-length lower bounds.

We can then form a Monte Carlo (MC) estimator for the variational gradient in \cref{e:variationa_grad_specific}:
\begin{align}
    \nabla_\phi \tilde{U}(w;\phi) \approx \frac{1}{n} \sum_{i=1}^{n} \Bigg[ &\alpha_{M} \nabla_{\phi_{M}} \log q(m_{0}^{(i)}|\info_N^{(i)};\phi_{M}) \nonumber \\
    &+\alpha_{\Param} \nabla_{\phi_{\Param_m}} \log q(\param_{m,0}^{(i)}|\info_N^{(i)};\phi_{\Param_m}) 
    \nonumber \\
    &+\alpha_{Z} \nabla_{\phi_{Z_m}} \log q(z_{m,0}^{(i)}|\info_N^{(i)};\phi_{Z_m}) \Bigg],
    \label{eq:variational_gradient_MC}
\end{align}
where the samples $m_{0}^{(i)}, \param_{m,0}^{(i)}, \nuis_{m,0}^{(i)}, z_{m,0}^{(i)}$ and corresponding trajectories $\info_N^{(i)}$ are generated following the same procedure described at the beginning of \cref{sec:one_point_utility}.
In practice, multiple steps of gradient update can be applied to $\phi$ at a given $w$ to more efficiently make use of the trajectory samples.

\subsection{Policy gradient}
\label{ss:policy_gradient}

In order to compute the policy gradient $\nabla_w \tilde{U}(w;\phi)$ in \cref{eq:policy_gradient}, we need to first form the approximate critic $\tilde{Q}^{\policy_w}_{k,\nu}$.
This can be achieved by training $\tilde{Q}^{\policy_w}_{k,\nu}$, for any given $\phi$, in a supervised learning manner to optimize $\nu$ 
towards the true $\tilde{Q}^{\policy_w}_{k}$ by minimizing a loss function based on the recursive definition in \cref{e:Q2}. Since the true $\tilde{Q}^{\policy_w}_{k+1}$ values would not be available, they are replaced with the current approximations $\tilde{Q}^{\policy_w}_{k+1,\nu}$:
\begin{align}
    \nu^{\ast} \in \argmin_{\nu} \Bigg\{\CL_{\phi}(\nu) := \frac{1}{n} \sum_{i=1}^{n} \sum_{k=0}^{N-1} \bigg[ &\tilde{Q}^{\policy_w}_{k,\nu}(\info^{(i)}_k, \design^{(i)}_k;\phi) \nonumber\\
    &- \(\tilde{r}_k(\info^{(i)}_k, \design^{(i)}_k,y^{(i)}_k;\phi) + \gamma \tilde{Q}^{\policy_w}_{k+1,\nu}(\info^{(i)}_{k+1}, \design^{(i)}_{k+1};\phi)\) \bigg]^2\Bigg\},
    \label{eq:critic_loss}
\end{align}
where $\gamma \in [0, 1]$ is a weighing factor inserted for regularization.
{In practice, we first sample entire trajectories of $\info_N^{(i)}$ following the procedure described at the beginning of \cref{sec:one_point_utility}. Then, we extract the partial sequences $\info_k^{(i)}$ and designs $\design_k^{(i)}$ resulting from the same sample trajectory $\info_N^{(i)}$ to use for all $k$, instead of generating new $\info_{k+1}^{(i)}$ from their latest posteriors which would be much more expensive. }
The gradient $\nabla_{\nu}\CL_{\phi}(\nu)$ can be obtained readily by differentiating $\tilde{Q}^{\policy_w}_{k,\nu}$ with respect to $\nu$, while the contribution from $\tilde{Q}^{\policy_w}_{k+1,\nu}$ is typically omitted as it is a stand-in of the true $\tilde{Q}^{\policy_w}_{k+1}$ that does not depend on $\nu$. The optimal value of $\nu^{\ast}$ needs not be precisely found, and often just a few steps of gradient update at this $\policy_w$ can already lead to good performance.

The loss $\CL_{\phi}$ from \cref{eq:critic_loss} can be used for both variational-one-point-TIG and -IIG. However, the TIG case has $\tilde{r}_k=0$ for $k=0,\dots,N-1$ (except for possible non-IG reward terms) and as a result it would take many $\nu$ updates from \cref{eq:critic_loss}---especially when $N$ is large---for the non-zero $\tilde{r}_N$ to propagate to the $\tilde{Q}^{\policy_w}_{k,\nu}$'s at early $k$'s. 
This can lead to slow or even divergent policy gradient updates. 
To mitigate this effect, we follow the REINFORCE algorithm~\cite{williams1992simple} and propose a modified loss as a more stable option for the TIG case:
\begin{align}
    \CL_{\phi,T}(\nu) = \frac{1}{n} \sum_{i=1}^{n} \sum_{k=0}^{N-1} \Bigg[ &\tilde{Q}^{\policy_w}_{k,\nu}(\info^{(i)}_k,\design^{(i)}_k;\phi) \nonumber \\
    &- \eta \(\tilde{r}_k(\info^{(i)}_k,\design^{(i)}_k,y^{(i)}_k;\phi) + \gamma \tilde{Q}^{\policy_w}_{k+1,\nu}(\info^{(i)}_{k+1},\design^{(i)}_{k+1};\phi)\) \nonumber \\
    &- (1-\eta) \(\sum_{t=k}^{N-1}\gamma^{t-k}\tilde{r}_t(\info^{(i)}_t,\design^{(i)}_t,y^{(i)}_t;\phi) + \gamma^{N-k} \tilde{r}_N(\info_N^{(i)};\phi) \) \Bigg]^2,
    \label{eq:critic_loss2}
\end{align}
where $\eta$ linearly increases from 0 to 1 during the training process.

Once $\tilde{Q}^{\policy_w}_{k,\nu}$ is obtained, we finally form the MC estimator for the policy gradient in \cref{eq:policy_gradient}:
\begin{align}
    \nabla_w \tilde{U}(w;\phi) \approx \frac{1}{n} \sum_{i=1}^{n} \sum_{k=0}^{N-1} \nabla_w \mu_{k,w}(\info^{(i)}_k) \nabla_{\design_k} \tilde{Q}^{\policy_w}_{k,\nu}(\info^{(i)}_k,\design^{(i)}_k;\phi),
    \label{eq:policy_gradient_MC}
\end{align}
{where $\info_N^{(i)}$ is generated following the same procedure described at the beginning of \cref{sec:one_point_utility} using policy $\policy_w$, and $\info_k^{(i)}$ and $\design^{(i)}_k$ are the partial sequences and designs, respectively, of the full sample sequence $\info_{N}^{(i)}$.}

\subsection{Implementation techniques and the overall algorithm}
\label{sec:final_alg}

\paragraph{Variational posteriors}
We use a neural network (NN) to represent the approximate posterior of model indicator, $q(m_{0}|\info_k;\phi_M)$; the NN takes $\info_k$ as input
and uses a softmax output activation to output model probability. For the approximate posteriors of PoIs and QoIs, respectively $q(\param_{m,0}|\info_k;\phi_{\Theta_m})$ and $q(z_{m,0}|\info_k;\phi_{Z_m})$, we adopt independent Gaussian mixture models (GMMs) with NNs predicting the GMM weights, means, and standard deviations. Truncated normal is used for variables with compact support. We also explore the use of normalizing flows (NFs)~\cite{Dong_25} for posterior approximations.  More details can be found in \cref{app:model_approximation,app:parameter_approximation}. 

\paragraph{Policy and critic networks}
We employ NNs to represent both $\policy_w$ and $\tilde{Q}^{\policy_w}_{k,\nu}$, therefore we also refer to these NNs as the policy network and critic network, respectively. The optimizations of the policy and critic networks are both carried out using Adam~\cite{kingma2014adam} with mini-batching.
More details about these networks can be found in \cref{app:actor_critic,app:critic}.

\paragraph{Target networks}
We adopt target networks for the actor and critic---i.e., secondary networks updated less frequently and with a damping factor---in order to promote stability in the training process by smoothing the learning target and mitigating 
drastic changes of Q-values and designs across training iterations.
Target networks have been shown to greatly improve robustness and effectiveness of policy-gradient-based methods~\cite{lillicrap2015continuous}.

\paragraph{Replay buffer} 
{We use a replay buffer to store past trajectory samples from previously encountered policies, and resample these stored samples for use during the policy optimization process. Replay buffer permits off-policy learning, which entails optimizing the policy using data generated from policies that differ from the current one \cite{watkins1992q}. 
This can help stabilize the optimization and improve sample efficiency by allowing multiple uses of the simulation data, for example those generated from policies in previous iterations and from exploration policies~\cite{mnih2015human}. %
}

\paragraph{Exploration policy} A balance of exploration and exploitation is important for the numerical optimizer to identify a good policy. Insufficient exploration limits understanding of the objective function's global landscape, while too much exploration 
can delay convergence. 
To inject exploration, 
we make use of an {exploration policy} during the policy optimization (i.e., training) phase by adding perturbation to the deterministic base policy:
$\design_k = \mu_{k,w}(\Info_k) + \mathcal{E}_{k,\text{explore}}$,
where $\mathcal{E}_{k,\text{explore}}\sim \CN(0,\Sigma_{k,\text{explore}})$ and $\Sigma_{k,\text{explore}}=\II\sigma_{k,\text{explore}}^2$ is a diagonal covariance whose entries reflect the exploration length scale for each dimension of $\designset_k$. 
A reasonable strategy is to set larger exploration in the early training iterations and reduce it gradually; details will be specified for each numerical case in \cref{sec:results}.
Once the final policy is obtained, its evaluation (i.e., testing phase) will use the deterministic policy only.

\paragraph{Overall algorithm} Pseudocode for the overall vsOED algorithm is presented in \cref{alg:PG-vsOED}. 
While we write the simple gradient ascent update formulas in the pseudocode for illustration, it can be replaced by any other gradient-based updates.

\begin{algorithm}
\caption{The vsOED algorithm. }
\label{alg:PG-vsOED}
\begin{algorithmic}[1]
\STATE{Initialize variational parameters $\phi$, actor (policy) parameters $w$, critic (action-value function) parameters $\nu$;}
\FOR{$l=1,\dots,n_{\text{updates}}$} 
\STATE{Simulate $n_{\text{traj}}$ trajectories: for the $i$th trajectory, sample $m_0^{(i)} \sim P_M$, $\param_{m,0}^{(i)} \sim
p\big(\param_{m,0} \vert m_0^{(i)}\big)$, $\nuis_{m,0}^{(i)} \sim
p\big(\nuis_{m,0} \vert m_0^{(i)}\big)$, $z_{m,0}^{(i)}
\sim p\big(z_{m,0}|\param_{m,0}^{(i)},\nuis_{m,0}^{(i)},m_0^{(i)}\big)$, and then for $k=0,\dots,N-1$ generate $\design_k^{(i)}=\mu_{k,w_{l}}(\info_k^{(i)})+\epsilon_{k,\text{explore}}^{(i)}$ and $y_k^{(i)}\sim p(y_k|m_0^{(i)},\param_{m,0}^{(i)},\nuis_{m,0}^{(i)},\design_k^{(i)},\info_k^{(i)})$;}
\STATE{Update newly generated information sequences $\left\{\info_N^{(i)}\right\}_{i=1}^{n_{\text{traj}}}$ into replay buffer;}
\STATE{Sample $n_{\text{batch}}$ trajectories from the replay buffer;}
\STATE{Use batch trajectories to estimate $\nabla_\phi \tilde{U}$ following \cref{eq:variational_gradient_MC}, update $\phi_{l+1}=\phi_{l}+a_{\phi,l} \nabla_\phi \tilde{U}$ with learning rate $a_{\phi,l}$ (can be done multiple times per $l$-iteration), 
and calculate
$\left\{\tilde{r}_k^{(i)}\right\}_{i=1}^{n_{\text{batch}}}$ with the new $\phi_{l+1}$; 
}
\STATE{
Update $\nu$ towards the $\nu^{\ast}$ in \cref{eq:critic_loss} (or \cref{eq:critic_loss2}), e.g., through multiple steps of gradient ascent\;}
\STATE{Estimate $\nabla_w \tilde{U}$ following \cref{eq:policy_gradient_MC}, and then update $w_{l+1}=w_{l}+a_{w,l} \nabla_w \tilde{U}$  with learning rate $a_{w,l}$; 
}
\ENDFOR
\STATE{Return final policy $\policy_{w}$;}
\end{algorithmic}
\end{algorithm}

\section{Numerical experiments}
\label{sec:results}

We demonstrate vsOED and compare it against other state-of-the-art sequential experimental design methods across a number of problems with varying complexity and that illuminate different challenges. We first describe the demonstration setup in \cref{ss:setup}, and then present results for four cases: Case 1---source location finding
in \cref{sec:source}; Case 2---constant elasticity of substitution (CES) in \cref{sec:ces}; Case 3---SIR model for disease spread 
in \cref{sec:sir}; and Case 4---convection-diffusion-reaction in
\cref{sec:conv_diff}. 

\subsection{Demonstration setup}
\label{ss:setup}

For \textbf{vsOED}, we employ GMMs and NFs for posterior approximation, and variational-one-point-TIG and -IIG for reward formulation: we adopt the naming convention where, for example,  \textbf{vsOED-G-I} stands for \textbf{G}MM with \textbf{I}IG, and \textbf{vsOED-N-T} for \textbf{N}Fs with \textbf{T}IG. 
Other sequential experimental design algorithms being compared include \textbf{Random} design,
\textbf{DAD}~\cite{foster2021deep}, \textbf{iDAD}~\cite{ivanova2021implicit}, and a stochastic policy based \textbf{RL} method \cite{blau2022optimizing}. \textbf{RL} can also be combined with TIG and IIG, denoted by \textbf{RL-T} and \textbf{RL-I} respectively.
{DAD} and {iDAD} require the derivative of 
the forward model, {vsOED} and {iDAD} can 
accommodate implicit likelihood, and {vsOED} can handle multiple models through the model indicator.
\Cref{tab:methods} summarizes which of these aspects have been considered or studied with these algorithms.
For DAD, iDAD, and RL, we directly use code from their original publications' Github repositories.

\begin{table}[htbp]
    \centering
    \caption{Properties of different sequential experimental design methods.}
    \begin{tabular}{ccccc}
    \toprule
      & Adaptive 
     & Implicit likelihood & No model derivative & Multiple models \\
    \midrule 
    Random  & \xmark & \cmark & \cmark & \cmark \\
    DAD~\cite{foster2021deep}  & \cmark & \xmark & \xmark & \xmark \\
    iDAD~\cite{ivanova2021implicit}  & \cmark & \cmark & \xmark & \xmark \\
    RL~\cite{blau2022optimizing}  & \cmark & \xmark & \cmark & \xmark \\
    \textbf{vsOED}  & \cmark & \cmark & \cmark & \cmark \\
    \bottomrule
    \end{tabular}
    \label{tab:methods}
\end{table}

In this work, we pay special attention to study algorithm efficiency in terms of trajectory samples since they entail the majority of forward model runs (i.e., evaluating $G_k$ and $H$). We make comparisons under two computational settings: (i) `fully trained' where the policies are trained using algorithms' default (usually a conservatively large) number of trajectory samples from their respective publications---8 trillion,  100 billion, and 200 million for RL, DAD, and iDAD, respectively---and we will use 10 million for vsOED; and (ii) `limited budget' where we restrict to a significantly smaller total budget of 10 million training trajectories for all algorithms and cases (except for Case 3---SIR model where the budget is set to 1 million).

To ensure fair comparisons, it is important to evaluate the final policies resulting from different algorithms in a common manner. 
To this end, we adopt PCE 
\cite{foster2020unified,foster2021deep}, specifically its sequential version, to evaluate the final expected utilities 
for policies resulting from `OED for model indicator' and `OED for PoIs' when likelihood is explicit and when there are no nuisance parameters (see \ref{app:reward_one_point} for details on PCE, including a new variant we derive to handle `OED for model indicator'). 
PCE itself is intractable to compute due to its outer expectation operator, and must be estimated numerically, for example by MC integration of the expectation. 
The final MC estimation of PCE results in a nested loop structure; in our numerical cases, we use 2000 trajectory samples for the outer loop and $10^6$ samples for the inner loop to achieve a high-quality estimation; when multiple models are involved, we split the inner samples to $10^6/|\CM|$ for each model so that the total number of inner loop samples remains the same overall. 
In the remainder of the paper, we will refer to `the MC estimate of PCE' simply as `PCE' for brevity.
PCE cannot be used for other OED scenarios involving QoIs and nuisance parameters because the prior predictive densities and likelihood in those scenarios are intractable;
similarly, PCE cannot be used for cases where likelihood is implicit. For those situations, we instead evaluate the policies by calculating a high-quality MC estimate of $\tilde{U}$ using $10^6$ trajectory samples (except for Case 3---SIR model where $3\times 10^5$ samples are used). In the remainder of the paper, we similarly refer to `the MC estimate of $\tilde{U}$' simply as `$\tilde{U}$' for brevity.

All numerical experiments are conducted on the University of Michigan Great Lakes Slurm High Performance Computing Cluster nodes, where
each node is equipped with a single Nvidia Tesla A40 or V100 GPU. All experiments are implemented in Python using PyTorch.

\subsection{Case 1: source location finding}
\label{sec:source}

We adapt the source location finding problem from \cite{foster2021deep}. 
We enlist three candidate models with uniform model prior $P(m)=1/3$ for $m=1,2,3$. The $m$th model contains $m$ sources randomly located in a two-dimensional (2D) domain, each emitting a signal that decays inversely with the square of the distance. The PoIs are the source locations $\Param_m=\{\Param_{m,i}\}_{i=1}^m$ where $\Param_{m,i} = [\Param_{m,i,x},\Param_{m,i,y}] \in \RR^2$ denotes the location of the $i$th source and is endowed with independent priors 
$\Param_{m,i,x},\Param_{m,i,y} \sim \CN(0,1^2)$.
The total intensity at location $[x,y]\in \RR^2$, aggregated from all sources, is
\begin{align}
    G(\Param_m, x,y; m) = a_{\text{bg}} + \sum_{i=1}^m \frac{1}{a_{\text{max}} + \norm{\Param_{m,i} - [x,y]}{2}^2}, \nonumber
\end{align}
where $a_{\text{bg}}=10^{-1}$ is the background signal and $a_{\text{max}}=10^{-4}$ is the maximum signal. 
The design variables $\design_k=[\design_{k,x},\design_{k,y}] \in[-4, 4]^2$ entail selecting the location of measurement within the allowable design space. 
The observation model is then ${Y}_k = G(\Param_m,\design_{k,x},\design_{k,y};m) + \mathcal{E}_k$,
where $\mathcal{E}_k \sim \CN(0, \sigma_{\epsilon}^2)$ with $\sigma_{\epsilon}=0.5$. 

Additionally, we are also interested in a QoI that is the log of the total flux magnitude over an infinite vertical wall located at $x=6$. 
The flux vector $J$ at a spatial location $[x,y]$ is given by Fick's law:
\begin{align}
    J(\Param_m,x,y;m) = -D\nabla_{(x,y)}{ G(\Param_m,x,y;m)}, 
    \nonumber
\end{align}
where $D=1$ is the scalar diffusivity. The total flux across the vertical wall at $x=6$ is then
\begin{align}
    \varphi(\Param_m;m) = \int_{-\infty}^{+\infty} J(\Param_m, x=6,y; m) \,\text{d}y 
    = \sum_{i=1}^m - \frac{\pi (\Param_{m,i,x}-6)}{\( 
a_{\text{max}} + (\Param_{m,i,x}-6)^2 \)^{\frac{3}{2}}}.
\label{e:flux}
\end{align}
At last, the final QoI is $Z_m=\log{\abs{\varphi(\Param_m;m)}}$.

\subsubsection{Case 1a: single-model}
\label{sec:source_uni}

Let us first consider the single-model case where $m$ is fixed at 2.

\begin{figure}[htbp]
  \centering
  \subfloat[Expected cumulative utility at different experiment stage $k$ for policies optimized for $N=30$ ]{\label{fig:source_un\Info_po\Info_EU_vs_stage}\includegraphics[width=0.49\linewidth]{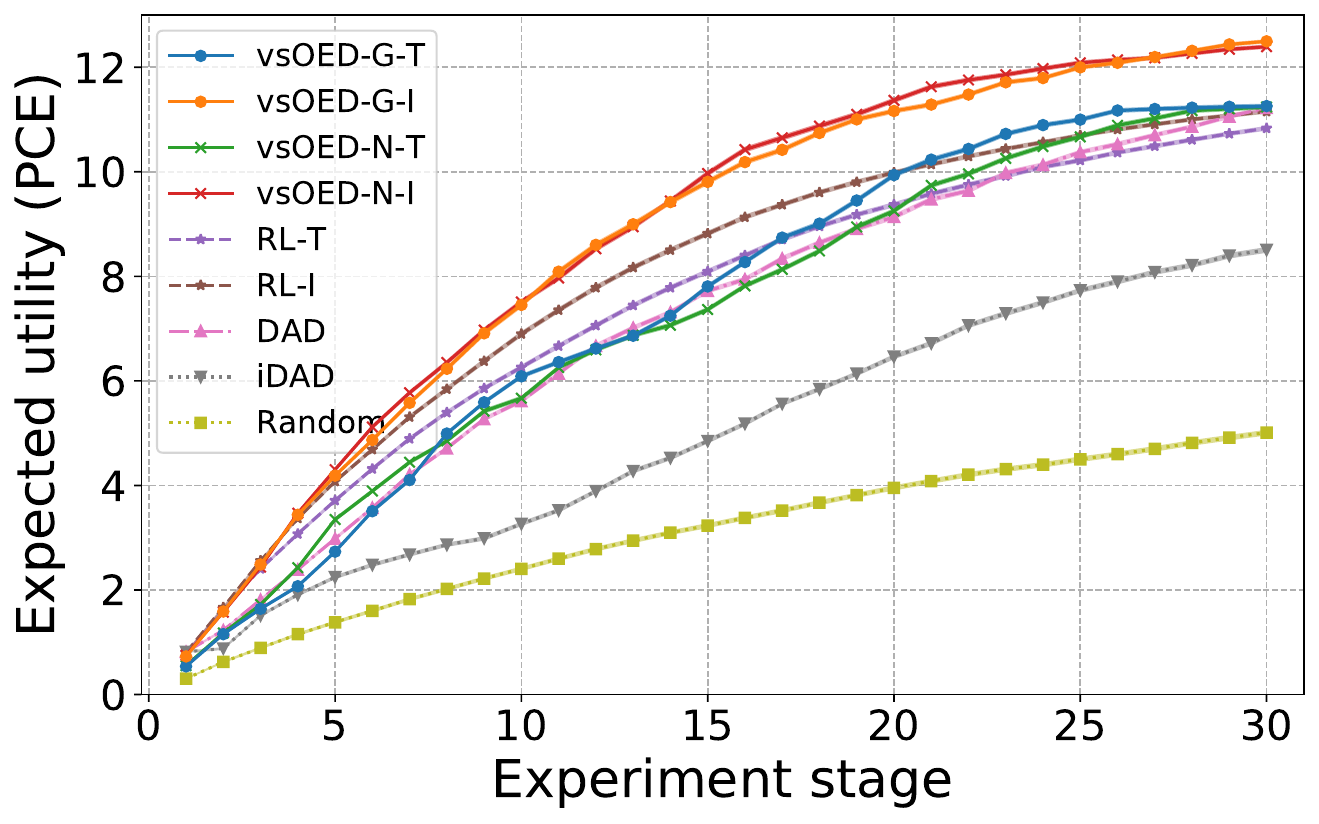}}
  \subfloat[Average expected utility over four training replicates versus design horizon $N$]{\label{fig:source_uni_poi_EU_vs_horizon}\includegraphics[width=0.49\linewidth]{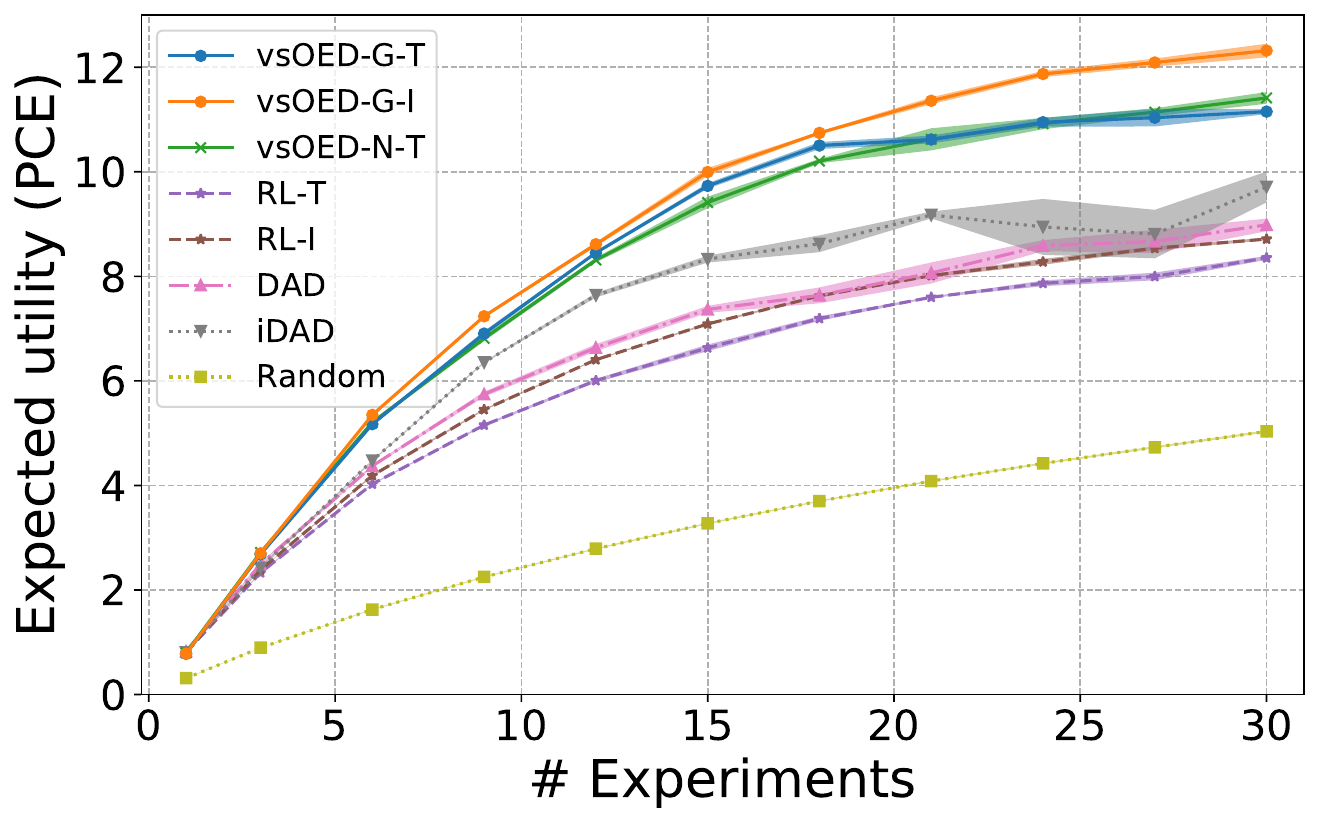}}
  \caption{Case 1a. Expected utility comparisons using policies resulting from different algorithms.
  The shaded regions represent the standard error.
  }
  \label{fig:source_un\Info_po\Info_EU}
\end{figure}

\paragraph{OED for PoIs (design for inference)} The first example entails design for inferring the source location PoIs $\Param$,
with $\alpha_\Param=1$ and $\alpha_Z=0$; this setup is identical to those from previous literature~\cite{foster2021deep, ivanova2021implicit, blau2022optimizing}. 
\Cref{fig:source_un\Info_po\Info_EU_vs_stage} presents the expected cumulative utility, evaluated using PCE, at different experiment stages $k$ for policies optimized for a horizon of $N=30$ experiments under the fully trained setting.
The plot suggests vsOED with IIG to achieve noticeably higher expected utilities throughout, while vsOED with TIG is lower in comparison but maintains a similar level as other algorithms at later stage numbers while lower in earlier stages.

\begin{figure}[htbp]
  \centering
  \subfloat[Example 1]{\label{fig:source_uni_poi_post_h3_comp_1}
  \includegraphics[width=0.48\linewidth]{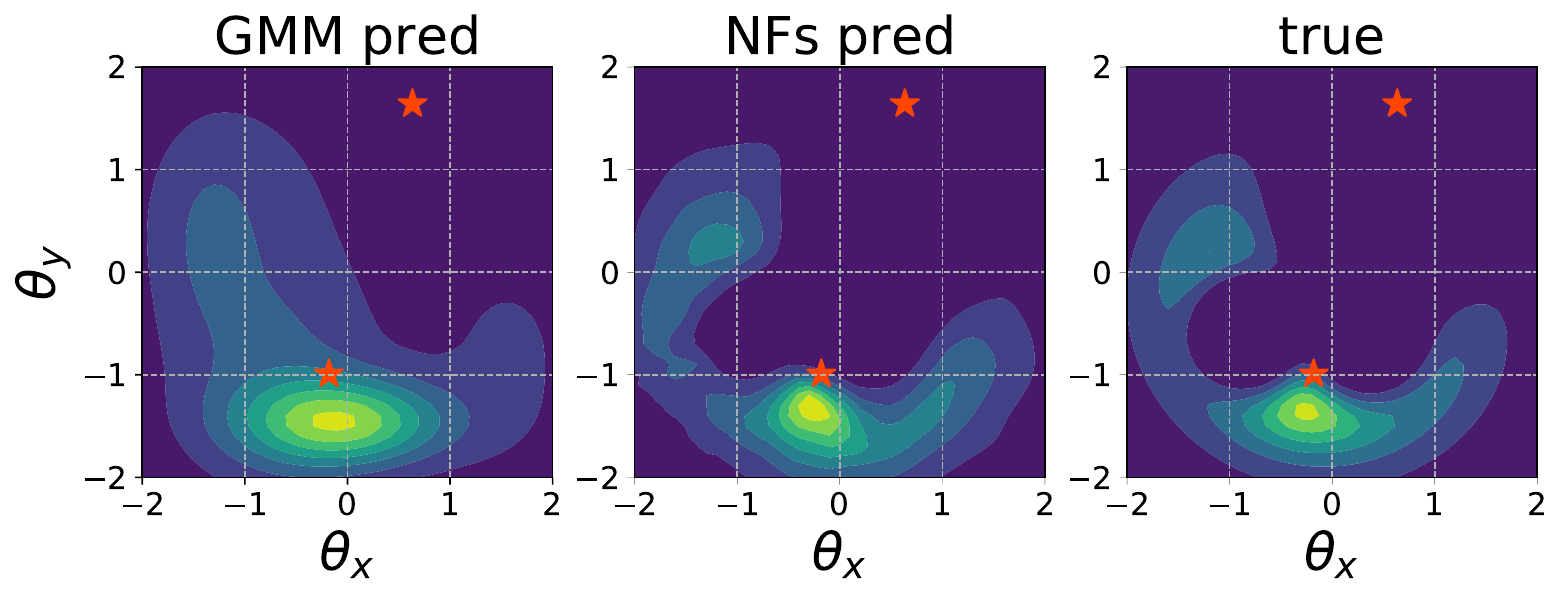}
  }
  \subfloat[Example 2]{\label{fig:source_uni_poi_post_h3_comp_2}
  \includegraphics[width=0.48\linewidth]{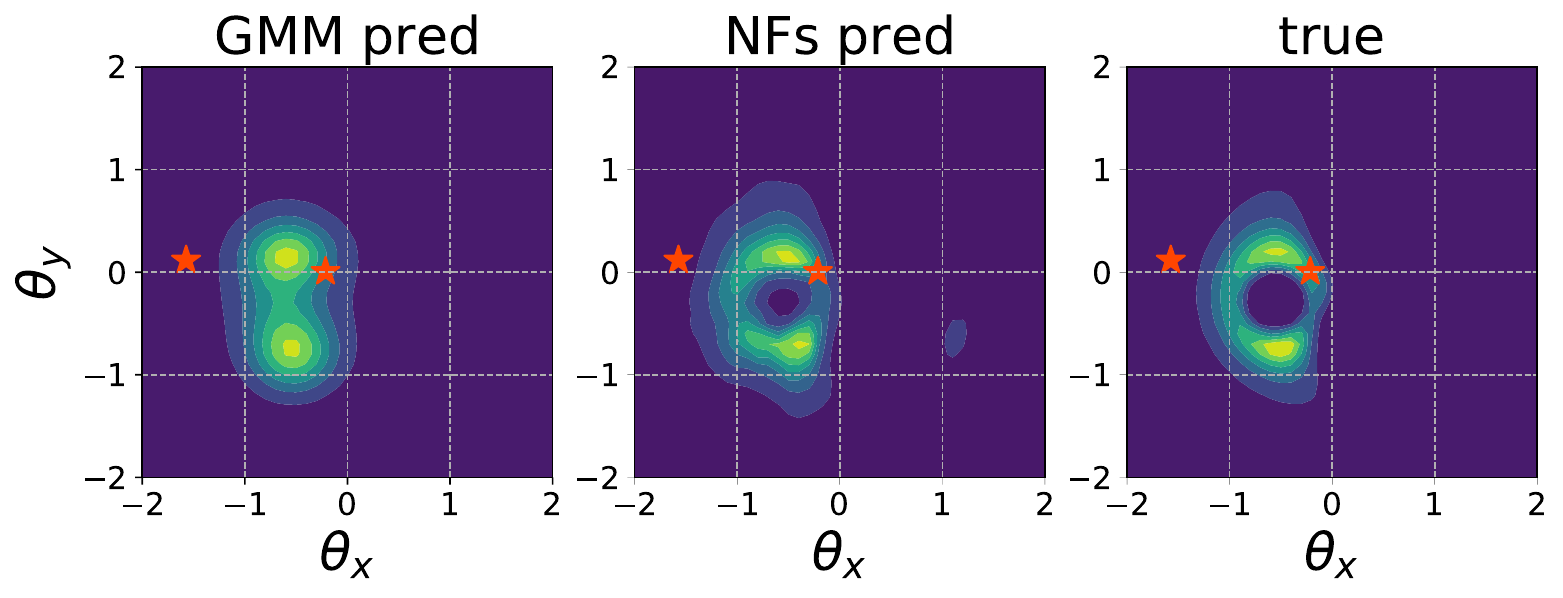}
  }
  \caption{Case 1a. Examples of GMM and NFs approximate posterior and true posterior for PoIs. The red stars indicate the true data-generating parameter values for these trajectory samples.
  }
  \label{fig:source_uni_poi_post_h3_comp}
\end{figure}

\Cref{fig:source_uni_poi_EU_vs_horizon} presents the average expected utility, averaged over four training replicates with different random seeds where each replicate is again evaluated using PCE, for policies trained under different design horizons $N$ under the limited budget setting---that is, each point corresponds to policies optimized specifically for that horizon.
Here with limited budget, vsOED outperforms other algorithms throughout.
IIG again performs better than TIG, especially for $N>15$. 
The standard errors are plotted as shaded regions 
and illustrate general training robustness
with iDAD exhibiting some instability at longer horizons.
Lastly, \cref{fig:source_uni_poi_post_h3_comp} provides examples showing that GMMs and NFs can effectively approximate posteriors even when they are highly non-Gaussian.

\begin{figure}[htbp]
  \centering
  \includegraphics[width=0.6\linewidth]{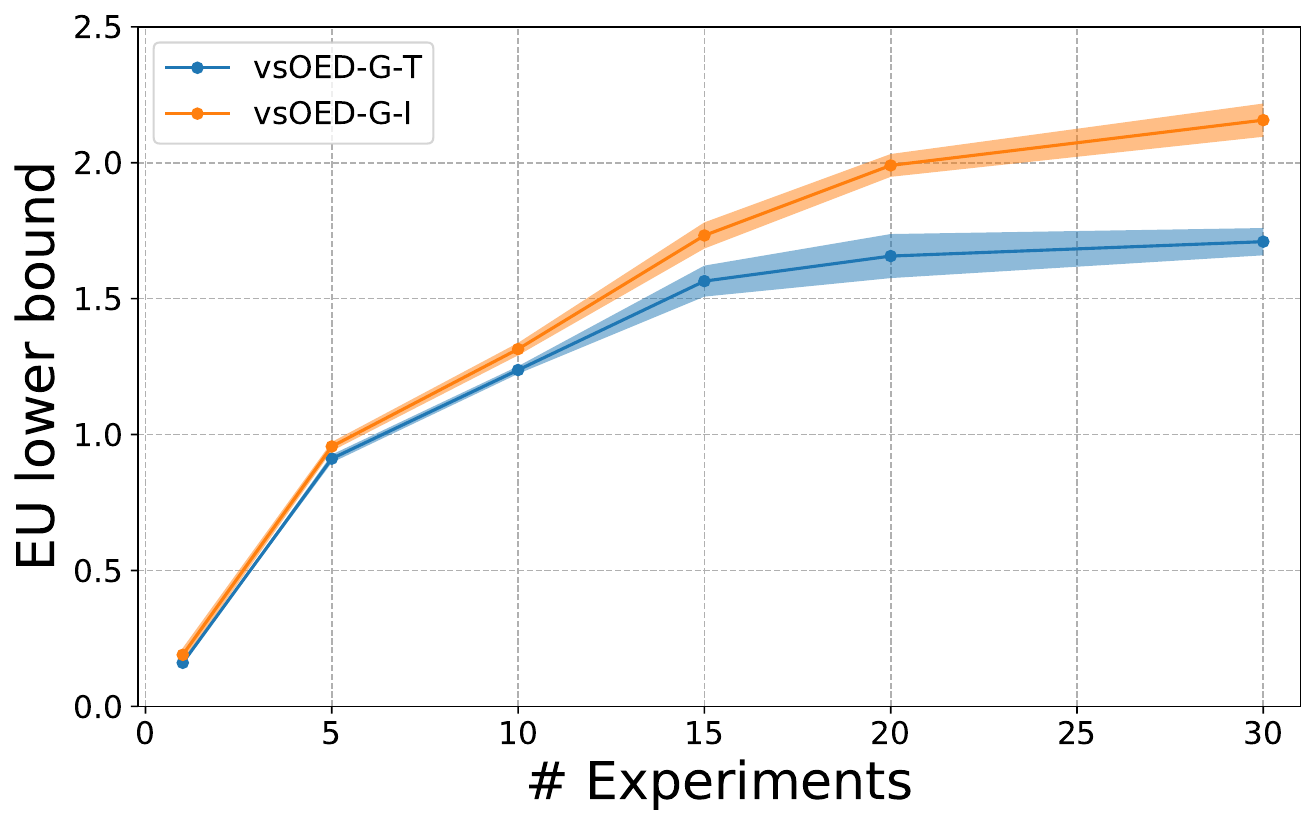}
  \caption{Case 1a.
  Average $\tilde{U}$ over four training replicates for  `OED for QoIs'. The shaded regions represent the standard error. 
  }
  \label{fig:source_uni_goal_EU_vs_horizon}
\end{figure}

\begin{figure}[htbp]
  \centering
  \subfloat[OED for PoIs]{\label{fig:source_uni_poi_policy}\includegraphics[width=0.5\linewidth]{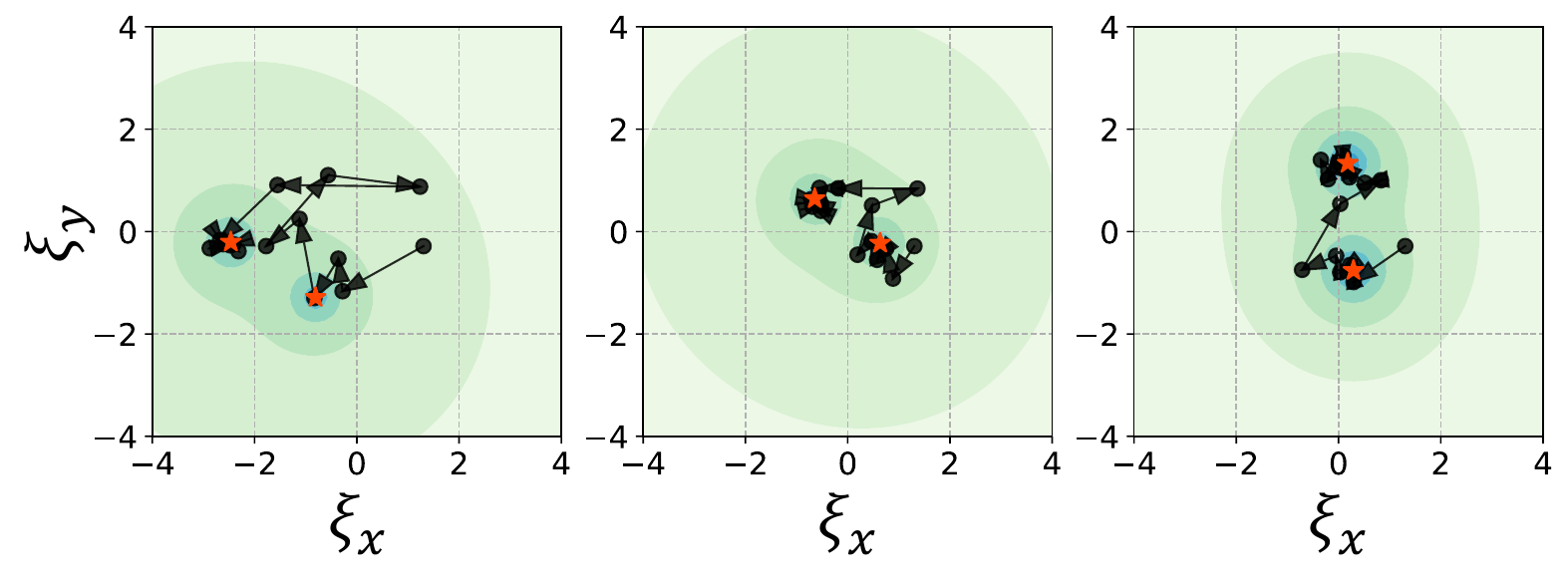}}
  \subfloat[OED for QoIs]{\label{fig:source_uni_goal_policy}\includegraphics[width=0.5\linewidth]{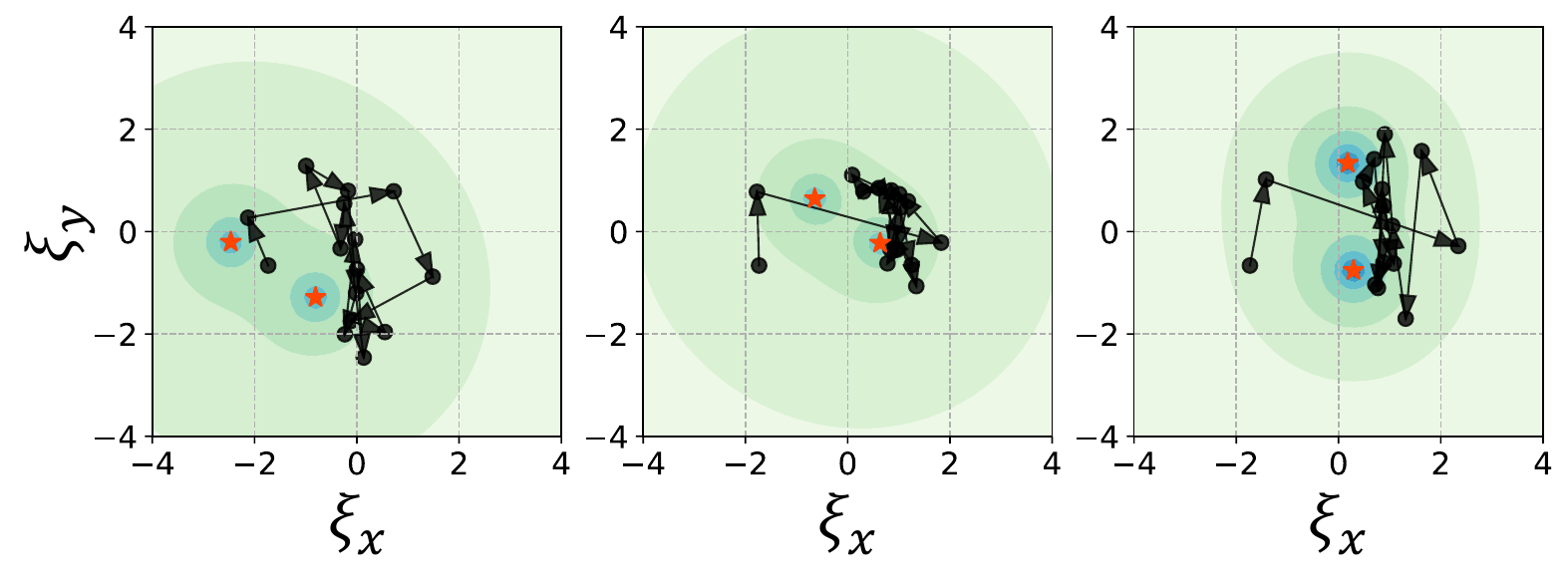}}
  \caption{Case 1a. Examples of policy trajectory for $N=15$. The contour background plots the true signal strength, and the red stars indicate the true source locations.}
  \label{fig:source_uni_policy}
\end{figure}

\begin{figure}[htbp]
  \centering
  \subfloat[Vertical and horizontal designs and posteriors]{\label{fig:source_uni_goal_ver_vs_hor}\includegraphics[width=0.48\linewidth]{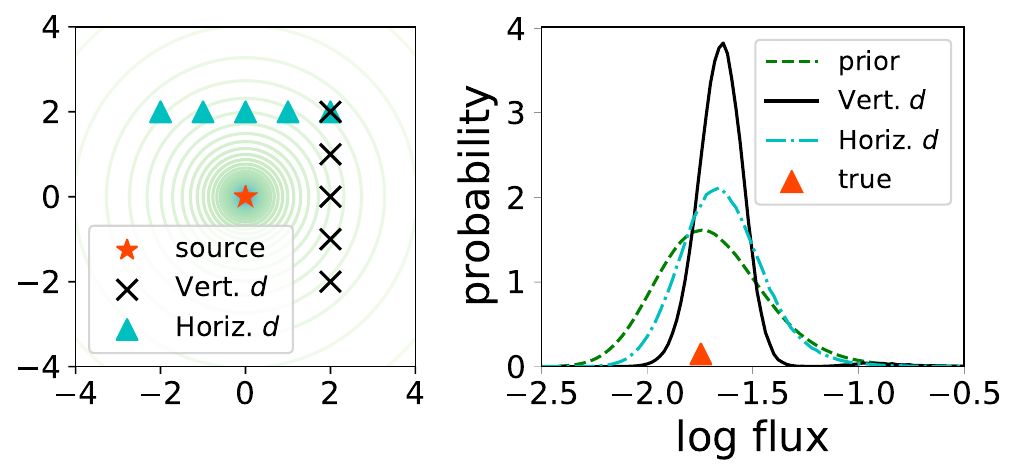}}
  \subfloat[Examples of QoI posterior at $N=15$]{\label{fig:source_uni_goal_post_h15}\includegraphics[width=0.52\linewidth]{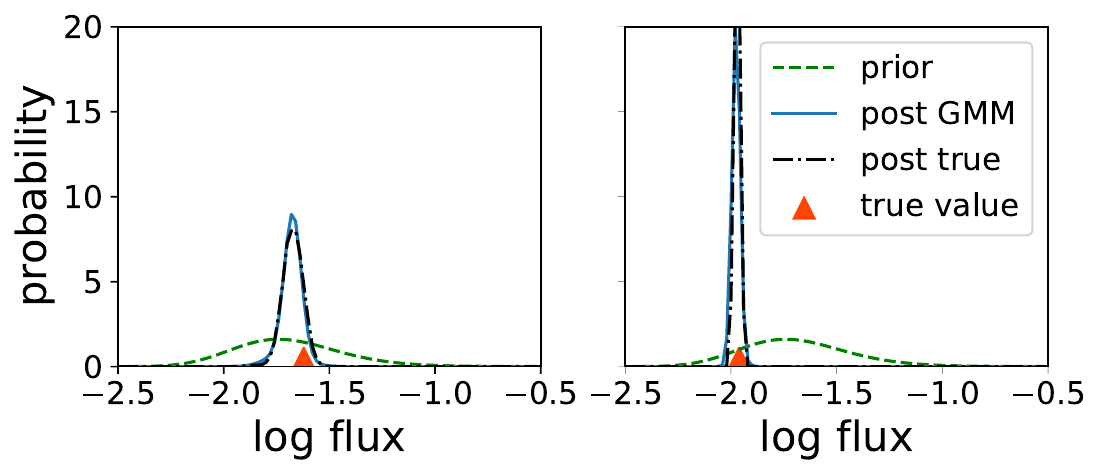}}
  \caption{Case 1a. Design and posterior comparisons for `OED for QoIs'.}
  \label{fig:source_uni_goal_post}
\end{figure}

\paragraph{OED for QoIs (design for prediction)} 
The second example involves goal-oriented design for predicting the flux-based QoI $Z$ introduced earlier, with
$\alpha_\Param=0$ and $\alpha_Z=1$. Only vsOED is used here since the other algorithms cannot accommodate `OED for QoIs'; furthermore, only GMM version of vsOED is presented since the QoI $Z$ is a scalar and unsuitable for the particular architecture  of NFs we adopt that requires at least two dimensions for decomposition (see \cref{app:NFs}).
\Cref{fig:source_uni_goal_EU_vs_horizon} plots the average $\tilde{U}$, averaged over four training replicates, 
versus $N$.
The standard errors are plotted as shaded regions illustrating training robustness for vsOED. As observed in `OED for PoIs', we again see higher expected utility estimates reached by IIG over TIG. 
\Cref{fig:source_uni_policy} presents examples of trajectory using policies resulting from `OED for PoIs' and `OED for QoIs' for $N=15$:
the former tends to move toward the estimated source locations while the latter forms a roughly vertical design pattern. We can explain this behavior through physics. In \cref{e:flux}, since the integration is over the entire $y$, the flux is solely dependent on the $x$-position of the source (i.e., $\Param_{m,i,x}$). Due to the isotropic nature of source emission, spreading measurements along a vertical line is thus more sensitive at detecting changes in $\param_x$. We see this effect in \cref{fig:source_uni_goal_ver_vs_hor}, where a greater QoI posterior shrinkage takes place under a vertical sensor design than a horizontal sensor design. \Cref{fig:source_uni_goal_post_h15} supports that GMMs can also capture well the QoI posteriors.

\paragraph{Hyperparameters and training stability} Hyperparameter settings and training stability results can be found in \cref{app:source}.

\subsubsection{Case 1b: multi-model}

We now consider a multi-model case for the source location finding problem, with three candidate models. Five scenarios are considered: OED for model indicator (i.e., design for model discrimination, $\alpha_{M}=1$, $\alpha_\Param=\alpha_Z=0$), OED for PoIs (i.e., design for parameter inference, $\alpha_\Param=1$, $\alpha_{M}=\alpha_Z=0$), OED for QoIs (i.e., design for goal-oriented prediction, $\alpha_Z=1$, $\alpha_{M}=\alpha_\Param=0$), OED for both model indicator and PoIs (`model-PoIs', $\alpha_{M}=\alpha_\Param=1$, $\alpha_Z=0$), and OED for both model indicator and QoIs (`model-QoIs', $\alpha_{M}=\alpha_Z=1$, $\alpha_\Param=0$).

\begin{figure}[htbp]
  \centering
  \subfloat[OED for model indicator]{\label{fig:source_multi_model_EU_vs_horizon}\includegraphics[width=0.4\linewidth]{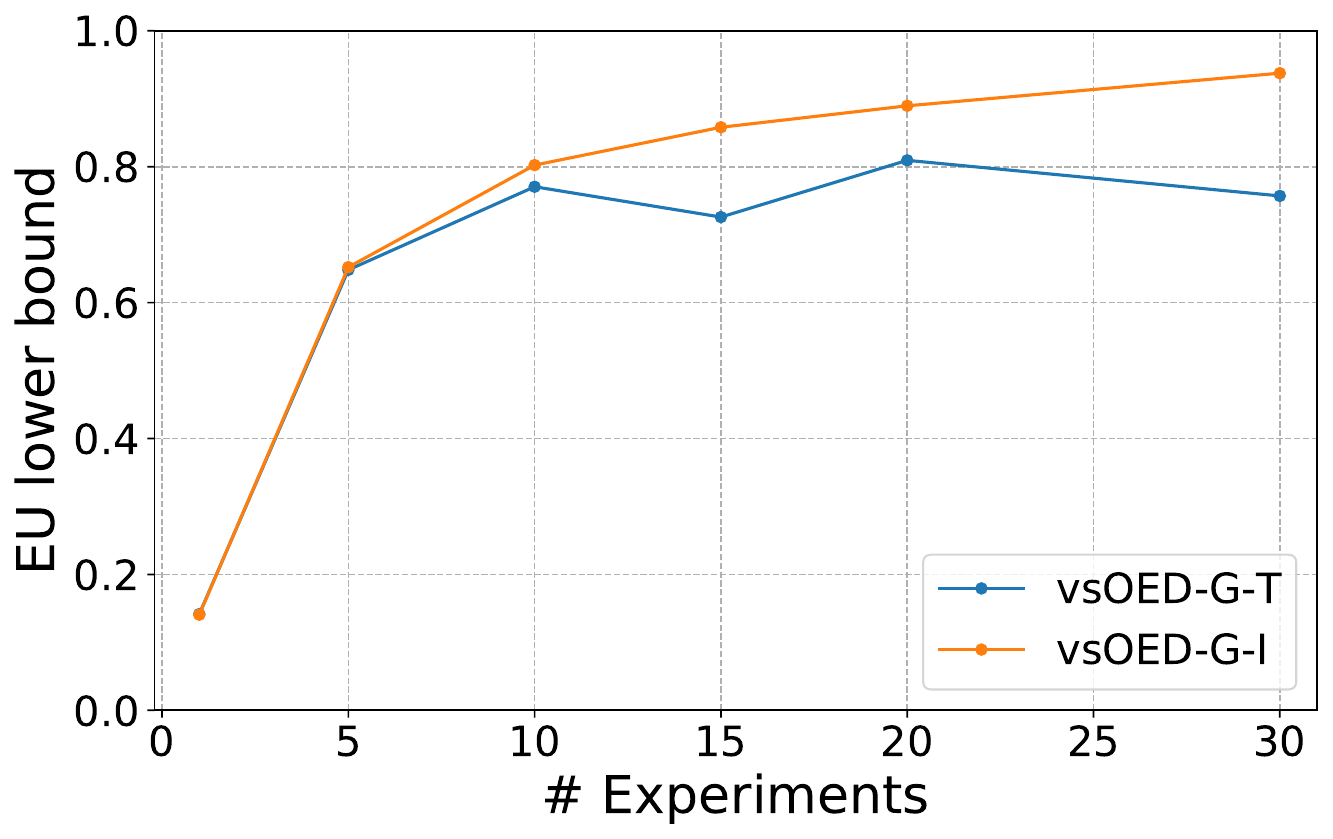}}
  \\
  \subfloat[OED for PoIs]{\label{fig:source_multi_poi_vs_horizon}\includegraphics[width=0.4\linewidth]{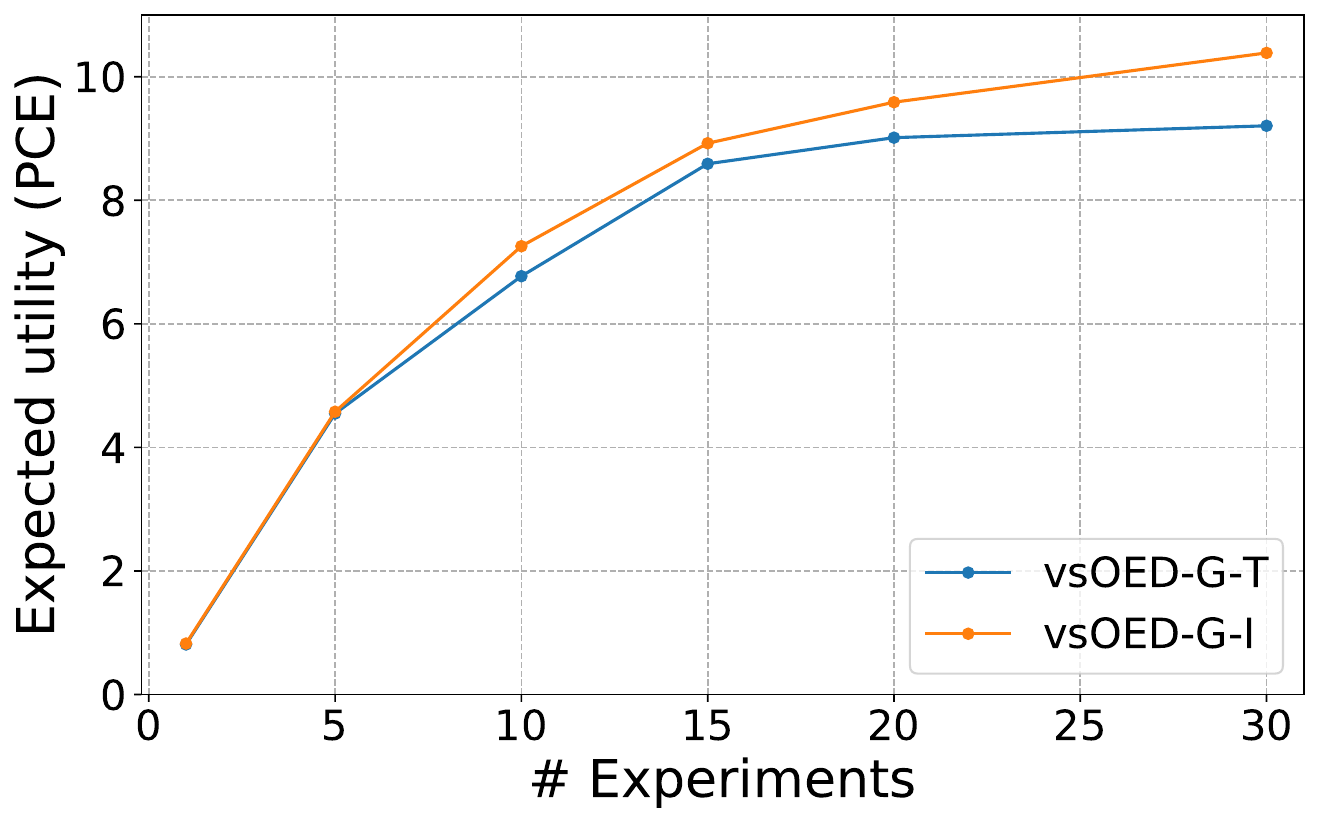}}
  \hspace{1em}
  \subfloat[OED for for QoIs]{\label{fig:source_multi_goal_EU_vs_horizon}\includegraphics[width=0.4\linewidth]{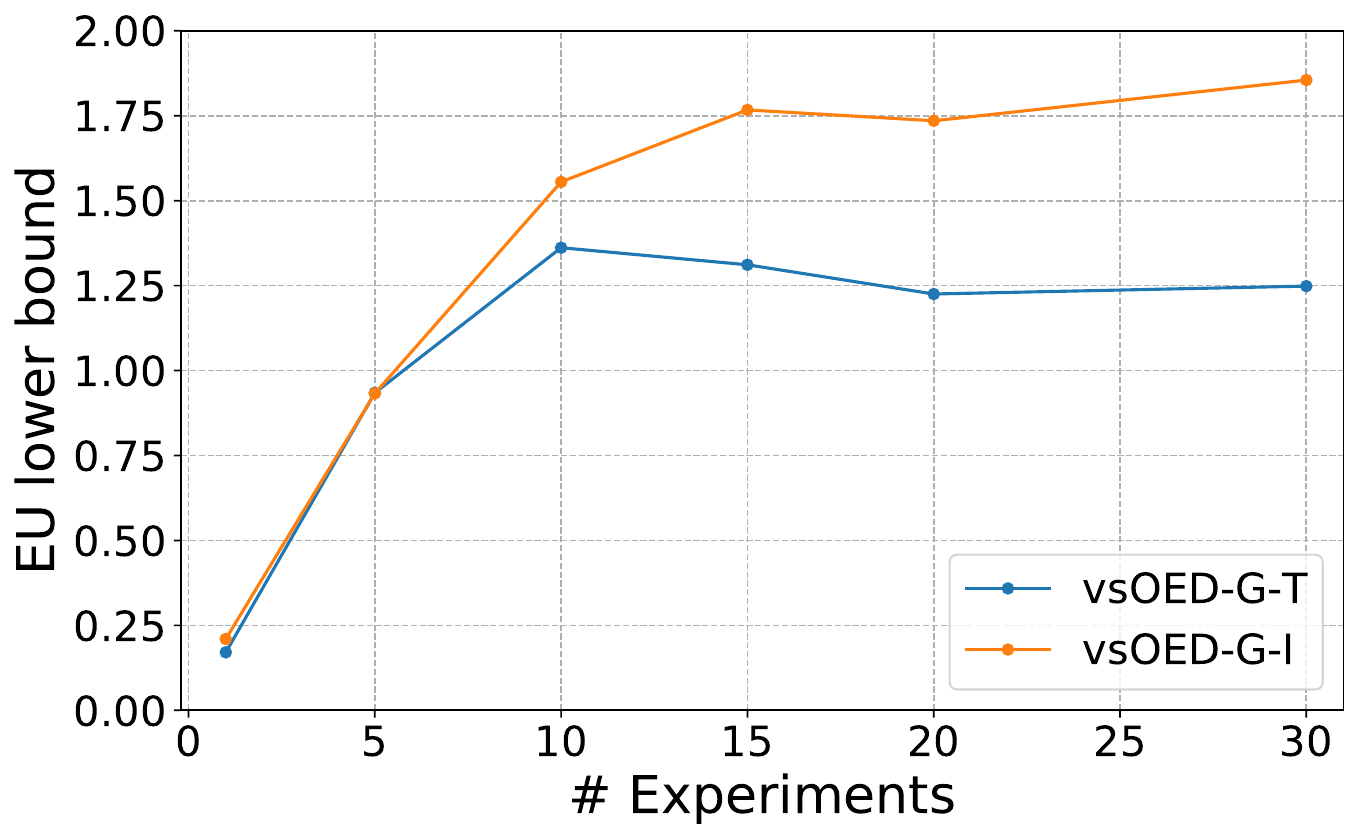}}
  \\
  \subfloat[Model-PoIs OED]{\label{fig:source_multi_model_poi_EU_vs_horizon}\includegraphics[width=0.4\linewidth]{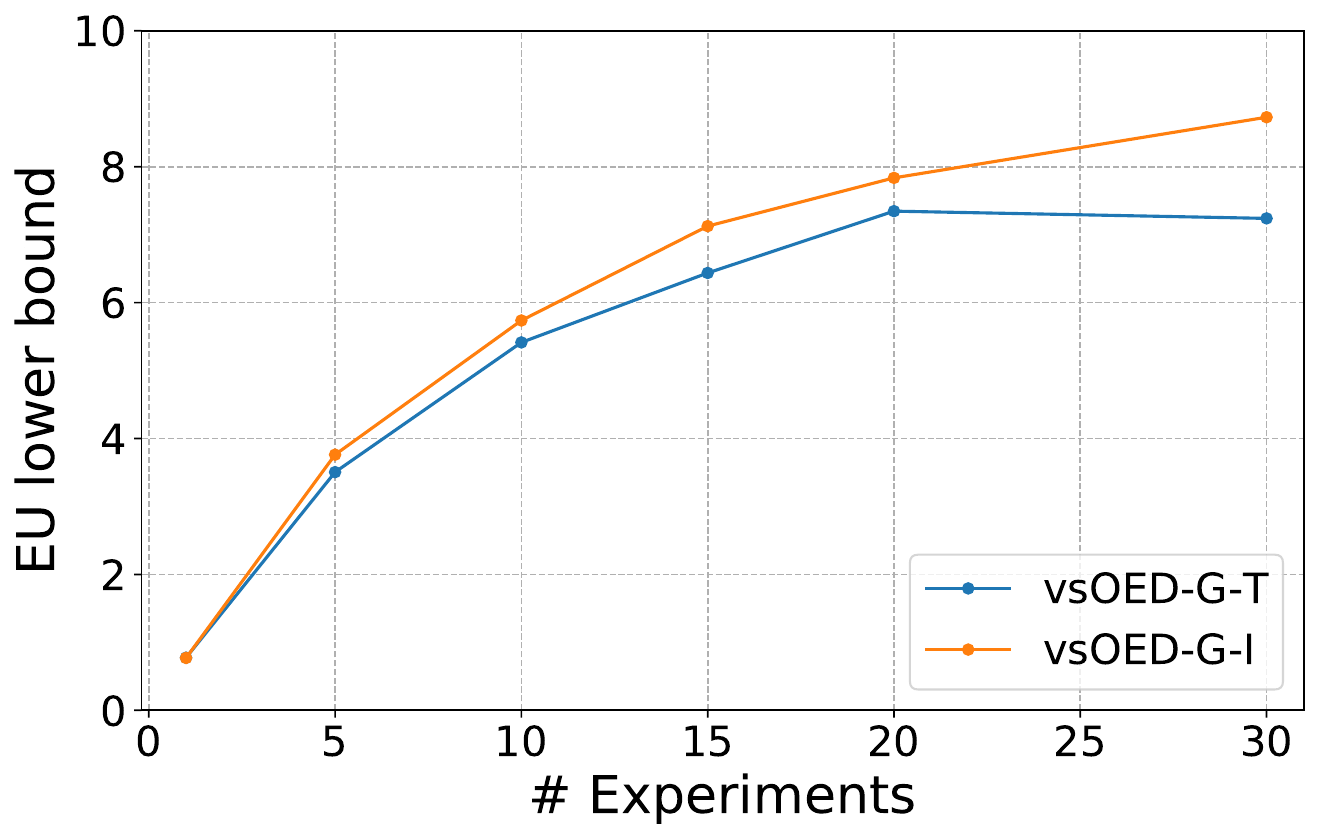}}
  \hspace{1em}
  \subfloat[Model-QoIs OED]{\label{fig:source_multi_model_goal_EU_vs_horizon}\includegraphics[width=0.4\linewidth]{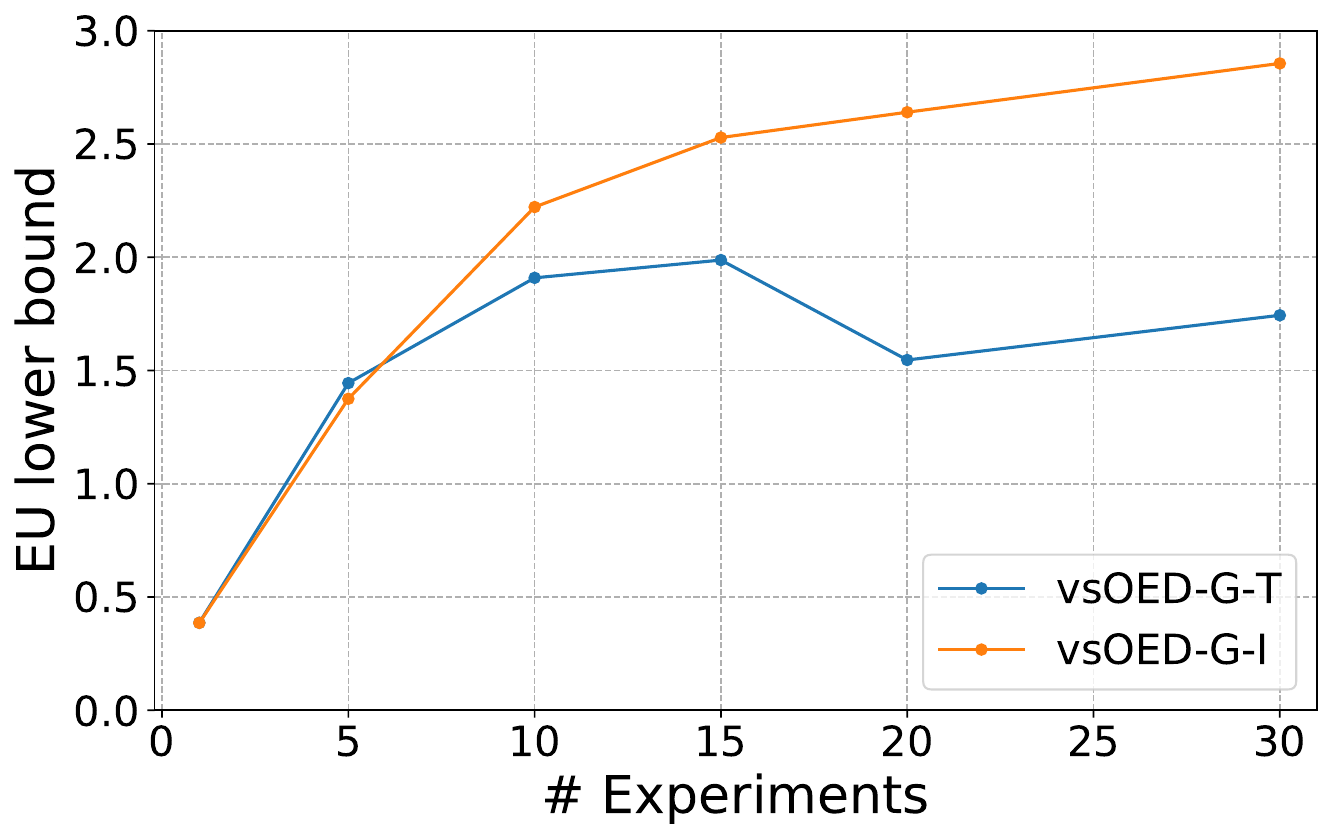}}
  \caption{Case 1b. Average expected utility or $\tilde{U}$ over two training replicates versus design horizon $N$ using policies resulting from the five OED scenarios. }
  \label{fig:source_multi}
\end{figure}

\begin{figure}[htbp]
  \centering
  \subfloat[OED for model indicator]{\label{fig:source_multi_model_policy}\includegraphics[width=0.9\linewidth]{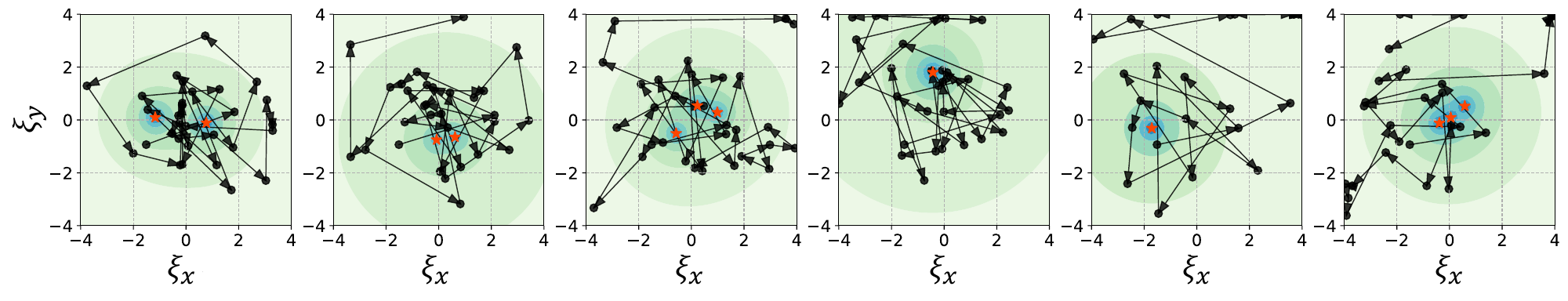}}
  \\
  \subfloat[OED for PoIs]{\label{fig:source_multi_poi_policy}\includegraphics[width=0.9\linewidth]{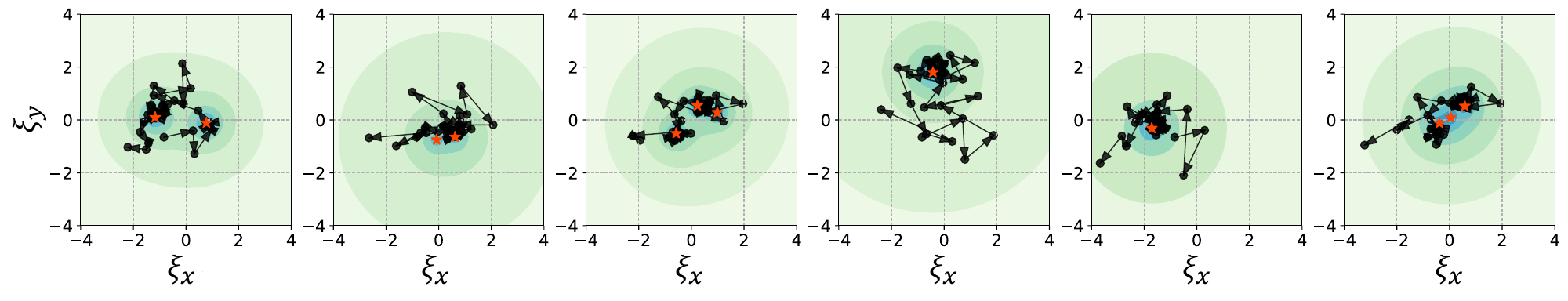}}
  \\
  \subfloat[OED for QoIs]{\label{fig:source_multi_goal_policy}\includegraphics[width=0.9\linewidth]{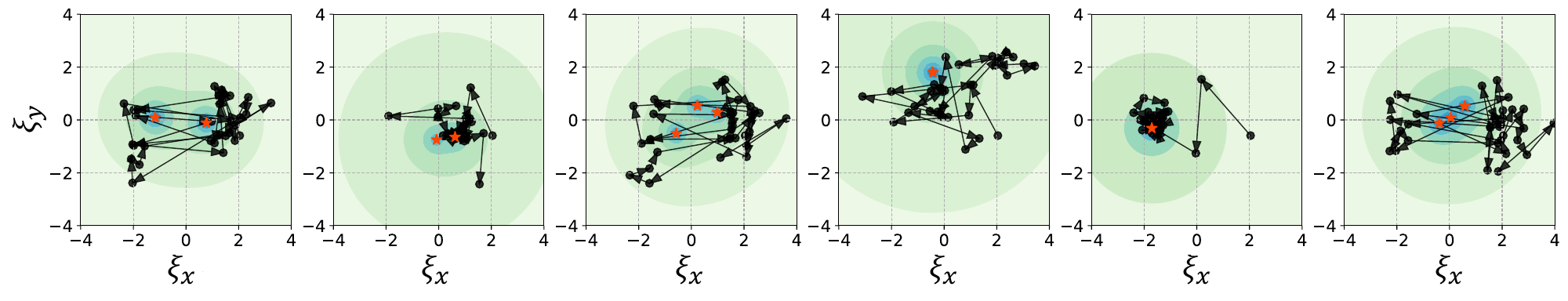}}
  \\
  \subfloat[Model-PoIs OED]{\label{fig:source_multi_model_poi_policy}\includegraphics[width=0.9\linewidth]{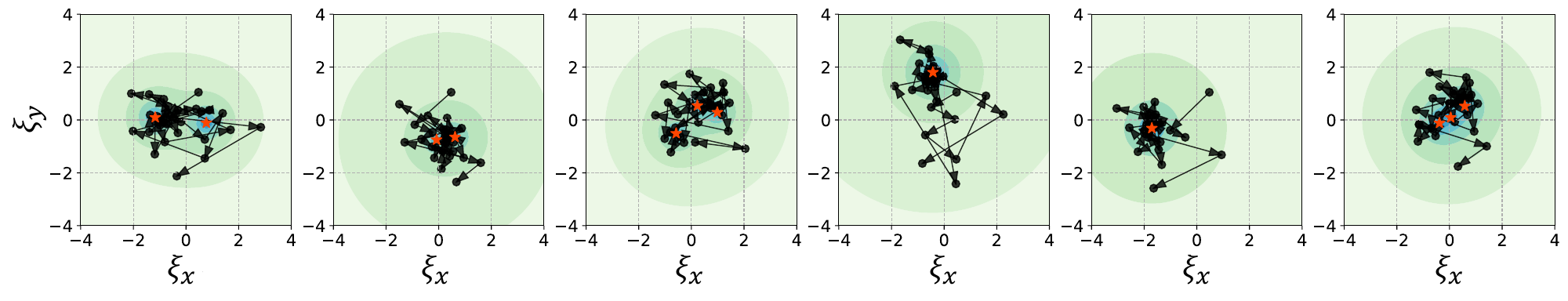}}
  \\
  \subfloat[Model-QoIs OED]{\label{fig:source_multi_model_goal_policy}\includegraphics[width=0.9\linewidth]{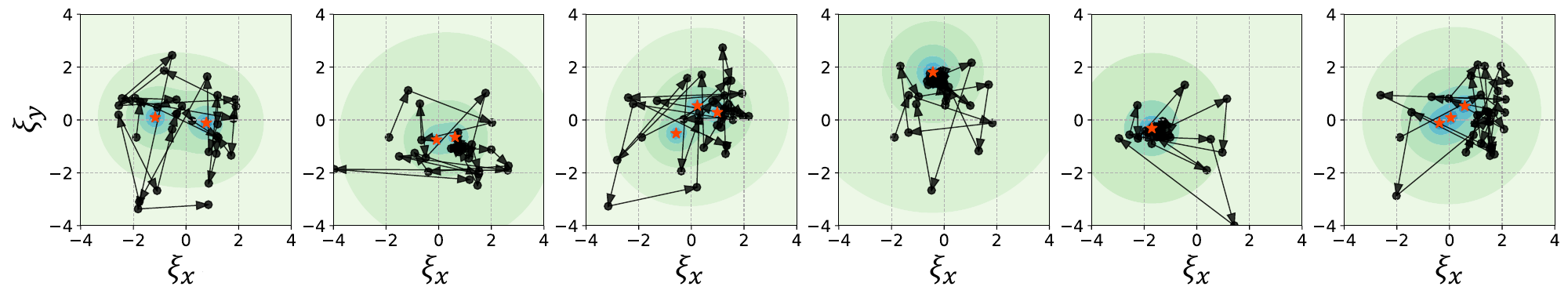}}
  \caption{Case 1b. Examples of policy trajectory for $N=30$. The contour background plots the true signal strength, and the red stars indicate the true source locations.
  }
  \label{fig:source_multi_policy}
\end{figure}

\paragraph{Results}
Only vsOED is used in this case since the other algorithms cannot handle multiple models. Only the GMM version of vsOED is presented for brevity as this example focuses on vsOED capabilities across the five OED scenarios.
\Cref{fig:source_multi} presents the average expected utilities or $\tilde{U}$ over two training replicates for the five OED scenarios.
IIG again demonstrates greater performance over TIG, especially for $N\geq 15$.
\Cref{fig:source_multi_policy} presents examples of trajectory using  policies resulting from the five OED scenarios for $N=30$. Generally, trajectories from `OED for model indicator' tend to explore more and appear more diffuse, while those from `OED for PoIs' tend to be more exploitative and remain closer to the estimated source location.
Similar to the single-model case, `OED for QoIs' promotes more vertically aligned design locations. The model-PoIs and model-QoIs scenarios appear more diffuse than `OED for PoIs' and `OED for QoIs', respectively, due to the addition of exploratory property from `OED for model indicator'.

\begin{figure}[htbp]
  \centering
  \includegraphics[width=1\linewidth]{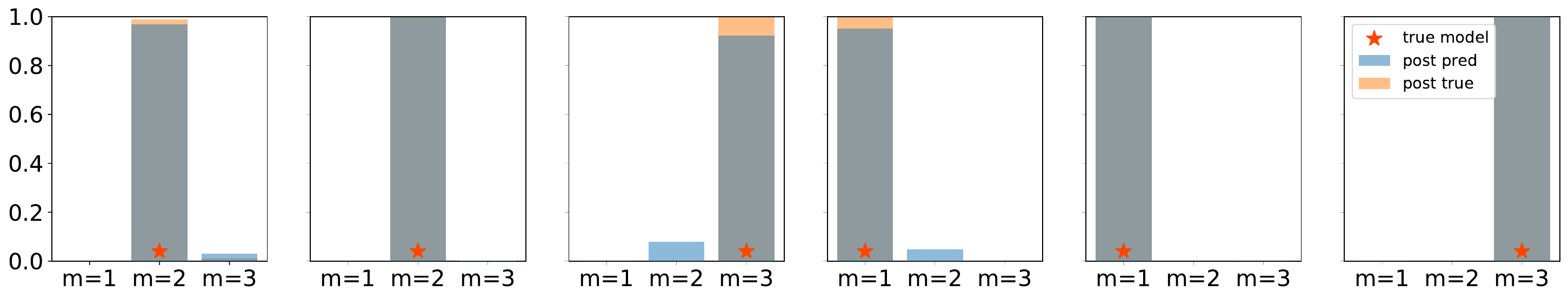}
  \caption{Case 1b. Examples of approximate posterior and true posterior for model indicator using policy resulting from `OED for model indicator', for $N=30$.}
  \label{fig:source_multi_model_post}
\end{figure}

\Cref{fig:source_multi_model_post} shows examples of approximate posterior and true posterior for model indicator using policy resulting from `OED for model indicator' with vsOED-G-I for $N=30$. The policy appears effective in discriminating the candidate models with highly concentrated posteriors, and the approximate model posteriors match well with the true model posteriors.
\Cref{tab:source_multi_model_kld} presents the PCE estimates of the EIG for model indicator and PoIs using policies resulting from the five OED scenarios, all using vsOED-G-I. 
The model discrimination OED finds the optimal policy in terms of maximizing the expected utility on model probability. 
Indeed, the `OED for model indicator' policy achieves the highest EIG estimate for model indicator, while model-PoIs and model-QoIs achieve higher values compared to their counterparts of `OED for PoIs' and `OED for QoIs'. Similarly, the policy from `OED for PoIs' achieves the highest EIG estimate on PoIs, with model-PoIs slightly lower.

\begin{table}[htbp]
    \centering
    \caption{Case 1b. PCE estimate of the EIG on model indicator and PoIs using policies resulting from the five OED scenarios. $\pm$ represents standard error. 
    }
    \begin{tabular}{lcc}
    \toprule
     & EIG on model indicator %
     & EIG on PoIs
     \\
    \midrule 
    OED for model indicator & $\mathbf{1.020}\pm 0.003$ 
    & $5.956\pm 0.065$
    \\
    OED for PoIs & $0.896 \pm0.005$ 
    & $\mathbf{10.567}\pm 0.048$
    \\
    OED for QoIs & $0.815\pm 0.005$ 
    & $6.999\pm 0.053$
    \\
    Model-PoIs OED & $0.950\pm 0.005$ 
    & $10.330\pm 0.049$
    \\
    Model-QoIs OED & $0.967\pm 0.004$ 
    & $7.830\pm 0.052$
    \\
    \bottomrule
    \end{tabular}
    \label{tab:source_multi_model_kld}
\end{table}

\paragraph{Hyperparameters and training stability} Hyperparameter settings and training stability results can be found in \cref{app:source}.

\subsection{Case 2: constant elasticity of substitution}
\label{sec:ces}

The CES case involves a single model and `OED for PoIs' only, and has been previously studied in \cite{Foster2019, foster2020unified, blau2022optimizing}. CES originates from behavioral economics, where participants are presented with two baskets of goods and asked to assess
the subjective difference in utility between the two baskets. The CES model \cite{arrow1961capital} represents the underlying utility function with latent PoIs $\Param=\{\rho,\beta,\log u\}$, where $\beta \in \mathbb{R}^3$ while $\rho$ and $u$ are scalars. We adopt independent priors $\rho \sim \text{Beta}\left(1,1\right), \beta \sim \text{Dirichlet}\left([1,1,1]\right), \log u \sim \CN\left(1, 3^2\right)$.
Note that the degree of freedom for $\beta$ is 2 since the sum of its components is constraint to $\beta_1+\beta_2+\beta_3=1$; hence, only $\beta_1$ and $\beta_2$ are included in $\Param$.
The design variables are $\design_k = [\design_{k,x}, \design_{k,x'}]$ and constrained to the design space $\design_{k,x} \in [0,100]^3$ and $\design_{k,x'} \in [0,100]^3$. The observation model is
\begin{align}
    Y_k = \text{clip}\bigg(\text{sigmoid}(\varphi_k), \epsilon, 1-\epsilon\bigg),
    \label{eq:CES_clip}
\end{align}
where the clipping threshold parameter is set to $\epsilon=2^{-22}$, and $\varphi_k \sim \CN\left(\mu_{\varphi}, \sigma_{\varphi}^2\right)$
with
\begin{align}
    \mu_{\varphi} = u \left[\left(\sum_{i=1}^3 (\design_{k,x,i})^\rho \beta_i\right)^{\frac{1}{\rho}} - \left(\sum_{j=1}^3 (\design_{k,x',j})^\rho \beta_j\right)^{\frac{1}{\rho}}\right], \qquad
    \sigma_{\varphi} = \tau u \bigg(1 + \norm{\design_{k,x}-\design_{k,x'}}{2}\bigg), \nonumber
\end{align}
and $\tau=0.005$. While the observation model in \cref{eq:CES_clip} differs from the form in \cref{eqn:observation}, the likelihood remains explicit and can be evaluated by applying a change of variable to the sigmoid function and integrating the tail probabilities beyond the clipped thresholds; we do not write the likelihood formula here for brevity, and refer interested readers to the associated vsOED code implementation on Github.

\begin{figure}[htbp]
  \centering
  \subfloat[Expected cumulative utility at different experiment stage $k$
for policies optimized for $N = 10$]{\label{fig:ces_EU_vs_stage}\includegraphics[width=0.49\linewidth]{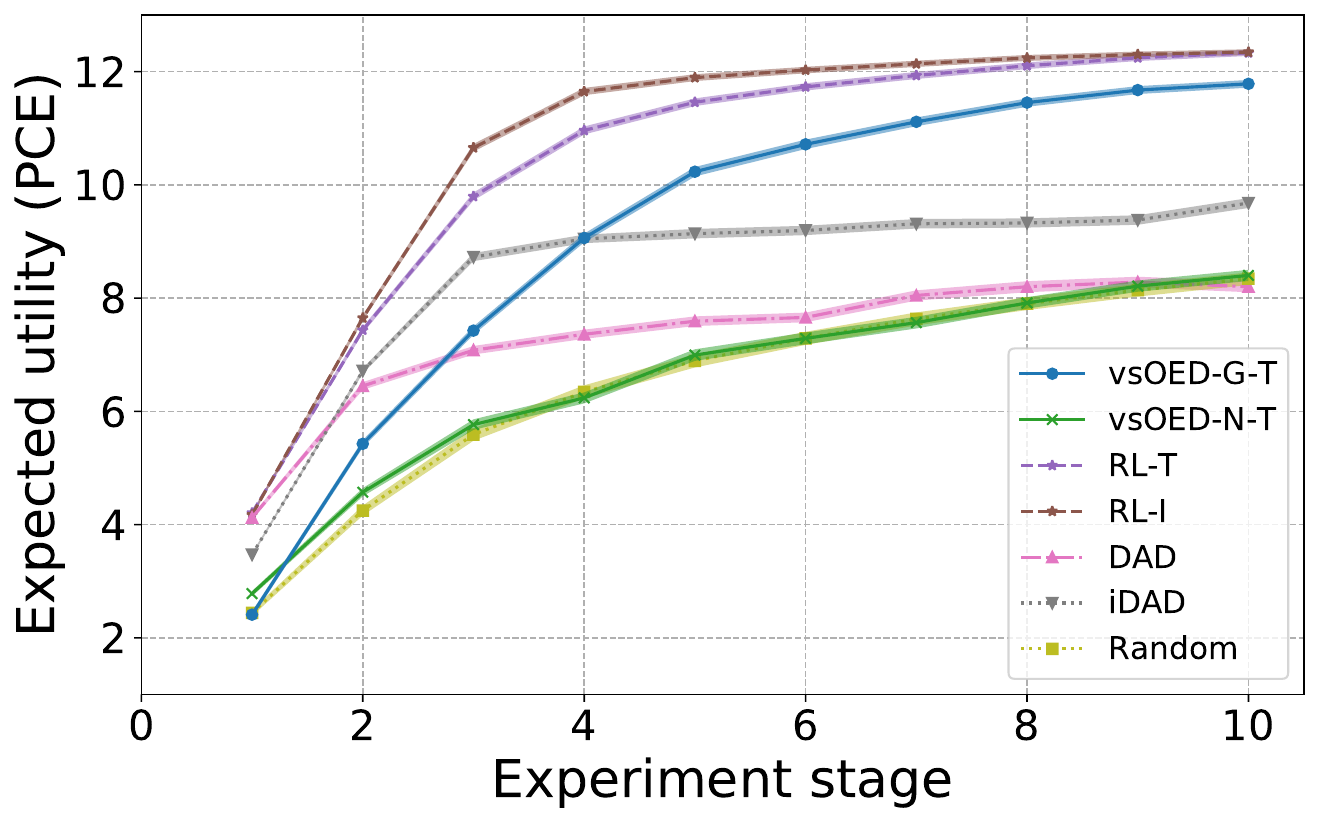}}
  \subfloat[Average expected utility over four training replicates versus design horizon $N$]{\label{fig:ces_EU_vs_horizon}\includegraphics[width=0.49\linewidth]{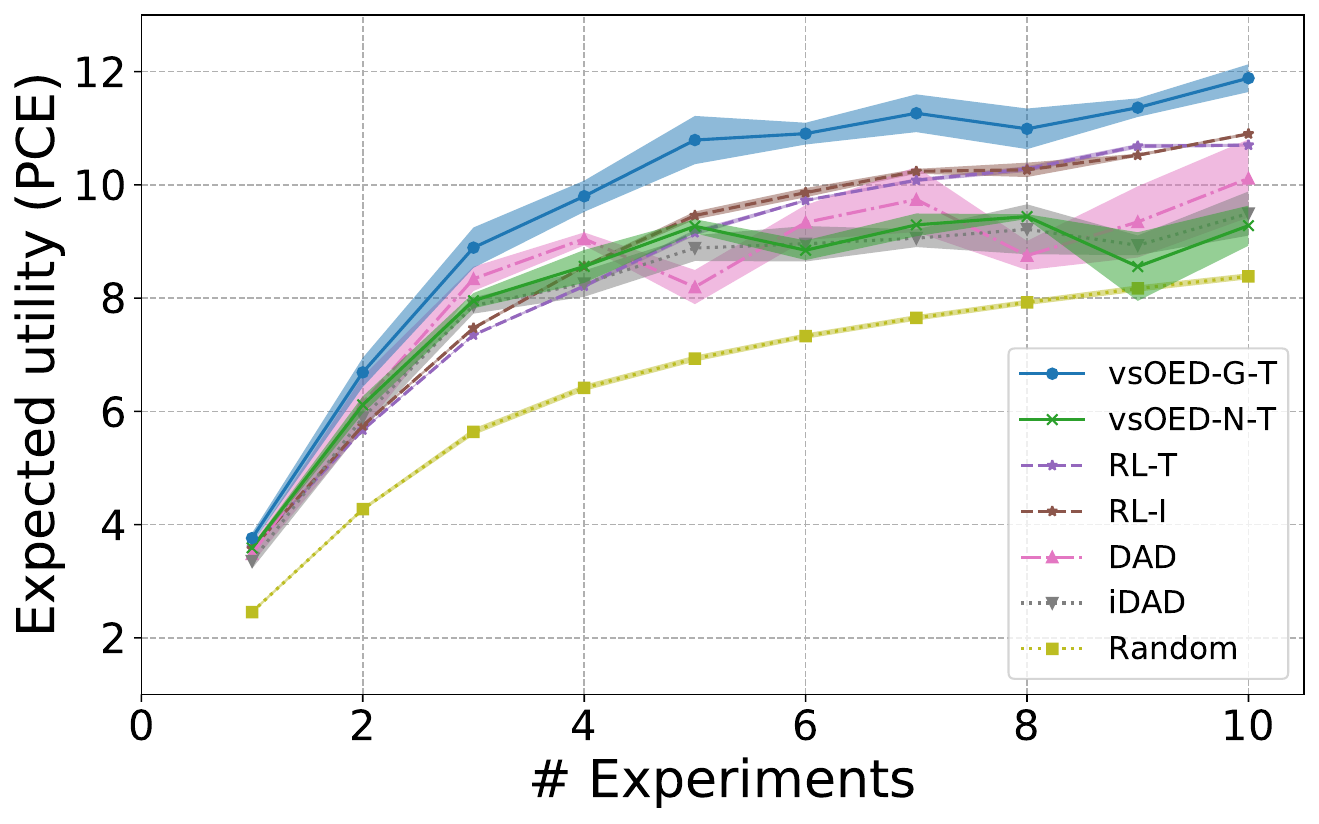}}
  \caption{Case 2. Expected utility comparisons using policies resulting from different algorithms. The shaded regions represent
    the standard error.  }
  \label{fig:ces_EU}
\end{figure}

\paragraph{Results} 
We use vsOED only with TIG since the horizons of this problem are relatively short, at most $N=10$, and \cref{fig:source_uni_poi_EU_vs_horizon} indicates that TIG and IIG perform similarly under those horizons. 
\Cref{fig:ces_EU_vs_stage} presents the expected cumulative utility, evaluated using PCE, at different experiment stages $k$ for policies optimized for a horizon of $N=10$ experiments under the fully trained setting. 
The plot suggests that vsOED with GMM and TIG is slightly inferior to fully-trained RL but significantly better than fully-trained DAD and iDAD. The lower values for vsOED with GMM and TIG in the earlier stages suggest that the policy prioritizes long-term expected utility accumulation over short-term rewards. 

\Cref{fig:ces_EU_vs_horizon} presents the average expected utility, averaged over four training replicates where each replicate is again evaluated using PCE, for policies trained under different design horizons $N$ under the limited budget setting. 
Here vsOED with GMM and TIG outperforms other approaches for all values of $N$. 
The standard errors are plotted as shaded regions and illustrate a greater robustness of vsOED-G-T and RL over vsOED-N-T and DAD.

For this case, vsOED with NFs appears significantly worse compared to vsOED with GMM. This is like due to NFs having more difficulty in representing distributions of random variables with compact support, $\rho$ and $\beta$ in this case.

\begin{figure}[htbp]
  \centering
  \label{fig:ces_post_h10}
  \includegraphics[width=0.8\linewidth]{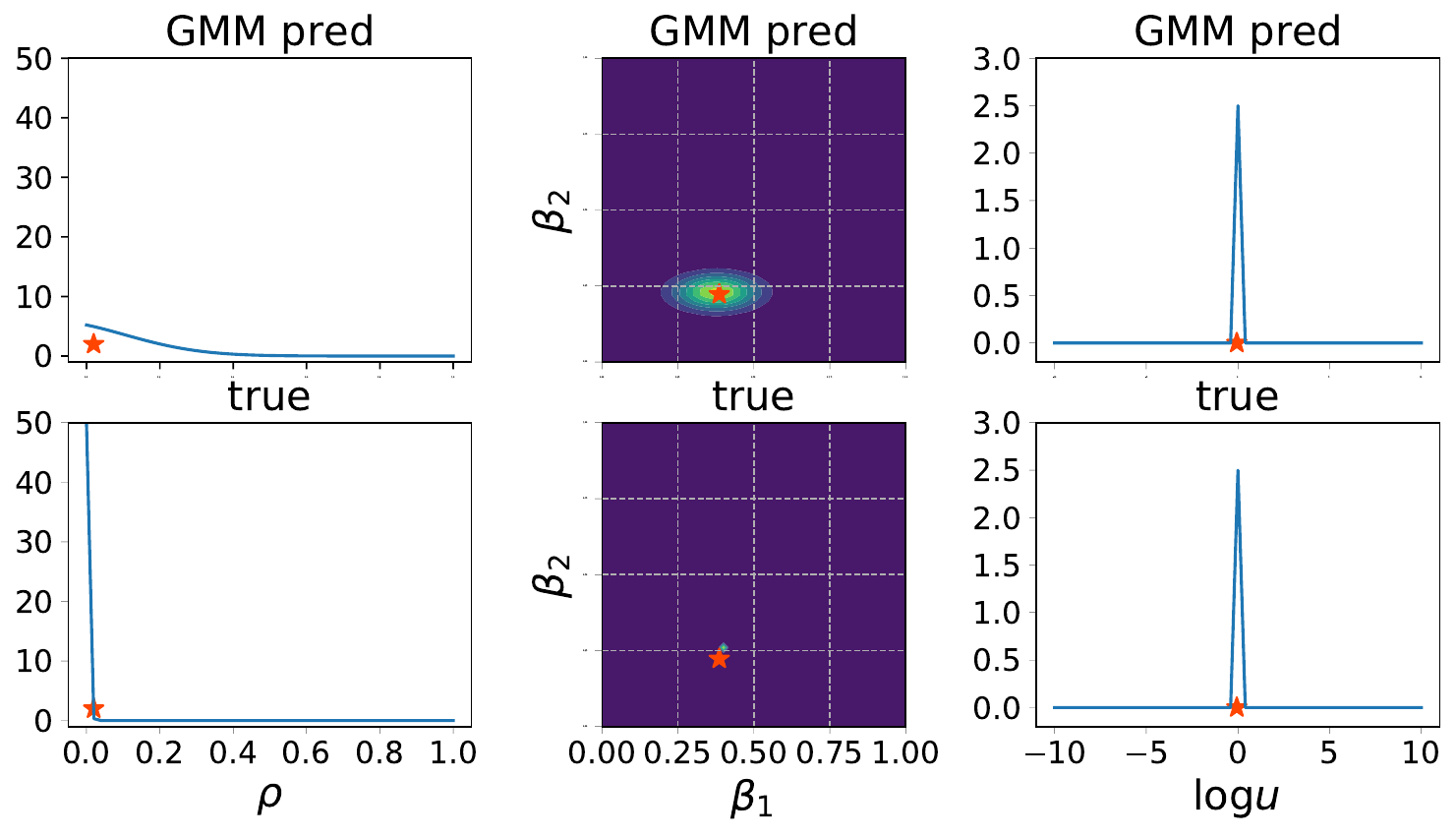}
  \caption{Case 2. Examples of GMM approximate posterior and true posterior at $N=10$. The red stars
indicate the true data-generating parameter values for these trajectory samples.
}
  \label{fig:ces_post}
\end{figure}

\Cref{fig:ces_post} provides examples comparing the GMM approximate posterior and the true posterior. GMM performs well for $\log{u}$, but tends to produce wider posteriors for $\rho$ and $\beta$. This is likely due to that many observations are clipped at the two ends per \cref{eq:CES_clip} which leads to numerous observations with identical values, and the non-uniqueness makes it difficult for GMM to learn the mapping from designs and observations to the posterior. Despite this challenge, vsOED with GMM is still able to find good policies.

\paragraph{Hyperparameters and training stability} Hyperparameter settings and training stability results can be found in \cref{app:ces_hyperparam}.

\subsection{Case 3: SIR model for disease spread}
\label{sec:sir}
The SIR case involves a single model and `OED for PoIs' only, and has been previously studied in~\cite{ivanova2021implicit}. As the model involves solving a stochastic differential equation (SDE), the case tests the ability of vsOED to handle an expensive forward model and implicit likelihood. 
SIR is a stochastic model \cite{kleinegesse2021gradient, cook2008} describing the spread of infectious diseases in a population. Individuals in the population are divided into three categories: susceptible, infected, and recovered. 
The number of individuals in these categories at any given time $t$ are respectively represented by $S(t)$, $I(t)$, and $R(t)$, which always sum to a fixed population size $S(t) + I(t) + R(t) = N_p, \forall t$. Thus, the population
state vector is defined to be ${X}(t) = [S(t), I(t)]$, where $R(t)$ is implied by the population constraint and can be ignored. 

An individual at time $t$ in the susceptible category
has a probability to become infected 
controlled  by rate parameter $\beta$,
an individual who is infected has a probability to recover 
controlled by rate parameter $\rho$, and an individual who is already recovered remains recovered.
The PoIs $\Theta=\{\log \beta,\log \rho\}$ entail the logarithm of the rates to ensure their non-negativity. We adopt independent priors 
$\log{\beta} \sim \CN\left(\log{0.5}, 0.5^2\right), \log{\rho} \sim \CN\left(\log{0.1}, 0.5^2\right)$.
The design variable for the $k$th experiment is $\design_k \in [0,100]$,
which is the time chosen for observing the number of infected $I(t=\design_k)$. The designs are also constrained by $\design_k \leq \design_{k+1}$. 
The overall SIR model is often defined by a continuous-time Markov chain and can be sampled via the Gillespie
algorithm \cite{allen2017}, but it generally yields discrete population states
that have undefined gradients. To circumvent these limitations, we follow \cite{ivanova2021implicit} and adopt a continuous-state alternative (i.e., where $X(t)$ can be real-valued) form based on an SDE formulation. 

In this SIR model formulation, the state dynamics follows the It\^{o} SDE:
\begin{align}
    \text{d}{X}(t) = {f}({X}(t))\,\text{d}t + {D}({X}(t) )\,\text{d}{W}(t),
    \label{eq:sir_sde}
\end{align}
where ${W}(t)$ is a vector of independent Wiener processes (i.e., Brownian motion), and ${f}$ and ${D}$ are the state-dependent drift vector and diffusion matrix, respectively, defined as~\cite{kleinegesse2021gradient}
\begin{align}
    {f}({X}(t)) = \begin{bmatrix}
-\beta\frac{S(t)I(t)}{N_p} \\
\beta\frac{S(t)I(t)}{N_p} - \rho I(t)
\end{bmatrix}, \qquad {D}({X}(t) ) = \begin{bmatrix}
-\sqrt{\beta\frac{S(t)I(t)}{N_p}} & 0 \\
\sqrt{\beta\frac{S(t)I(t)}{N_p} } & -\sqrt{\rho I(t)}
\end{bmatrix}.
\label{eq:drift_diffusion}
\end{align}
Given \cref{eq:sir_sde,eq:drift_diffusion}, we can simulate state ${X}(t)$ by solving the SDE via finite-differencing, such as the Euler--Maruyama method; here we directly adopt the solver from~\cite{ivanova2021implicit}. For a fair comparison, we follow \cite{ivanova2021implicit} and use the solutions of \cref{eq:sir_sde} as data and do not consider an additional Poisson observational model that increases the noise in simulated data as suggested in \cite{kleinegesse2021gradient}---that is, $Y_k=I(\xi_k)$. The likelihood is implicit for this case, since computing the probability of all stochastic transitions leading to the observed value would be intractable. Hence, we can sample from the observation model, but not evaluate its probability density. 

Solving the SDE is quite computationally expensive, we thus limit the computational budget to 1 million trajectory samples for both vsOED and iDAD. To accelerate the training process, we pre-generate and store 1 million simulations and the access the stored simulations during the training. A new set of $3\times 10^{5}$ simulations are 
used as evaluation data. 

\begin{figure}[htbp]
  \centering
  \includegraphics[width=0.55\linewidth]{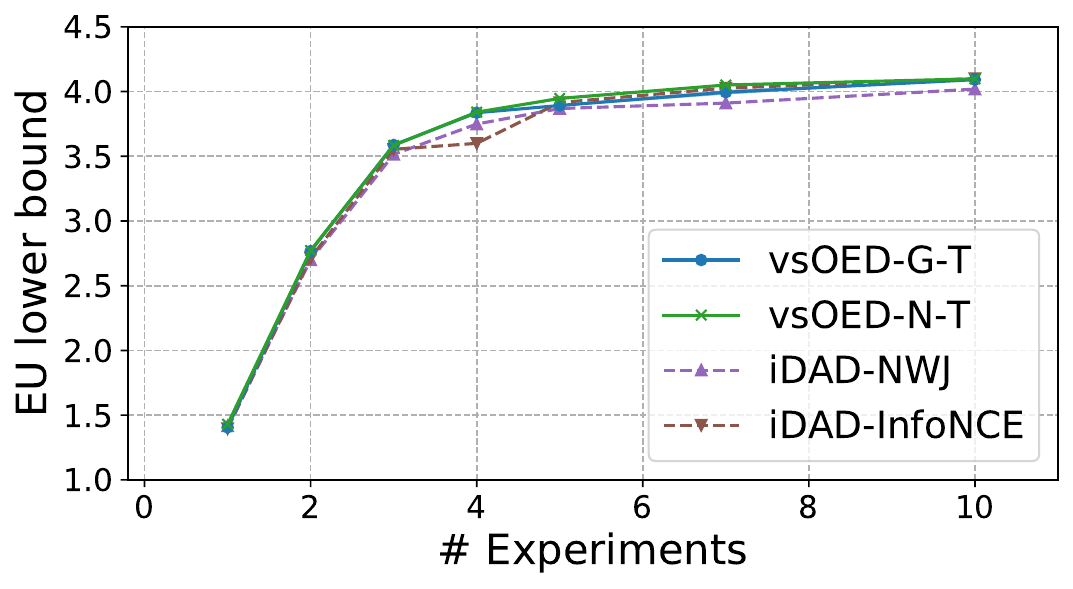}
  \caption{Case 3. Average $\tilde{U}$ over four training replicates versus design horizon $N$. The shaded regions represent the standard error.}
  \label{fig:sir_EU_vs_horizon}
\end{figure}

\paragraph{Results}
Only vsOED and iDAD are used in this case since the other algorithms cannot handle implicit likelihood. We use vsOED only with TIG since the horizons of this problem are relatively short, at most $N=10$, and \cref{fig:source_uni_poi_EU_vs_horizon} indicates that TIG and IIG perform similarly under those horizons. 
\Cref{fig:sir_EU_vs_horizon} presents the average $\tilde{U}$, averaged over four training replicates, for policies trained under different design horizons $N$. vsOED and iDAD appear to perform similarly, with vsOED slightly better at some spots. However, we note that the comparison is not entirely commensurable since iDAD is trained based on a different EIG lower bound estimator than $\tilde{U}$ (which vsOED uses and is also used here for evaluation), and that iDAD additionally uses the forward model derivatives. The absence of requiring model derivatives in vsOED can be potentially valuable in situations where model derivatives are inaccessible.

\begin{figure}[htbp]
  \centering
  \subfloat[$I(t)$]{\label{fig:sir_data}\includegraphics[width=0.49\linewidth]{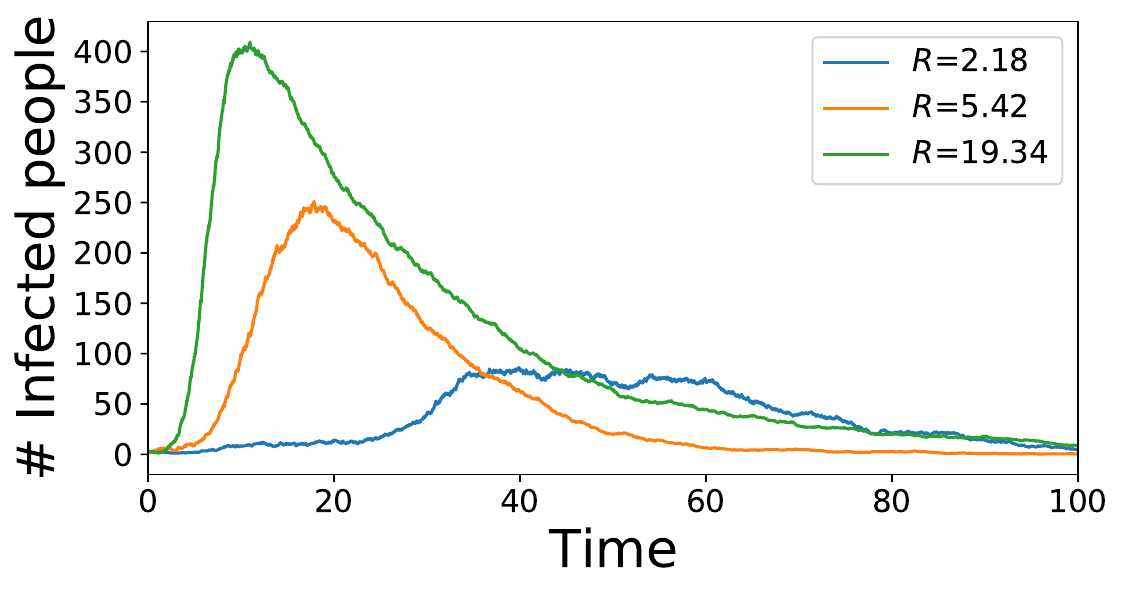}}
  \subfloat[$\design_k$]{\label{fig:sir_policy}\includegraphics[width=0.49\linewidth]{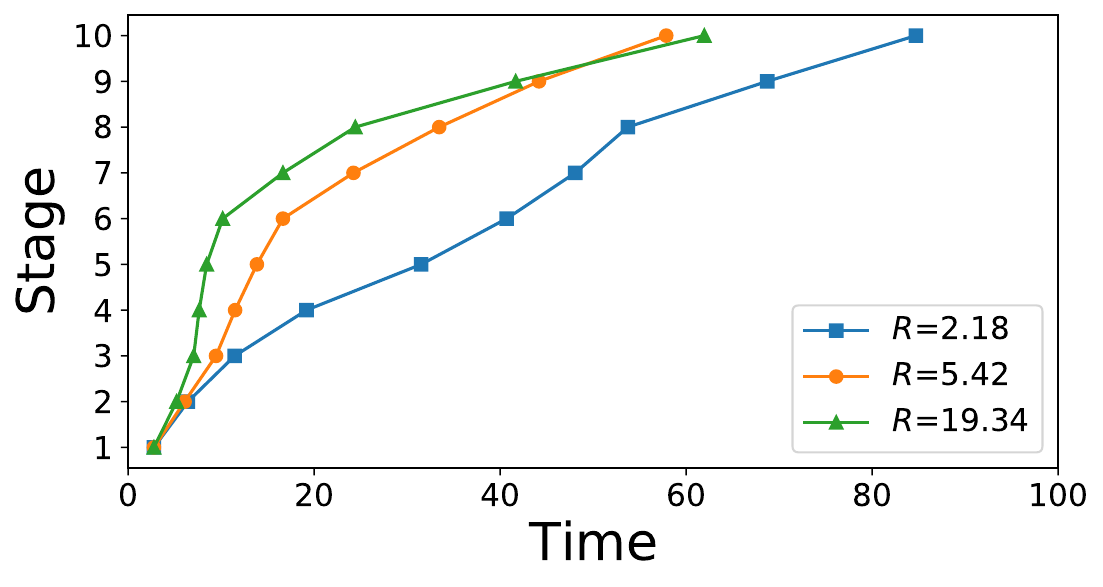}}
  \caption{Case 3. Examples of infected state trajectory $I(t)$ and corresponding designs (observation times) $\design_k$ for $N=10$ with three realizations of $(\beta, \rho)$ and different ratios $r=\beta/\rho$.
  }
  \label{fig:sir_data_policy}
\end{figure}

\Cref{fig:sir_data_policy} presents examples of infected state  trajectory $I(t)$ and corresponding designs (observation times) $\design_k$
for $N=10$
with three realizations of $(\beta, \rho)$ and different ratios $r=\beta/\rho$. We observe that smaller $r$ corresponds to a more spread out design of observation times, which aligns with the results in~\cite{ivanova2021implicit}.

\paragraph{Hyperparameters and training stability} Hyperparameter settings and training stability results can be found in \cref{app:sir_hyperparam}.

\subsection{Case 4: convection-diffusion-reaction}
\label{sec:conv_diff}

The convection-diffusion-reaction case involves nuisance parameters and physical state, and has been previously studied in~\cite{shen2021bayesian} under the single-model setting.
The case entails designing mobile sensor movements within a contaminant plume whose dynamics are governed by a 2D convection-diffusion-reaction partial differential equation (PDE).
In this paper, we consider three candidate models with uniform model prior $P(m)=1/3$ for $m=1,2,3$. The $m$th model contains $m$ Gaussian-profiled contaminant sources randomly located in a 2D domain. 
The PoIs are the source locations $\Param_m = \{\Param_{m,i}\}_{i=1}^{m}$ where $\Param_{m,i}=[\Param_{m,i,x}, \Param_{m,i,y}]\in \RR^2$ denotes the location of the $i$th source and is endowed with independent priors $\Param_{m,i,x},\Param_{m,i,y} \sim \CU(0,1^2)$. 
For the $m$th model, the contaminant concentration $G$ at location $[x,y]$ and time $t$ is governed by
\begin{align}
    \frac{\partial G(\Param_m,\Nuis_m,x,y,t;m)}{\partial t} = \nabla^2_{(x,y)} G - u(\Nuis_m) \cdot \nabla_{(x,y)} G + S(\Param_m, x,y,t;m),
    \nonumber
\end{align}
for $[x,y]\in [-1, 2]^2$ and $0 < t \le 0.2$.
Here $u=[v\cos{\beta},v\sin{\beta}]^{\top} \in \RR^2$ is a time-invariant convection velocity that is described by nuisance parameters $\Nuis_m = \{v, \beta\}$ with independent priors on the convection speed (magnitude) $v \sim \CU(0, 20)$ and convection angle $\beta \sim \CU(0, 2\pi)$. The source function is
\begin{align}
    S(\Param_m,x,y,t;m) = \sum_{i=1}^m \frac{s}{2\pi h^2} \exp\( -\frac{\norm{\Param_{m,i} - [x,y]}{2}^2}{2 h^2} \), \nonumber
\end{align}
where $s=2$ is the known source strength and $h=0.05$ is the known source width.
The initial condition is $G(\Param_m,\Nuis_m,x,y,0;m)=0$ and homogeneous Neumann boundary conditions are applied to all sides of the computational domain. 
The design variables $\design_k=[\design_{k,x},\design_{k,y}] \in [0,1]^2$ entail selecting the location of concentration measurement within the allowable design space at prescribed time intervals $t_k=0.01 (k+1)$. 
The observation model is then $Y_k = G(\Param_m,\Nuis_m,\design_{k,x},\design_{k,y},t_k;m)+\mathcal{E}_k$ where $\mathcal{E}_k\sim\CN(0,\sigma_{\epsilon}^2)$ with $\sigma_{\epsilon}=0.05$. 
Moreover, the measurement sensor is initially located at $[x_0,y_0] = [0.5,0.5]$, and a sensor movement penalty of $-0.1\norm{\design_k-\design_{k-1}}{2}$ for $k=1,\ldots,N-1$ (and $-0.1\norm{\design_k-[x_0,y_0]}{2}$ for $k=0$) is incurred in the immediate rewards to reflect the cost of moving the sensor. 
 
Similar to \cref{sec:source}, we are interested in a QoI that is the log of the integrated flux magnitude crossing the right boundary $x=1$ of the design space (i.e., spanning from $y=0$ to $y=1$) at a future time
$t=0.2$. The QoI is $Z_m=\log{(\abs{\varphi(\Param_m,\Nuis_m; m)})}$, where
\begin{align}
    \varphi(\Param_m,\Nuis_m;m) = \int_{0}^{1} - \frac{\partial G(\Param_m, \Nuis_m, x=1, y, t=0.2; m)}{\partial x} \, \text{d}y. \nonumber
\end{align}

We apply a second-order finite volume method to solve the PDE numerically. The computational domain $[-1,2]^2$ is discretized into a uniform grid with cell size $\Delta x = \Delta y = 0.01$. Second-order fractional step method is used for time integration, with a step size $\Delta t = 5.0 \times 10^{-4}$. The integrated flux $\varphi$ is computed by first estimating the derivative term using finite difference on the grid values at the target boundary, and then numerically integrated using midpoint rule. 

While we can directly use the finite volume solver as the forward model, it can be expensive and inefficient since each solve calculates the concentrations for all $[x,y]$ and all $t$, while in the OED problem only a small subset of these values are needed for each forward model evaluation. 
Therefore, to accelerate computations, we pre-build NN surrogate models to replace $G(\Param_m,\Nuis_m,\xi,t_k;m)$ and $\varphi(\Param_m,\Nuis_m; m)$, one for each $m$ and $t_k$. The architectures of these NN surrogates can be found in %
\cref{app:convec_diff_hyperparam}.

\begin{figure}[htbp]
  \centering
  \subfloat[$m=1$]{\label{fig:conv_diff_surrogate_comparison_1}\includegraphics[width=0.48\linewidth]{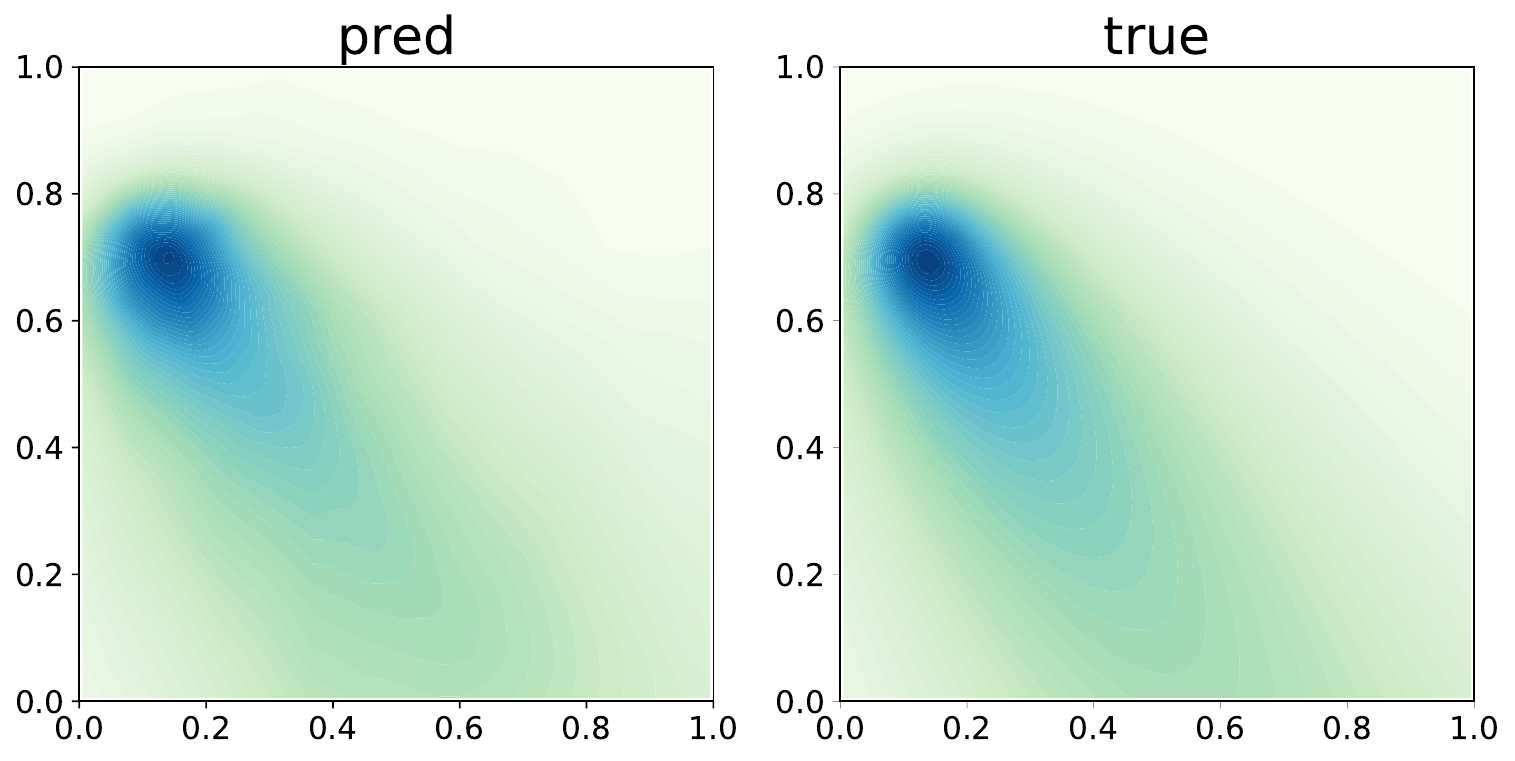}}
  \hspace{1em}
  \subfloat[$m=3$]{\label{fig:conv_diff_surrogate_comparison_3}\includegraphics[width=0.48\linewidth]{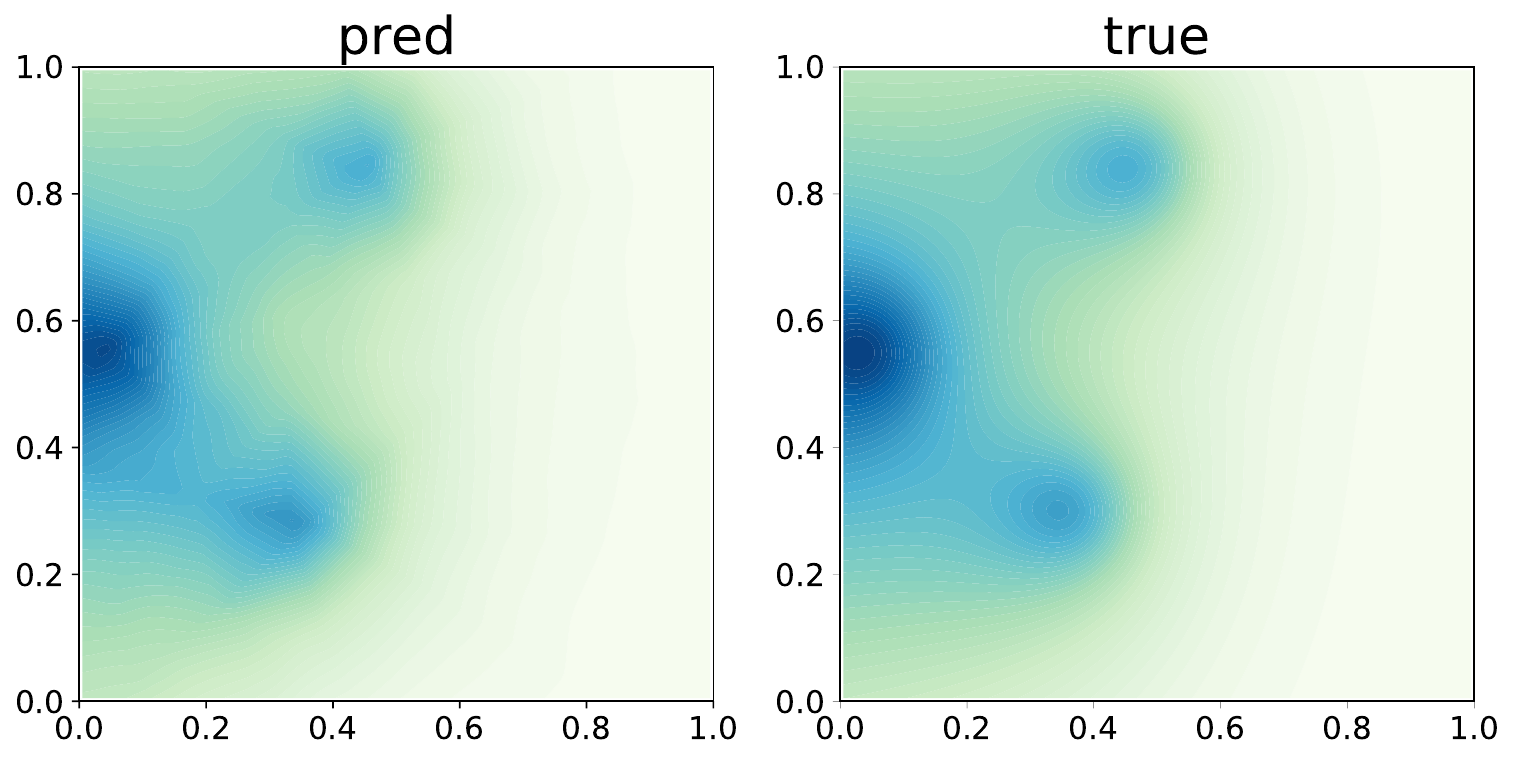}}
  \caption{Case 4. Examples of the true and surrogate concentration fields $G$ at $t=0.2$. }
  \label{fig:conv_diff_surrogate_comparison}
\end{figure}

A separate $G$ surrogate is built for each $m$ and $t_k$, and a separate $\varphi$ surrogate is built for each $m$. 
For each $m$, 20,000 simulations are generated with random samples of $\Param_m$ and $\Nuis_m$, with 18,000 used for training and 2,000 for testing. The testing mean squared errors (MSE) are shown in \cref{tab:conv_diff_surrogate_error} for $G$ surrogates at the end time $t=0.2$ and the three $\varphi$ surrogates, illustrating excellent accuracy. \Cref{fig:conv_diff_surrogate_comparison} presents examples of the true and surrogate concentration fields $G$ at the end time $t=0.2$; they demonstrate excellent agreement.

\begin{table}[htbp]
    \centering
    \caption{Case 4. Testing MSE of the surrogate models.}
    \begin{tabular}{ccc}
    \toprule
    Model & Surrogate $G$ at $t=0.2$ & Surrogate $\varphi$ \\
    \midrule 
    $m=1$ & $3.094 \times 10^{-5}$ & $4.141 \times 10^{-5}$ \\
    $m=2$ & $3.284 \times 10^{-4}$ & $4.986 \times 10^{-4}$ \\
    $m=3$ & $1.650 \times 10^{-3}$ & $2.080 \times 10^{-3}$ \\
    \bottomrule
    \end{tabular}
    \label{tab:conv_diff_surrogate_error}
\end{table}

\paragraph{Results}
Only vsOED is used in this case since the other algorithms cannot handle nuisance parameters and multiple models. We use vsOED only with TIG since the horizons of this problem are relatively short, at most $N=15$, and \cref{fig:source_uni_poi_EU_vs_horizon} indicates that TIG and IIG perform similarly under those horizons. Only the GMM version of vsOED is presented for brevity. 
Similar to Case 1b, we consider five OED scenarios: OED for model indicator (i.e., design for model discrimination, $\alpha_{M}=1$, $\alpha_\Param=\alpha_Z=0$), OED for PoIs (i.e., design for parameter inference, $\alpha_\Param=1$, $\alpha_{M}=\alpha_Z=0$), OED for QoIs (i.e., design for goal-oriented prediction, $\alpha_Z=1$, $\alpha_{M}=\alpha_\Param=0$), OED for both model indicator and PoIs (`model-PoIs', $\alpha_{M}=\alpha_\Param=1$, $\alpha_Z=0$), and OED for both model indicator and QoIs (`model-QoIs', $\alpha_{M}=\alpha_Z=1$, $\alpha_\Param=0$). 

\begin{figure}[htbp]
  \centering
  \subfloat[Posteriors for model indicator from `OED for model indicator'.]{\label{fig:conv_diff_model_post}\includegraphics[width=0.9\linewidth]{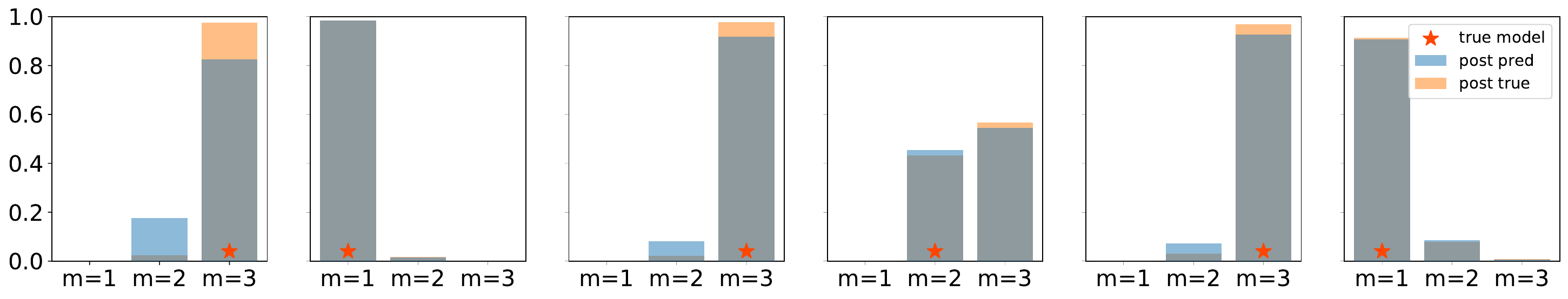}}
  \\
  \subfloat[Posteriors for PoIs from `OED for PoIs' for $m=1$, along with the true posterior for the nuisance parameters.]{\label{fig:conv_diff_poi_post}\includegraphics[width=0.9\linewidth]{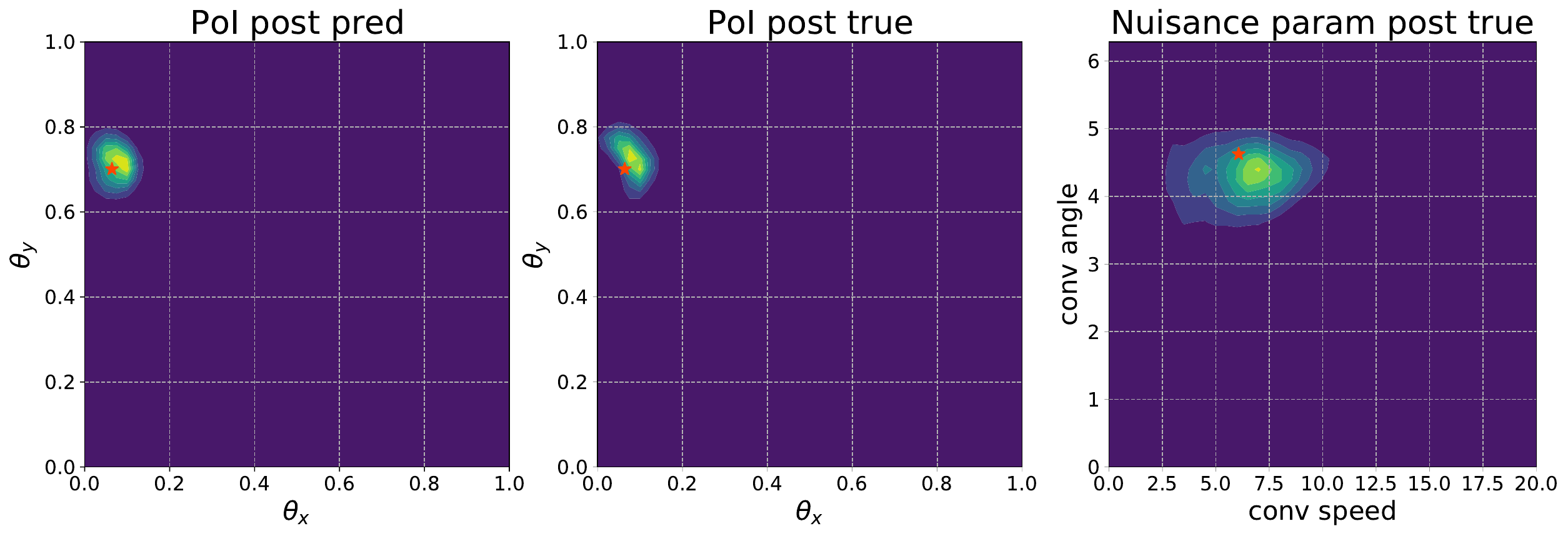}}
  \caption{Case 4. Examples of approximate posterior and true posterior for model indicator and PoIs using policy resulting from `OED for model indicator' and `OED for PoIs', respectively, for $N=10$.  }
  \label{fig:conv_diff_post}
\end{figure}

\begin{figure}[htbp]
  \centering
  \subfloat[OED for model indicator]{\label{fig:conv_diff_model_policy}\includegraphics[width=1\linewidth]{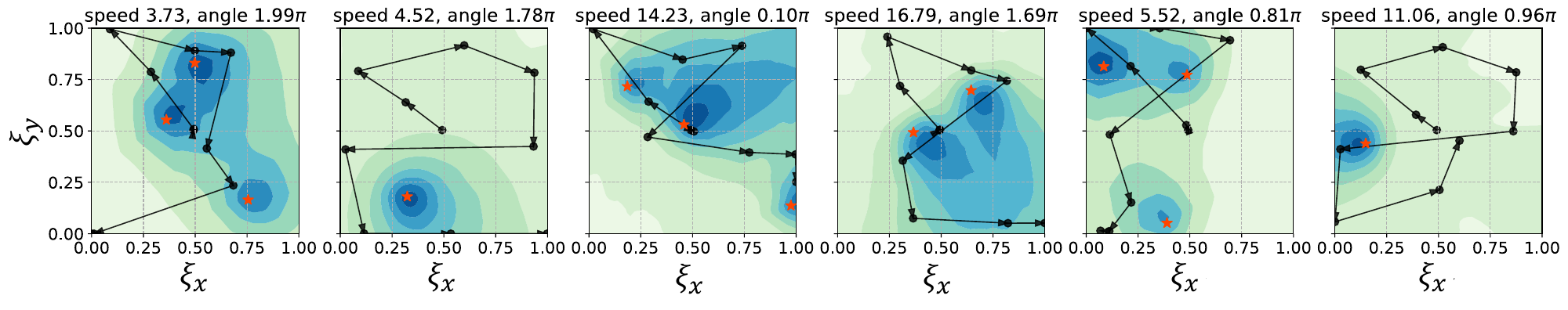}}
  \\
  \subfloat[OED for PoIs]{\label{fig:conv_diff_poi_policy}\includegraphics[width=1\linewidth]{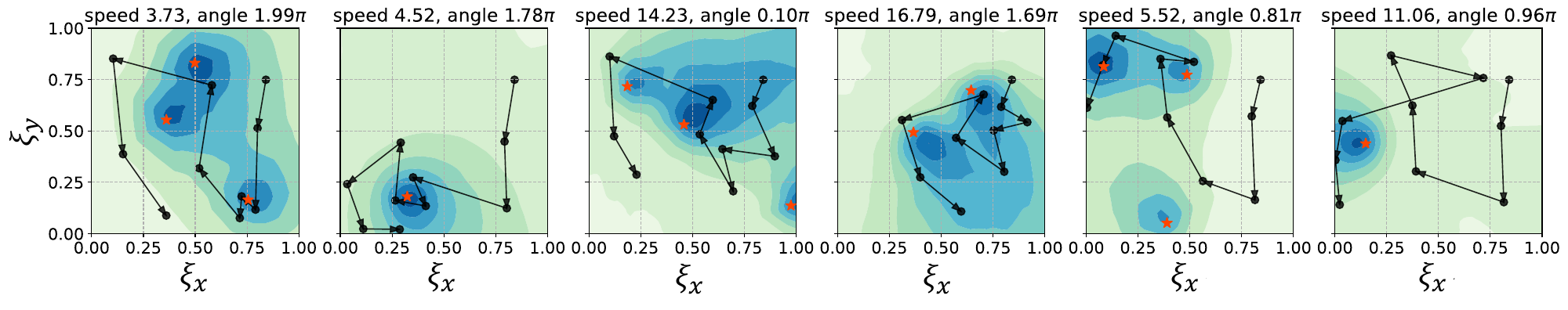}}
  \\
  \subfloat[OED for QoIs]{\label{fig:conv_diff_goal_policy}\includegraphics[width=1\linewidth]{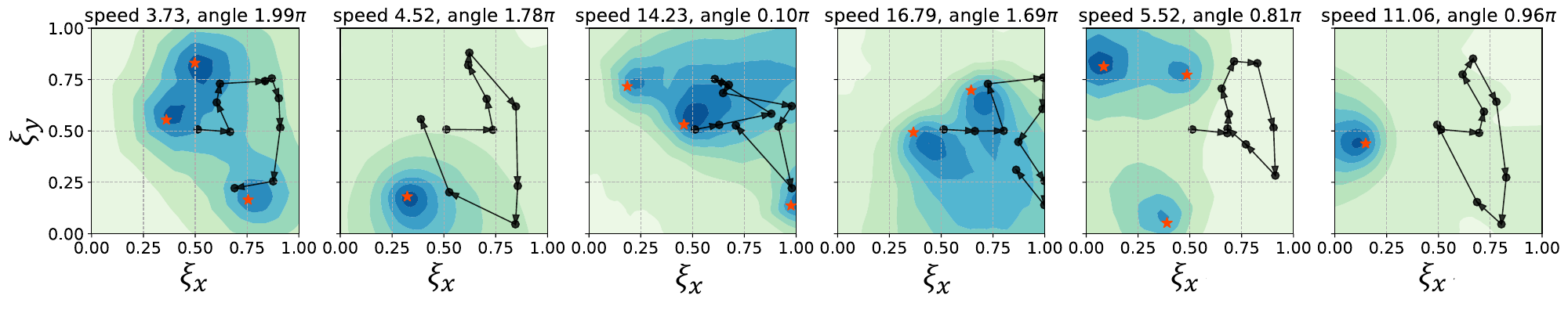}}
  \\
  \subfloat[Model-PoIs OED]{\label{fig:conv_diff_model_poi_policy}\includegraphics[width=1\linewidth]{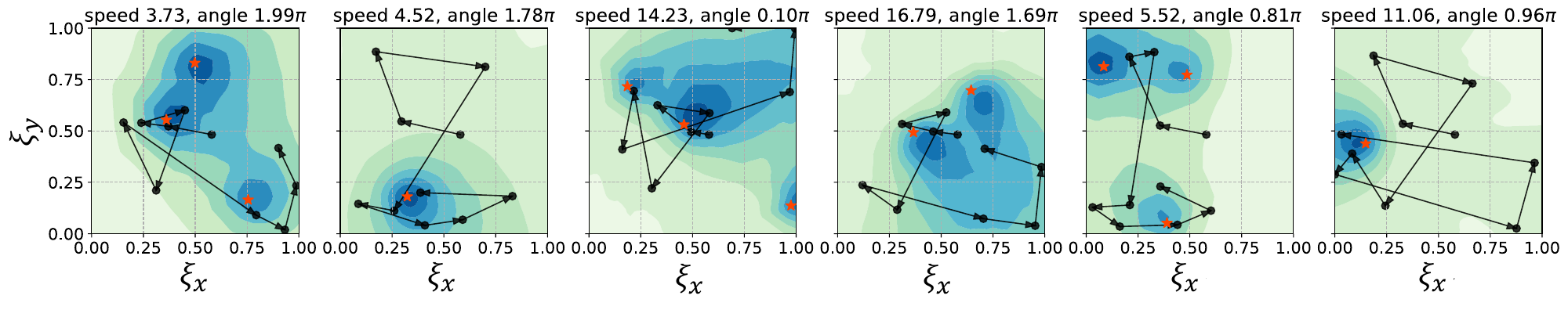}}
  \\
  \subfloat[Model-QoIs OED]{\label{fig:conv_diff_model_goal_policy}\includegraphics[width=1\linewidth]{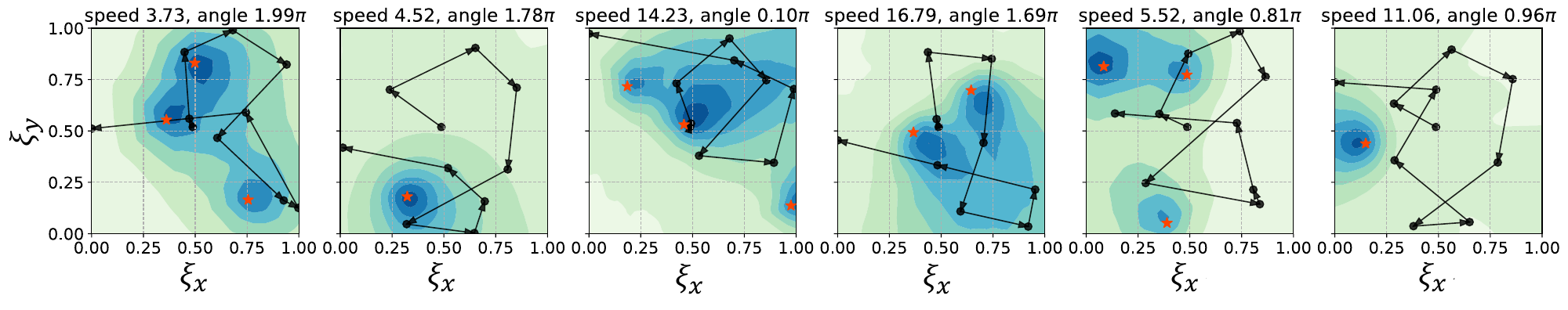}}
  \caption{Case 4. Examples of policy trajectory for $N=10$. The background contour plots the true contaminant concentration at the end time $t=0.2$, and the red stars indicate the true source locations.
  }
  \label{fig:conv_diff_policy}
\end{figure}

\Cref{fig:conv_diff_post} presents examples comparing the GMM approximate posterior and true posterior for model indicator and PoIs (from $m=1$) using policies respectively resulting from `OED for model indicator' and `OED for PoIs' for $N=10$; the GMMs again approximate the posteriors well for this case. 
Notably, an example PoI posterior and the corresponding nuisance parameter posterior are shown in \cref{fig:conv_diff_poi_post}, both plotted over the full support of their respective uniform priors. We see that the PoI posterior has shrunk more substantially than that for the nuisance parameter, which is consistent with the expected behavior of `OED for PoIs' that targets to reduce the PoI uncertainty. 
\Cref{fig:conv_diff_policy} presents examples of trajectory using policies from the five OED scenarios for $N=10$. 
The overall policy behavior is similar to the multi-model source location finding problem of Case 1b.
Trajectories from `OED for model indicator' tend to explore more and appear more diffuse compared to the other scenarios, often extending to the boundaries of the design space and leading to narrow posteriors of model indicator as shown in \cref{fig:conv_diff_model_post}. 
Trajectories from `OED for PoIs' tend to be more exploitative, remaining closer to the estimated sources while leveraging the background convection. Trajectories from `OED for QoIs' exhibit a vertical design tendency, due to the same effects as explained in Case 1b.

\paragraph{Hyperparameters, training stability, and surrogate architectures} Hyperparameter settings, training stability, and surrogate architectures can be found in \cref{app:convec_diff_hyperparam}. 

\section{Conclusions}
\label{sec:conclusions}

We introduced vsOED, a novel method for solving Bayesian sequential OED problem using an actor-critic reinforcement learning framework powered by
policy gradient techniques and the variational Barber--Agakov bound to the EIG.
vsOED is capable of accommodating nuisance parameters, implicit likelihoods, and multiple candidate models, while supporting a flexible design criterion that can target designs for model discrimination, parameter inference, goal-oriented  prediction, or their weighted combinations.

We provided key theoretical results including theorems proving the equivalence between incremental and terminal information gain rewards, equivalence between EIG and its one-point reward formulations, and the lower bound property of vsOED when posteriors are approximated. 
We then developed numerical methods under an actor-critic framework, deriving and estimating the policy gradient and utilizing Gaussian mixture models and normalizing flows to approximate the posteriors. 

Finally, we demonstrated vsOED across four numerical cases---source location finding, constant elasticity of substitution, SIR model for disease spread, and convection-diffusion-reaction. These scenarios involved challenges of multiple models, implicit likelihoods, and nuisance parameters. 
Comparisons with existing algorithms (DAD, iDAD, and RL) showed that vsOED achieves superior sample efficiency under limited budgets and avoids reliance on model derivatives.

Despite its strengths, vsOED has limitations. 
Its performance is sensitivity to inaccurate posterior approximations, especially for posteriors with compact support, multiple modes, and highly non-Gaussian features. 
Developing more accurate and adaptive posterior representations, particularly in high-dimensional spaces, will be highly valuable. 
Additionally, vsOED currently does not handle discrete designs or stochastic policies, which could expand its applicability, as demonstrated by~\cite{blau2022optimizing}. 
Further enhancements could also stem from advanced reinforcement learning techniques, such as proximal policy optimization, trust region policy optimization, and soft actor-critic~\cite{schulman2017proximal, schulman2015trust, haarnoja2018soft, fujimoto2018addressing}.

\section*{Acknowledgments}

This research is based upon work supported in part by the U.S. Department of Energy, Office of Science, Office of Advanced Scientific Computing Research, under Award Numbers DE-SC0021397 and DE-SC0021398.  
This work relates to Department of Navy award N00014-23-1-2735 issued by
the Office of Naval Research.
This research is supported in part through computational resources and services provided by Advanced Research Computing at the University of Michigan, Ann Arbor.

\bibliography{references}
\bibliographystyle{elsarticle-num} 

\clearpage
\appendix
\section{Proofs} 
\label{app:proofs}

\subsection{Information gain jointly with model indicator} 
\label{app:total_info_gain}

Akin to the total entropy in 
\cite{borth1975total}, 
the IG jointly on the model indicator and PoIs is:
\begin{align}
    & \DKL\(\, P_{M, \Param_m|\info_{k_2}}\,||\,P_{M, \Param_m|\info_{k_1}} \,\) \nonumber \\
    &= \sum_{m=1}^{|\CM_m|} \int p(m,\param_m|\info_{k_2}) \log \frac{p(m,\param_m|\info_{k_2})}{p(m,\param_m|\info_{k_1})} \,\text{d}\param_m \nonumber\\
    &= \sum_{m=1}^{|\CM_m|} P(m|\info_{k_2}) \int p(\param_m|m,\info_{k_2}) \log \frac{P(m|\info_{k_2})\,p(\param_m|m,\info_{k_2})}{P(m|\info_{k_1})\,p(\param_m|m,\info_{k_1})} \,\text{d}\param_m \nonumber \\
    &= \sum_{m=1}^{|\CM_m|} P(m|\info_{k_2}) \log \frac{P(m|\info_{k_2})}{P(m|\info_{k_1})} + \sum_{m=1}^{|\CM_m|} P(m|\info_{k_2}) \int p(\param_m|m,\info_{k_2}) \log \frac{p(\param_m|m,\info_{k_2})}{p(\param_m|m, \info_{k_1})} \,\text{d}\param_m \nonumber \\
    &= \DKL\(\, P_{M|\info_{k_2}}\,||\,P_{M|\info_{k_1}} \,\) + \EE_{M|\info_{k_2}} \[ \DKL\(\, p_{\Param_m|\info_{k_2}}\,||\,p_{\Param_m|\info_{k_1}} \,\) \], \nonumber 
\end{align}
where $0\leq k_1 \leq k_2 \leq N$, and $p(\param_m|m, \info_{k})=p(\param_m|\info_{k})$ per our notation convention. 
When setting $k_1=0$ and $k_2=N$, we recover the terminal reward in \cref{eq:terminal_info_gN} under the special case of $\alpha_M=\alpha_\Param=1$ and $\alpha_Z=0$.
Similarly, the IG jointly on the model indicator and QoIs is:
\begin{align}
    &\DKL \( p_{M,Z_m|\info_{k_2}} \,||\, p_{M,Z_m|\info_{k_1}}\) \nonumber \\
    &= \sum_{m=1}^{|\CM_m|} \int p(m,z_m|\info_{k_2}) \log \frac{p(m,z_m|\info_{k_2})}{p(m,z_m|\info_{k_1})} \,\text{d}z_m \nonumber\\
    &= \sum_{m=1}^{|\CM_m|} P(m|\info_{k_2}) \int p(z_m|m,\info_{k_2}) \log \frac{P(m|\info_{k_2})\,p(z_m|m,\info_{k_2})}{P(m|\info_{k_1})\,p(z_m|m,\info_{k_1})} \,\text{d}z_m \nonumber \\
    &= \sum_{m=1}^{|\CM_m|} P(m|\info_{k_2}) \frac{P(m|\info_{k_2})}{P(m|\info_{k_1})} + \sum_{m=1}^{|\CM_m|} P(m|\info_{k_2}) \int p(z_m|m, \info_{k_2}) \log \frac{p(z_m|m, \info_{k_2})}{p(z_m|m, \info_{k_1})} \,\text{d}z_m \nonumber \\
    &= \DKL \( P_{M|\info_{k_2}} \,||\, P_{M|\info_{k_1}}\) + \EE_{M|\info_{k_2}} \[ \DKL \( p_{Z_m|\info_{k_2}} \,||\, p_{Z_m|\info_{k_1}}\) \], \nonumber 
\end{align}
where $0\leq k_1 \leq k_2 \leq N$, and $p(z_m|m, \info_{k})=p(z_m| \info_{k})$ per our notation convention. 
When setting $k_1=0$ and $k_2=N$, we recover the terminal reward in \cref{eq:terminal_info_gN} under the special case of $\alpha_M=\alpha_Z=1$ and $\alpha_\Param=0$.

\subsection{Information gain jointly on PoIs and QoIs} 
\label{app:joint_info_gain}

When nuisance parameters $\Nuis_m$ are absent, the IG jointly on the PoIs and QoIs given model $m$ is:
\begin{align}
    &\DKL \( p_{\Param_m,Z_m|\info_{k_2}} \,||\, p_{\Param_m,Z_m|\info_{k_1}}\) \nonumber\\
    &= \iint p(\param_m,z_m|\info_{k_2}) \log \frac{p(\param_m,z_m|\info_{k_2})}{p(\param_m,z_m|\info_{k_1})} \,\text{d}z_m\,\text{d}\param_m \nonumber\\
    &= \iint p(\param_m,z_m|\info_{k_2}) \log \frac{p(\param_m|\info_{k_2})\,p(z_m|\param_m,\info_{k_2})}{p(\param_m|\info_{k_1})\,p(z_m|\param_m,\info_{k_1})} \,\text{d}z_m\,\text{d}\param_m \nonumber\\
    &= \iint p(\param_m,z_m|\info_{k_2}) \log \frac{p(\param_m|\info_{k_2})\,p(z_m|\param_m)}{p(\param_m|\info_{k_1})\,p(z_m|\param_m)} \,\text{d}z_m\,\text{d}\param_m \nonumber\\
    &= \iint p(\param_m,z_m|\info_{k_2}) \log \frac{p(\param_m|\info_{k_2})}{p(\param_m|\info_{k_1})} \,\text{d}z_m\,\text{d}\param_m \nonumber\\
    &= \int p(\param_m|\info_{k_2}) \log \frac{p(\param_m|\info_{k_2})}{p(\param_m|\info_{k_1})} \,\text{d}\param_m \nonumber\\
    &= \DKL \( p_{\Param_m|\info_{k_2}} \,||\, p_{\Param_m|\info_{k_1}}\), \nonumber
\end{align}
where $0\leq k_1 \leq k_2 \leq N$, and the third equality is due to $Z_m$ only depending on $\Param_m$ when $\Nuis_m$ is absent (see \cref{eqn:predictive}). Hence, the IG on the QoIs is fully absorbed into the IG on the PoIs when nuisance parameters are absent.

\subsection{Proof of Theorem 1 (Terminal-incremental equivalence)} 
\label{app:terminal_incre}

\begin{proof}
We first decompose $U_T(\policy)$
into four summing parts:
\begin{align}
    U_T(\policy) = U_{T,\text{NIG}}(\policy) + U_{T,M}(\policy) + U_{T,\Param}(\policy) + U_{T,Z}(\policy), \nonumber
\end{align}
where $U_{T,\text{NIG}}(\policy)$ 
captures any non-IG reward contributions, and the other three parts are (with the conditioning on $\Info_0$ written out explicitly for the prior terms)
\begin{align}
    U_{T,M}(\policy) &= \alpha_M \EE_{Y_{0:N-1}|\policy,s_0}\[ \DKL \( P_{M|\Info_{N}} \,||\, P_{M|\Info_0}\) \], \nonumber \\
    U_{T,\Param}(\policy) &= \alpha_\Param \EE_{Y_{0:N-1}|\policy,s_0}\EE_{M|\Info_N} \[  \DKL\(\, p_{\Param_m|\Info_N}\,||\,p_{\Param_m|\Info_0} \,\)\], \nonumber \\
    U_{T,Z}(\policy) &= \alpha_Z \EE_{Y_{0:N-1}|\policy,s_0}\EE_{M|\Info_N} \[ \DKL\(\, p_{Z_m|\Info_N}\,||\,p_{Z_m|\Info_0} \,\) \]. \nonumber
\end{align}
Similarly, $U_I(\policy)$
is also decomposed into four summing parts:
\begin{align}
    U_I(\policy) = U_{I,\text{NIG}}(\policy) + U_{I,M}(\policy) + U_{I,\Param}(\policy) + U_{I,Z}(\policy), \nonumber
\end{align}
where $U_{I,\text{NIG}}(\policy)$ %
captures any non-IG reward contributions, and the other three parts are
\begin{align}
    U_{I,M}(\policy) &= \alpha_M \EE_{Y_{0:N-1}|\policy,s_0}\[ \sum_{k=0}^{N-1} \DKL \( P_{M|\Info_{k+1}} \,||\, P_{M|\Info_k}\) \] \nonumber \\
    U_{I,\Param}(\policy) &= \alpha_\Param \EE_{Y_{0:N-1}|\policy,s_0} \[ \sum_{k=0}^{N-1} \EE_{M|\Info_{k+1}} \[ \DKL\(\, p_{\Param_m|\Info_{k+1}}\,||\,p_{\Param_m|\Info_k} \,\)\]\] \nonumber \\
    U_{I,Z}(\policy) &= \alpha_Z \EE_{Y_{0:N-1}|\policy,s_0} \[\sum_{k=0}^{N-1} \EE_{M|\Info_{k+1}} \[ \DKL\(\, p_{Z_m|\Info_{k+1}}\,||\,p_{Z_m|\Info_k} \,\) \]\]. \nonumber
\end{align}
Since TIG and IIG formulations only entail the IG contributions, the non-IG reward contributions are not affected by this choice, and hence
\begin{align}
    U_{T,\text{NIG}}(\policy) = U_{I,\text{NIG}}(\policy). \nonumber
\end{align}

For the part corresponding to the IG contribution from model indicator:
\begin{align}
    &U_{I,M}(\policy) - U_{T,M}(\policy) \nonumber \\
    &= \alpha_M \EE_{Y_{0:N-1}|\policy,s_0}\[ \sum_{k=0}^{N-1} \DKL \( P_{M|\Info_{k+1}} \,||\, P_{M|\Info_k}\) - \DKL \( P_{M|\Info_{N}} \,||\, P_{M|\Info_0}\)\] \nonumber \\
    &= \alpha_M \EE_{Y_{0:N-1}|\policy,s_0}\[ \sum_{k=0}^{N-1} \sum_{m=1}^{|\CM_m|} P(m|\Info_{k+1}) \log \frac{P(m|\Info_{k+1})}{P(m|\Info_k)} - \sum_{m=1}^{|\CM_m|} P(m|\Info_N) \log \frac{P(m|\Info_N)}{P(m|\Info_0)} \] \nonumber \\
    &= \alpha_M \sum_{m=1}^{|\CM_m|} \EE_{Y_{0:N-1}|\policy,s_0}\[ \sum_{k=0}^{N-1}  P(m|\Info_{k+1}) \log \frac{P(m|\Info_{k+1})}{P(m|\Info_k)} - P(m|\Info_N) \log \frac{P(m|\Info_N)}{P(m|\Info_0)} \] \nonumber \\
    &= \alpha_M \sum_{m=1}^{|\CM_m|} \EE_{Y_{0:N-1}|\policy,s_0}\Bigg[ \sum_{k=0}^{N-2}  P(m|\Info_{k+1}) \log \frac{P(m|\Info_{k+1})}{P(m|\Info_k)} + P(m|\Info_N) \log \frac{P(m|\Info_N)}{P(m|\Info_{N-1})} 
    \nonumber\\
    & \qquad\qquad\qquad\qquad\qquad
    - P(m|\Info_N) \log \frac{P(m|\Info_N)}{P(m|\Info_0)} \Bigg] \nonumber \\
    &= \alpha_M \sum_{m=1}^{|\CM_m|} \EE_{Y_{0:N-1}|\policy,s_0}\[ \sum_{k=0}^{N-2}  P(m|\Info_{k+1}) \log \frac{P(m|\Info_{k+1})}{P(m|\Info_k)} - P(m|\Info_N) \log \frac{P(m|\Info_{N-1})}{P(m|\Info_0)}\] \nonumber \\
    &= \alpha_M \sum_{m=1}^{|\CM_m|} \EE_{Y_{0:N-2}|\policy,s_0}\Bigg[ \sum_{k=0}^{N-2}  P(m|\Info_{k+1}) \log \frac{P(m|\Info_{k+1})}{P(m|\Info_k)} 
    \nonumber \\
    &\qquad\qquad\qquad\qquad\qquad
    - 
    \EE_{Y_{N-1}|Y_{0:N-2},\policy,s_0}
    \[ P(m|\Info_N) \log \frac{P(m|\Info_{N-1})}{P(m|\Info_0)}\]\Bigg] \nonumber \\
    &= \alpha_M \sum_{m=1}^{|\CM_m|} \EE_{Y_{0:N-2}|\policy,s_0}\[ \sum_{k=0}^{N-2}  P(m|\Info_{k+1}) \log \frac{P(m|\Info_{k+1})}{P(m|\Info_k)} - P(m|\Info_{N-1}) \log \frac{P(m|\Info_{N-1})}{P(m|\Info_0)}\] \nonumber \\
    &\quad \vdots \nonumber \\
    &= \alpha_M \sum_{m=1}^{|\CM_m|} \EE_{Y_0|\policy,s_0}\[ \sum_{k=0}^{0}  P(m|\Info_{k+1}) \log \frac{P(m|\Info_{k+1})}{P(m|\Info_k)} - P(m|\Info_1) \log \frac{P(m|\Info_1)}{P(m|\Info_0)}\] \nonumber \\
    &= 0, \nonumber
\end{align}
where the seventh equality is due to
\begin{align}
    &\EE_{Y_{N-1}|Y_{0:N-2},\policy,s_0} 
    \[P(m|\Info_N) \log \frac{P(m|\Info_{N-1})}{P(m|\Info_0)}\] \nonumber \\
    &= \int p(y_{N-1}|Y_{0:N-2},\policy,s_0) P(m|\Info_N)  \log \frac{P(m|\Info_{N-1})}{P(m|\Info_0)} \,\text{d}y_{N-1} \nonumber \\
    &= \int p(y_{N-1}|\policy,\Info_{N-1}) P(m|y_{N-1},\policy,\Info_{N-1})  \log \frac{P(m|\Info_{N-1})}{P(m|\Info_0)} \,\text{d}y_{N-1} \nonumber \\
    &= \int p(y_{N-1},m|\policy,\Info_{N-1}) \log \frac{P(m|\Info_{N-1})}{P(m|\Info_0)} \,\text{d}y_{N-1} \nonumber \\
    &= P(m|\Info_{N-1}) \log \frac{P(m|\Info_{N-1})}{P(m|\Info_0)}, \nonumber
\end{align}
and the eighth equality results from repeatedly applying the steps between the third and seventh equalities $N-1$ times.

For the part corresponding to the IG contribution from PoIs:
\begin{align}
    &U_{I,\Param}(\policy) - U_{T,\Param}(\policy) \nonumber \\
    &= \alpha_\Param \EE_{Y_{0:N-1}|\policy,s_0}\[ \sum_{k=0}^{N-1} \EE_{M|\Info_{k+1}} \DKL\(\, p_{\Param_m|\Info_{k+1}}\,||\,p_{\Param_m|\Info_k} \,\) - \EE_{M|\Info_N}  \DKL\(\, p_{\Param_m|\Info_N}\,||\,p_{\Param_m|\Info_0} \,\) \] \nonumber \\
    &= \alpha_\Param \sum_{m=1}^{|\CM_m|} \EE_{Y_{0:N-1}|\policy,s_0}\Bigg[ \sum_{k=0}^{N-1} P(m|\Info_{k+1}) \DKL\(\, p_{\Param_m|\Info_{k+1}}\,||\,p_{\Param_m|\Info_k}\,\) 
    \nonumber \\
    &\qquad\qquad\qquad\qquad\qquad
    - P(m|\Info_N)   \DKL\(\, p_{\Param_m|\Info_N}\,||\,p_{\Param_m|\Info_0} \,\)\Bigg] \nonumber \\
    &= \alpha_\Param \sum_{m=1}^{|\CM_m|} \EE_{Y_{0:N-1}|\policy,s_0}\Bigg[ \sum_{k=0}^{N-2} P(m|\Info_{k+1}) \DKL\(\, p_{\Param_m|\Info_{k+1}}\,||\,p_{\Param_m|\Info_k} \,\) 
    \nonumber \\
    &\qquad\qquad\qquad\qquad\qquad
    + P(m|\Info_N)   \DKL\(\, p_{\Param_m|\Info_N}\,||\,p_{\Param_m|\Info_{N-1}} \,\) \nonumber \\
    &\qquad\qquad\qquad\qquad\qquad - P(m|\Info_N)   \DKL\(\, p_{\Param_m|\Info_N}\,||\,p_{\Param_m|\Info_0} \,\) \Bigg] \nonumber \\
    &= \alpha_\Param \sum_{m=1}^{|\CM_m|} \EE_{Y_{0:N-1}|\policy,s_0}\Bigg[ \sum_{k=0}^{N-2} P(m|\Info_{k+1}) \DKL\(\, p_{\Param_m|\Info_{k+1}}\,||\,p_{\Param_m|\Info_k} \,\) \nonumber\\
    &\qquad\qquad\qquad\qquad\qquad + P(m|\Info_N) \int p(\param_m|\Info_N) \log \frac{p(\param_m|\Info_0)}{p(\param_m|\Info_{N-1})} \,\text{d}\param_m \Bigg] \nonumber \\
    &= \alpha_\Param \sum_{m=1}^{|\CM_m|} \EE_{Y_{0:N-2}|\policy,s_0}\Bigg[ \sum_{k=0}^{N-2} P(m|\Info_{k+1}) \DKL\(\, p_{\Param_m|\Info_{k+1}}\,||\,p_{\Param_m|\Info_k} \,\) \nonumber\\
    &\qquad\qquad\qquad\qquad\qquad + 
    \EE_{Y_{N-1}|Y_{0:N-2},\policy,s_0} 
    \[ P(m|\Info_N) \int p(\param_m|\Info_N) \log \frac{p(\param_m|\Info_0)}{p(\param_m|\Info_{N-1})} \,\text{d}\param_m \] \Bigg] \nonumber \\
    &= \alpha_\Param \sum_{m=1}^{|\CM_m|} \EE_{Y_{0:N-2}|\policy,s_0}\Bigg[ \sum_{k=0}^{N-2} P(m|\Info_{k+1}) \DKL\(\, p_{\Param_m|\Info_{k+1}}\,||\,p_{\Param_m|\Info_k}\,\)  
    \nonumber\\
    &\qquad\qquad\qquad\qquad\qquad 
    - P(m|\Info_{N-1})  \DKL\(\, p_{\Param_m|\Info_{N-1}}\,||\,p_{\Param_m|\Info_0} \) \Bigg] \nonumber \\
    &\quad \vdots \nonumber \\
    &= \alpha_\Param \sum_{m=1}^{|\CM_m|} \EE_{Y_0|\policy,s_0}\Bigg[ \sum_{k=0}^{0} P(m|\Info_{k+1}) \DKL\(\, p_{\Param_m|\Info_{k+1}}\,||\,p_{\Param_m|\Info_k} \,\)  - P(m|\Info_1)  \DKL\(\, p_{\Param_m|\Info_{1}}\,||\,p_{\Param_m|\Info_0}\) \Bigg]  \nonumber \\
    &= 0, \nonumber
\end{align}
where 
the sixth equality is due to
\begin{align}
    & \EE_{Y_{N-1}|Y_{0:N-2},\policy,s_0} 
    \[P(m|\Info_N) \int p(\param_m|\Info_N) \log \frac{p(\param_m|\Info_0)}{p(\param_m|\Info_{N-1})} \,\text{d}\param_m \nonumber \] \\
    &= \int P(y_{N-1}|Y_{0:N-2},\policy,s_0) P(m|\Info_N) \int p(\param_m|\Info_N) \log \frac{p(\param_m|\Info_0)}{p(\param_m|\Info_{N-1})} \,\text{d}\param_m \,\text{d}y_{N-1} \nonumber \\
    &= \int P(y_{N-1}|\policy,\Info_{N-1}) P(m|y_{N-1},\policy,\Info_{N-1}) \int p(\param_m|m,y_{N-1},\policy,\Info_{N-1}) \log \frac{p(\param_m|\Info_0)}{p(\param_m|\Info_{N-1})} \,\text{d}\param_m \,\text{d}y_{N-1} \nonumber \\
    &= \iint P(y_{N-1},m,\param_m|\policy,\Info_{N-1}) \log \frac{p(\param_m|\Info_0)}{p(\param_m|\Info_{N-1})} \,\text{d}\param_m \,\text{d}y_{N-1} \nonumber \\
    &= P(m|\Info_{N-1}) \int P(\param_m|\Info_{N-1}) \log \frac{p(\param_m|\Info_0)}{p(\param_m|\Info_{N-1})} \,\text{d}\param_m \nonumber \\
    &= - P(m|\Info_{N-1}) \DKL\(\, p_{\Param_m|\Info_{N-1}}\,||\,p_{\Param_m|\Info_{0} }\,\), \nonumber 
\end{align}
and the seventh equality results from repeatedly applying the steps between the second and sixth equalities $N-1$ times.

For the part corresponding to the IG contribution from QoIs, the derivation is identical as above for the PoIs except $\Param$ is replaced with $Z$, to arrive at
\begin{align}
    U_{I,Z}(\policy) - U_{T,Z}(\policy) = 0. \nonumber
\end{align}

Combining the equivalence results from all four parts, we obtain
\begin{align}
    U_I(\policy) = U_T(\policy) \nonumber
\end{align}
for any policy $\policy$.
\end{proof}

\clearpage

\subsection{Proof of Theorem 2 (One-point equivalence)} 
\label{app:one_point_expected_utility}

\begin{proof}
We begin by proving the equivalence of expected utility under TIG and one-point-TIG:
\begin{align}
    U_T(\policy) 
    &= \EE_{Y_{0:N-1}|\policy,s_0} \Bigg[\sum_{k=0}^{N-1}r_k(S_k,\design_k,y_k)+r_N(S_N)\Bigg] \nonumber \\
    &= \EE_{Y_{0:N-1}|\policy,s_0} \Bigg[ \alpha_{M} \DKL (\, P_{M|\Info_N}\,||\,P_M \,) \nonumber \\
 &\qquad\qquad\qquad\qquad + \EE_{M|\Info_N} \Bigg[ \alpha_{\Param} \DKL(\, p_{\Param_m|\Info_N}\,||\,p_{\Param_m} \,) + \alpha_Z \DKL (\, p_{Z_m|\Info_N}\,||\,p_{Z_m} \,) \Bigg] \nonumber \\
    &= \EE_{Y_{0:N-1}|\policy,s_0} \Bigg[ \alpha_M \EE_{M|\Info_N} \log \frac{P(M|\Info_N)}{P(M)} \nonumber \\
 &\qquad\qquad\qquad\qquad + \EE_{M|\Info_N} \[ \alpha_\Param \EE_{\Param_m|\Info_N} \log \frac{p(\Param_m|\Info_N)}{p(\Param_m)} + \alpha_Z \EE_{Z_m|\Info_N} \log \frac{p(Z_m|\Info_N)}{p(Z_m)} \] \Bigg] \nonumber \\
    &= \EE_{M,Y_{0:N-1}|\policy,s_0} \Bigg[ \alpha_M \log \frac{P(M|\Info_N)}{P(M)} \nonumber \\
 &\qquad\qquad\qquad\qquad + \alpha_\Param \EE_{\Param_m|\Info_N} \log \frac{p(\Param_m|\Info_N)}{p(\Param_m)} + \alpha_Z \EE_{Z_m|\Info_N} \log \frac{p(Z_m|\Info_N)}{p(Z_m)} \Bigg] \nonumber \\
    &= \EE_{M,Y_{0:N-1}|\policy,s_0} \Bigg[ \alpha_M \EE_{\Param_m,Z_m|\Info_N} \log \frac{P(M|\Info_N)}{P(M)} \nonumber \\
 &\qquad\qquad\qquad\qquad + \alpha_\Param \EE_{\Param_m,Z_m|\Info_N} \log \frac{p(\Param_m|\Info_N)}{p(\Param_m)} + \alpha_Z \EE_{\Param_m,Z_m|\Info_N} \log \frac{p(Z_m|\Info_N)}{p(Z_m)} \Bigg] \nonumber \\
    &= \EE_{M,\Param_m,Z_m,Y_{0:N-1}|\policy,s_0} \[ \alpha_M \log \frac{P(M|\Info_N)}{P(M)} + \alpha_\Param \log \frac{p(\Param_m|\Info_N)}{p(\Param_m)} + \alpha_Z \log \frac{p(Z_m|\Info_N)}{p(Z_m)} \] \nonumber \\
    &= \EE_{M,\Param_m,\Nuis_m,Z_m,Y_{0:N-1}|\policy,s_0} \[ \alpha_M \log \frac{P(M|\Info_N)}{P(M)}   + \alpha_\Param \log \frac{p(\Param_m|\Info_N)}{p(\Param_m)} + \alpha_Z \log \frac{p(Z_m|\Info_N)}{p(Z_m)} \] \nonumber \\
    &= \EE_{M_{0},\Param_{m,0},\Nuis_{m,0},Z_{m,0}}\EE_{Y_{0:N-1}|\policy,s_0,M_{0},\Param_{m,0},\Nuis_{m,0},Z_{m,0}} \Bigg[ \alpha_M \log \frac{P(M_{0}|\Info_N)}{P(M_{0})}  \nonumber \\
    &\hspace{18em} +   \alpha_\Param \log \frac{p(\Param_{m,0}|\Info_N)}{p(\Param_{m,0})} + \alpha_Z \log \frac{p(Z_{m,0}|\Info_N)}{p(Z_{m,0})} \Bigg] \nonumber \\
    &= \EE_{M_{0},\Param_{m,0},\Nuis_{m,0},Z_{m,0}}\EE_{\Info_N|\policy,s_0,M_{0},\Param_{m,0},\Nuis_{m,0}} \Bigg[ \alpha_M \log \frac{P(M_{0}|\Info_N)}{P(M_{0})}  \nonumber \\
    &\hspace{18em} +   \alpha_\Param \log \frac{p(\Param_{m,0}|\Info_N)}{p(\Param_{m,0})} + \alpha_Z \log \frac{p(Z_{m,0}|\Info_N)}{p(Z_{m,0})} \Bigg] \nonumber \\
    &= \check{U}_T(\policy), \nonumber
\end{align}
where at several occasions we use the equivalence between $\EE_{Y_{0:k-1}|\policy,s_0,\ldots}$ and $\EE_{\Info_k|\policy,s_0,\ldots}$, and between $\EE_{\ldots|\Info_k}$ and $\EE_{\ldots|Y_{0:k-1},\policy,s_0}$, in the eighth equality the random variables $M$, $\Param_m$, $\Nuis_m$, $Z_m$ are replaced with their oracle versions $M_0$, $\Param_{m,0}$, $\Nuis_{m,0}$, $Z_{m,0}$ which have identical distributions, and the ninth equality omits the conditioning on $Z_{m,0}$ in the inner expectation since $\Info_N$ does not depend on $Z_{m,0}$.

Next, we have $U_T(\policy) = U_I(\policy)$ through \cref{prop:terminal_incremental} and its proof in \cref{app:terminal_incre}. Finally, we show the equivalence between $\check{U}_I(\policy)$ and $\check{U}_T(\policy)$ by cancelling out all intermediate posteriors:
\begin{align}
    & \check{U}_I(\policy) \nonumber\\
    &= \EE_{M_{0},\Param_{m,0},\Nuis_{m,0},Z_{m,0}}\EE_{\Info_N|\policy,s_0,M_{0},\Param_{m,0},\Nuis_{m,0}}\sum_{k=0}^{N-1} \Bigg[ \alpha_M \log \frac{P(M_0|\Info_{k+1})}{P(M_0|\Info_k)} + \alpha_\Param \log \frac{p(\Param_{m,0}|\Info_{k+1})}{p(\Param_{m,0}|\Info_k)} \nonumber \\ 
    &\hspace{19.5em} + \alpha_Z \log \frac{p(Z_{m,0}|\Info_{k+1})}{p(Z_{m,0}|\Info_k)} \Bigg] 
    \nonumber\\
    &= \EE_{M_{0},\Param_{m,0},\Nuis_{m,0},Z_{m,0}}\EE_{\Info_N|\policy,s_0,M_{0},\Param_{m,0},\Nuis_{m,0}} \Bigg[ \alpha_M \log \frac{P(M_0|\Info_N)}{P(M_0)} + \alpha_\Param \log \frac{p(\Param_{m,0}|\Info_N)}{p(\Param_{m,0})} \nonumber \\ &\hspace{19.5em} + \alpha_Z \log \frac{p(Z_{m,0}|\Info_N)}{p(Z_{m,0})} \Bigg]
    \nonumber\\
    &= \check{U}_T(\policy). \nonumber
\end{align}
Combining the above equivalence results together, we arrive at 
\begin{align}
U_T(\policy) = \check{U}_T(\policy)= \check{U}_I(\policy) =  U_I(\policy) \nonumber
\end{align}
for any policy $\policy$.
\end{proof}

\subsection{Omitting prior terms in the expected utility} 
\label{app:omit_prior}

When the prior terms $p(\cdot)$ are omitted, the expected utility is shifted
by the amount: 
\begin{align}
    &\EE_{M_{0},\Param_{m,0},\Nuis_{m,0},Z_{m,0}}\EE_{\Info_N|\policy,s_0,M_{0},\Param_{m,0},\Nuis_{m,0}} \[ \alpha_M \log P(M_0) + \alpha_\Param \log p(\Param_{m,0}) + \alpha_Z \log p(Z_{m,0}) \] \nonumber \\
    &= \EE_{M_{0},\Param_{m,0},\Nuis_{m,0},Z_{m,0}} \[ \alpha_M \log P(M_0) + \alpha_\Param \log p(\Param_{m,0}) + \alpha_Z \log p(Z_{m,0}) \], \nonumber
\end{align}
which does not depend on the policy $\policy$. In other words, the expected utility is shifted by a constant. Hence, whether including or omitting the prior terms will not affect the optimized policy. The same result holds for both TIG and IIG rewards, and their one-point versions and variational one-point versions.

\subsection{Proof of Theorem 3 (Variational lower bound)} 
\label{app:variational_expected_utility_lower_bound}

\begin{proof}
First, we prove the equivalence between the expected utility under {variational-one-point-TIG} and {variational-one-point-IIG}:
\begin{align}
    &\tilde{U}_I(\policy;\phi) \nonumber \\
    &=\EE_{M_{0},\Param_{m,0},\Nuis_{m,0},Z_{m,0}}\EE_{\Info_N|\policy,s_0,M_{0},\Param_{m,0},\Nuis_{m,0}} \sum_{k=0}^{N-1} \Bigg[ \alpha_M \log \frac{q(M_0|\Info_{k+1};\phi_M)}{q(M_0|\Info_k;\phi_M)} \nonumber \\
    &\hspace{16em} + \alpha_\Param \log  \frac{q(\Param_{m,0}|\Info_{k+1};\phi_{\Param_m})}{q(\Param_{m,0}|\Info_k;\phi_{\Param_m})} + \alpha_Z \log \frac{q(Z_{m,0}|\Info_{k+1};\phi_{Z_m})}{q(Z_{m,0}|\Info_k;\phi_{Z_m})}   \Bigg] \nonumber \\
    &=\EE_{M_{0},\Param_{m,0},\Nuis_{m,0},Z_{m,0}}\EE_{\Info_N|\policy,s_0,M_{0},\Param_{m,0},\Nuis_{m,0}} \Bigg[ \nonumber \\
    &\hspace{5em}\alpha_M \left(\log \frac{q(M_0|\Info_{1};\phi_M)}{P(M_0)} + \log \frac{q(M_0|\Info_{2};\phi_M)}{q(M_0|\Info_1;\phi_M)} + \cdots + \log \frac{q(M_0|\Info_{N};\phi_M)}{q(M_0|\Info_{N-1};\phi_M)}\right) \nonumber \\
    &\hspace{5em} + \alpha_\Param \left(\log \frac{q(\Param_{m,0}|\Info_{1};\phi_{\Param_m})}{p(\Param_{m,0})} + \log \frac{q(\Param_{m,0}|\Info_{2};\phi_{\Param_m})}{q(\Param_{m,0}|\Info_1;\phi_{\Param_m})} + \cdots + \log \frac{q(\Param_{m,0}|\Info_{N};\phi_{\Param_m})}{q(\Param_{m,0}|\Info_{N-1};\phi_{\Param_m})}\right) \nonumber \\
    &\hspace{5em} + \alpha_Z \left(\log \frac{q(Z_{m,0}|\Info_{1};\phi_{Z_m})}{p(Z_{m,0})} + \log \frac{q(Z_{m,0}|\Info_{2};\phi_{Z_m})}{q(Z_{m,0}|\Info_1;\phi_{Z_m})} + \cdots + \log \frac{q(Z_{m,0}|\Info_{N};\phi_{Z_m})}{q(Z_{m,0}|\Info_{N-1};\phi_{Z_m})}\right) \Bigg] \nonumber \\
    &=\EE_{M_{0},\Param_{m,0},\Nuis_{m,0},Z_{m,0}}\EE_{\Info_N|\policy,s_0,M_{0},\Param_{m,0},\Nuis_{m,0}} \Bigg[ \alpha_M \log \frac{q(M_0|\Info_N;\phi_M)}{P(M_0)}  \nonumber \\
    &\hspace{16em} + \alpha_\Param \log \frac{q(\Param_{m,0}|\Info_N;\phi_{\Param_m})}{p(\Param_{m,0})} 
    + \alpha_Z \log \frac{q(Z_{m,0}|\Info_N;\phi_{Z_m})}{p(Z_{m,0})} \Bigg] \nonumber \\
    &= \tilde{U}_T(\policy;\phi). \nonumber
\end{align}
Due to the cancellation of all intermediate variational posteriors, only the prior and final variational posterior terms survive. Since the prior density is usually accessible analytically or omitted per \cref{app:omit_prior}, the accuracy of the variational expected utility only depends on the quality of the final variational posterior approximations, $q(\cdot|\Info_N;\phi_{(\cdot)})$. 

Next, we prove the lower bound for {variational-one-point-TIG}:
\begin{align}
    & {U}_T(\policy) - \tilde{U}_T(\policy;\phi) \nonumber\\
    &= \check{U}_T(\policy) - \tilde{U}_T(\policy;\phi) \nonumber \\
    &= \EE_{M_{0},\Param_{m,0},\Nuis_{m,0},Z_{m,0}}\EE_{\Info_N|\policy,s_0,M_{0},\Param_{m,0},\Nuis_{m,0}}\Bigg[  \alpha_M \log \frac{P(M_0|\Info_N)}{q(M_0|\Info_N;\phi_M)} + \alpha_\Param \log \frac{p(\Param_{m,0}|\Info_N)}{q(\Param_{m,0}|\Info_N;\phi_{\Param_m})} \nonumber \\
    &\hspace{17.5em} + \alpha_Z \log \frac{p(Z_{m,0}|\Info_N)}{q(Z_{m,0}|\Info_N;\phi_{Z_m})} \Bigg] \nonumber \\
    &= \alpha_M \EE_{M_{0},\Param_{m,0},\Nuis_{m,0},Z_{m,0}}\EE_{\Info_N|\policy,s_0,M_{0},\Param_{m,0},\Nuis_{m,0}} \Bigg[ \log \frac{P(M_0|\Info_N)}{q(M_0|\Info_N;\phi_M)} \Bigg] \nonumber \\
    &\quad + \alpha_\Param \EE_{M_{0},\Param_{m,0},\Nuis_{m,0},Z_{m,0}}\EE_{\Info_N|\policy,s_0,M_{0},\Param_{m,0},\Nuis_{m,0}} \Bigg[ \log \frac{p(\Param_{m,0}|\Info_N)}{q(\Param_{m,0}|\Info_N;\phi_{\Param_m})} \Bigg] \nonumber \\
    &\quad + \alpha_Z \EE_{M_{0},\Param_{m,0},\Nuis_{m,0},Z_{m,0}}\EE_{\Info_N|\policy,s_0,M_{0},\Param_{m,0},\Nuis_{m,0}} \Bigg[ \log \frac{p(Z_{m,0}|\Info_N)}{q(Z_{m,0}|\Info_N;\phi_{Z_m})} \Bigg] \nonumber \\
    &= \alpha_M \EE_{M_{0}, \Info_N|\policy, s_0} \Bigg[ \log \frac{P(M_0|\Info_N)}{q(M_0|\Info_N;\phi_M)} \Bigg] \nonumber \\
    &\quad + \alpha_\Param \EE_{M_{0},\Param_{m,0}, \Info_N|\policy, s_0} \Bigg[ \log \frac{p(\Param_{m,0}|\Info_N)}{q(\Param_{m,0}|\Info_N;\phi_{\Param_m})} \Bigg] \nonumber \\
    &\quad + \alpha_Z \EE_{M_{0}, Z_{m,0},  \Info_N|\policy, s_0} \Bigg[ \log \frac{p(Z_{m,0}|\Info_N)}{q(Z_{m,0}|\Info_N;\phi_{Z_m})} \Bigg] \nonumber \\
    &= \alpha_M \EE_{\Info_N|\policy, s_0} \EE_{M_0|\Info_N} \Bigg[ \log \frac{P(M_0|\Info_N)}{q(M_0|\Info_N;\phi_M)} \Bigg] \nonumber \\
    &\quad + \alpha_\Param \EE_{M_0,\Info_N|\policy, s_0} \EE_{\Param_{m,0}|M_0,\Info_N} \Bigg[ \log \frac{p(\Param_{m,0}|\Info_N)}{q(\Param_{m,0}|\Info_N;\phi_{\Param_m})} \Bigg] \nonumber \\
    &\quad + \alpha_Z \EE_{M_0,\Info_N|\policy, s_0} \EE_{Z_{m,0}|M_0,\Info_N} \Bigg[ \log \frac{p(Z_{m,0}|\Info_N)}{q(Z_{m,0}|\Info_N;\phi_{Z_m})} \Bigg] \nonumber \\
    &= \alpha_M \EE_{\Info_N|\policy, s_0} \Bigg[ \DKL \( P_{M_0|\Info_N} \,||\, q_{M_0|\Info_N;\phi_M}\) \Bigg] \nonumber \\
    &\quad + \alpha_\Param \EE_{M_0,\Info_N|\policy, s_0} \Bigg[ \DKL \( p_{\Param_{m,0}|\Info_N} \,||\, q_{\Param_{m,0}|\Info_N;\phi_{\Param_m}}\) \Bigg] \nonumber \\
    &\quad + \alpha_Z \EE_{M_0,\Info_N|\policy, s_0} \Bigg[ \DKL \( p_{Z_{m,0}|\Info_N} \,||\, q_{Z_{m,0}|\Info_N;\phi_{Z_m}}\) \Bigg] \nonumber \\
    & \ge 0, \nonumber
\end{align}
where the first equality invokes \cref{prop:one_point_expected_utility},
the sixth equality is due to $p(\Param_{m,0}|\Info_N)$ being equivalent to $p(\Param_{m,0}|M_0,\Info_N)$ and $p(Z_{m,0}|\Info_N)$ being equivalent to $p(Z_{m,0}|M_0,\Info_N)$, and the final inequality is due to the non-negativity of KL divergence terms and that we require $\alpha_{M}, \alpha_{\Param}, \alpha_{Z} \in [0,1]$. 
The bound is tight if and only if $q(\cdot|\Info_N;\phi_{(\cdot)}) = p(\cdot|\Info_N)$ and so all the KL divergence terms become zero; (except the trivial case when $\alpha_M= \alpha_\Param= \alpha_Z = 0$, under which $U$, $\check{U}$, and $\tilde{U}$ will always be identically zero).

Putting everything together, we arrive at
\begin{align}
\tilde{U}_I(\policy;\phi) = \tilde{U}_T(\policy;\phi) \leq \check{U}_T(\policy) = \check{U}_I(\policy) = U_T(\policy) = U_I(\policy), \nonumber
\end{align}
for any $\policy$ and $\phi$, where the last three equalities result from \cref{prop:one_point_expected_utility}.
\end{proof}

\section{Algorithm details} 
\label{app:alg_details}

\subsection{Neural network architecture for approximate posteriors of model indicator}
\label{app:model_approximation}

The overall architecture for a NN-based approximate posterior of model indicator, $q(m|\info_k;\phi_M)$, is shown in \cref{tab:arch_model_post}; the same architecture is used for all numerical cases in this paper. 
The NN takes $\info_k$ as input, and outputs the log-probabilities of each candidate model, $\log q(m|\info_k;\phi_M)$.
Separate NNs are trained for each stage $k$ when the IIG formulation is used. 
As shown in the first part of \cref{app:variational_expected_utility_lower_bound}, the quality of the intermediate approximate posteriors does not directly contribute to the accuracy of the overall variational expected utility, 
and thus one may train these intermediate approximate posteriors more `roughly', for example,
by using simpler NN architectures and with shared weights among the NNs. 

\begin{table}[hbt]
    \centering
    \caption{Architecture for the NN-based approximate posteriors of model indicator.}
    \begin{tabular}{cccc}
    \toprule
    Layer & Description & Dimension & Activation \\
    \midrule 
    Input & $\info_k$ & $k(N_{\design}+N_y)$ & - \\
    Hidden 1 & {Dense} & 256 & ReLU \\
    Hidden 2 & Dense & 256 & ReLU \\
    Hidden 3 & Dense & 256 & ReLU \\
    Output & Dense & $|\mathcal{M}_m|$ %
    & LogSoftmax \\
    \bottomrule
    \end{tabular}
    \label{tab:arch_model_post}
\end{table}

\subsection{Neural network architectures for approximate posteriors of PoIs and QoIs}
\label{app:parameter_approximation}

We use GMMs and NFs as approximate posteriors for PoIs, $q(\param_m|\info_k;\phi_{\Param_m})$, and for QoIs, $q(z_m|\info_k;\phi_{Z_m})$\footnote{In this section, we have dropped the oracle subscript $0$ (from $\param_{m,0}$ and $z_{m,0}$) to simplify notation.}. 
The same architecture is used for both PoI and QoI cases, and we  introduce below only in the context of PoIs for simplicity.
Separate GMMs/NFs are trained for each stage $k$ when the IIG formulation is used.

\subsubsection{Independent Gaussian mixture models}

An independent GMM 
approximates a complex distribution 
through a weighted sum of multiple independent Gaussians:
\begin{align}
    q(\param_m|\info_k;\phi_{\Param_m}) = \sum_{i=1}^{n_{\text{mix}}} w_i(\info_k;\phi_{\Param_m}) \, \CN(\param_m; \mu_i(\info_k;\phi_{\Param_m}), \Sigma_i(\info_k;\phi_{\Param_m})),
    \label{eq:GMM}
\end{align}
where for the $i$th Gaussian, $w_i(\info_k;\phi_{\Param_m})$ is its mixture weight, $\mu_i(\info_k;\phi_{\Param_m}) \in \RR^{N_{\param_m}}$ is its mean, and $\Sigma_i(\info_k;\phi_{\Param_m}) \in \RR^{N_{\param_m} \times N_{\param_m}}$ is its diagonal covariance matrix with square root of the diagonal terms being the standard deviations. 
The weights, means, and standard deviations of the GMM are predicted using NNs, together referred to as the GMM net. These NNs share a common backend network that learns shared features. The architectures for the feature net, weight net, mean net and standard deviation net are provided in \cref{tab:arch_GMM_feature,tab:arch_GMM_weight,tab:arch_GMM_mean_sd}. 
The `Linear mapping' in \cref{tab:arch_GMM_mean_sd} refers to the process of mapping the output to a specific range that is problem dependent. This mapping ensures that the predicted means and standard deviations fall within the desired range. 
Additionally, a `nugget' of $10^{-27}$ is added to \cref{eq:GMM}
to prevent numerical underflow.
When certain PoIs have compact support, independent truncated normal distributions \cite{burkardt2014truncated} 
are used for the dimensions corresponding to those PoIs.
The specific ranges of the linear mapping and the usage of truncated normal will be stated in each numerical case. The same GMM net architecture is used across all numerical cases.

\begin{table}[htbp]
    \centering
    \caption{Architecture for the feature net within the GMM net.}
    \begin{tabular}{cccc}
    \toprule
    Layer & Description & Dimension & Activation \\
    \midrule 
    Input & $\info_k$ & $k(N_{\design}+N_y)$ & - \\
    Hidden 1 & Dense & 256 & ReLU \\
    Output & Dense & 256 & ReLU \\
    \bottomrule
    \end{tabular}
    \label{tab:arch_GMM_feature}
\end{table}

\begin{table}[htbp]
    \centering
    \caption{Architecture for the weight net within the GMM net.}
    \begin{tabular}{cccc}
    \toprule
    Layer & Description & Dimension & Activation \\
    \midrule 
    Input & Feature($\info_k$) & 256 & - \\
    Hidden 1 & Dense & 256 & ReLU \\
    Hidden 2 & Dense & 256 & ReLU \\
    Output & Dense & $n_{\text{mix}}$ & Softmax \\
    \bottomrule
    \end{tabular}
    \label{tab:arch_GMM_weight}
\end{table}

\begin{table}[htbp]
    \centering
    \caption{Architecture for the mean net and standard deviation net within the GMM net.}
    \begin{tabular}{cccc}
    \toprule
    Layer & Description & Dimension & Activation \\
    \midrule 
    Input & Feature($\info_k$) & 256 & - \\
    Hidden 1 & Dense & 256 & ReLU \\
    Hidden 2 & Dense & 256 & ReLU \\
    Hidden 3 & Dense & $n_{\text{mix}} N_{\param_m}$ & Sigmoid \\
    Output & Identity & $n_{\text{mix}} N_{\param_m}$ & Linear mapping \\
    \bottomrule
    \end{tabular}
    \label{tab:arch_GMM_mean_sd}
\end{table}

\subsubsection{Normalizing flows}
\label{app:NFs}

An NF is an invertible mapping from a target random variable $\Param_m \sim p_{\Param_m}(\param_m)$\footnote{While we use the prior as the target distribution here for the purpose of introducing NFs, generally we will target posteriors such as $p(\param_m|\info_k)$.} to a standard normal random variable, 
${\sf Z}_m \sim p_{{\sf Z}_m}({\sf z}_m)$$=\mathcal{N}({0},\mathbb{I})$\footnote{${\sf Z}$ denotes the standard normal random variable, not to be confused with the QoIs $Z$.},
of the same dimension: ${\sf Z}$$_m$  =$f(\Param_m)$ and $\Param_m$ = $g({\sf Z}$$_m )$ where $g \coloneqq f^{-1}$. In practice, we approximate $f$ via a  mapping parameterized with  $\phi_{\Param_m}$, which produces an approximate transformation $\widetilde{{\sf Z}}_m =f(\Param_m;\phi_{\Param_m})$ and its inverse $g(\,\cdot\, ; \phi_{\Param_m}) \coloneqq f(\,\cdot\, ;\phi_{\Param_m})^{-1}$ produces $\Param_m = g(\widetilde{{\sf Z}}_m ; \phi_{\Param_m})$. If acting on the exact standard normal ${\sf Z}_m $, then $\widetilde{\Param}_m = g({{\sf Z}_m }; \phi_{\Param_m})$ and also ${{\sf Z}_m }=f(\widetilde{\Param}_m;\phi_{\Param_m})$. 

In general, the approximate mappings used in NFs are often structured as compositions of successive simple invertible mappings: $f(\widetilde{\Param}_m;\phi_{\Param_m}) = f_n \circ f_{n-1} \circ ... \circ f_1(\widetilde{\Param}_m) = f_n(f_{n-1}(...(f_1({\widetilde{\Param}_m}))...))$ and $g({{\sf Z}_m };\phi_{\Param_m})=g_1 \circ ... g_{n-1}\circ g_n({{\sf Z}_m }) = g_1(g_{2}(...(g_n({{{\sf Z}_m }}))...))$ with $g_i=f_i^{-1}$ and $n\geq 1$. Note that all intermediate mappings $f_i$ and $g_i$ depend on ${\phi_{\Param_m}}$, but we omit their subscripts to simplify notation. 
The log density of $\widetilde{\Param}_m$ can be tracked via the change-of-variable formula:
\begin{align}
    \nonumber
    \log q_{\widetilde{\Param}_m}({\widetilde{\Param}_m={\param_m};\phi_{\Param_m}}) &= \log p_{{\sf Z}_m }(f_n \circ f_{n-1} \circ ... \circ f_1(\widetilde{\Param}_m={\param_m}))  \\ 
    &\quad\quad\quad + \sum_{i=1}^n \log \left|\text{det} \frac{\partial f_i \circ f_{i-1} \circ ... f_1(\widetilde{\Param}_m)}{\partial {\widetilde{\Param}_m}}\right|_{\widetilde{\Param}_m=\param_m},
    \label{eq:composing}
\end{align}
where $\frac{\partial f_{i}(\widetilde{\Param}_m) }{\partial \widetilde{\Param}_m}$is the Jacobian of $f_{i}$.
By applying successive transformations on ${\sf Z}_m $, the density of the resulting variable can be highly expressive~\cite{Rezende_16_Variational, Tabak_10_Density} and effective for multi-modal, skewed, or other non-standard distribution shapes. 

Among a range of choices for the architecture of invertible mappings~\cite{Papamakarios_21_NFsReview, Kobyzev_20_NFsReview}, 
we adopt the coupling layers~\cite{Dinh_16_NVP} as a special type of invertible neural network (INN)~\cite{Kruse_21_Hint, Radev_20_Bayesflow} for its efficient density evaluations and sampling in both forward ($f$) and inverse ($g$) directions. 
Several papers have shown that composing coupling layers can create flexible flows \cite{Kingma_18_Glow, Ardizzone_18_INN, Kruse_21_Hint}, and recent work by Draxler \textit{et al.} \cite{Draxler_24_Universality} shows that coupling layers form a distributional universal approximator. 
The basic form of the coupling layer that completes one full transformation starts by partitioning ${\sf Z}$$_m$  = $[{\sf Z}$$_{m_1}$, ${\sf Z}$$_{m_2}]^{\top}$ into two parts of approximately equal dimension---that is, ${\sf Z}$$_{m_1} $ $\in \RR^{N_{\param_{m,1}}}$, ${\sf Z}$$_{m_2}$ $ \in \RR^{N_{\param_{m,2}}}$, and $N_{\param_{m,1}}+N_{\param_{m,2}}=N_{\param_m}$---and then composes together $n=2$ transformations that transform one part at a time. This maps ${\sf Z}$$_m$ to an approximate target $\widetilde{{\Param}}_m$, i.e.,
$g({\sf Z}_m ;\phi_{\Param_m}) = g_1 \circ g_2({\sf Z}_m) =\widetilde{{\Param}}_m$, and is defined as:
\begin{align}
   g_2( {\sf Z}_m ) &= \begin{bmatrix}
\widetilde{{\Param}}_{m_1} = [{\sf Z}_{m_1} - {\sf t}_2( {\sf Z}_{m_2})] \odot \text{exp}[-{\sf s}_2( {\sf Z}_{m_2} )] \\
{\sf Z}_{m_2}
\end{bmatrix},  \nonumber \\
g_1(g_2( {\sf Z}_m )) &= \begin{bmatrix}
\widetilde{{\Param}}_{m_1} \\
\widetilde{{\Param}}_{m_2} = [{\sf Z}_{m_2} -{\sf t}_1(\widetilde{{\Param}}_{m_1}))] \odot \text{exp}(-{\sf s}_1(\widetilde{{\Param}}_{m_1})) 
\end{bmatrix}, \nonumber
\end{align}
where $\odot$ denotes element-wise product, and ${\sf s}_1, {\sf t}_1 : \mathbb{R}^{N_{\param_{m,1}}} \rightarrow \mathbb{R}^{N_{\param_{m,2}}}$ and ${\sf s}_2, {\sf t}_2 : \mathbb{R}^{N_{\param_{m,2}}} \rightarrow \mathbb{R}^{N_{\param_{m,1}}}$ are arbitrary functions. The parameterizations of these functions make up $\phi_{\Param_m}$; for instance, if these functions are represented by NNs, then $\phi_{\Param_m}$ encompasses all NNs' weight and bias parameters. 

The inverse of $g(\,\cdot\,;\phi_{\Param_m})$, which is $f(\widetilde{\Param}_m;\phi_{\Param_m}) = f_2 \circ f_1(\widetilde{\Param}_m)={\sf Z}_m$, similarly involves partitioning $\widetilde{\Param}_m = [\widetilde{\Param}_{m_1}, \widetilde{\Param}_{m_2}]$ and can be shown to be: 
\begin{align}
f_1(\widetilde{\Param}_m) &= \begin{bmatrix}
\widetilde{\Param}_{m_1} \\
{\sf Z}_{m_2} = \widetilde{\Param}_{m_2} \odot \exp({\sf s}_1(\widetilde{\Param}_{m_1})) + {\sf t}_1(\widetilde{\Param}_{m_1})
\end{bmatrix},  \nonumber\\
f_2(f_1(\widetilde{\Param}_m)) &= \begin{bmatrix}
{\sf Z}_{m_1} = \widetilde{\Param}_{m_1} \odot \exp ({\sf s}_2({\sf Z}_{m_2})) + {\sf t}_2({\sf Z}_{m_2}) \\
{\sf Z}_{m_2} 
\end{bmatrix} . \nonumber
\end{align}
The Jacobians of $f_1$ and $f_2$ are triangular matrices: 
\begin{align}
    \frac{\partial f_1(\widetilde{\Param}_m)}{\partial \widetilde{\Param}_m} = \begin{bmatrix}
    \mathbb{I}_{N_{\param_{m,1}}} & {0} \\
    \frac{\partial {\sf Z}_{m_2}} %
    {\partial \widetilde{\Param}_{m_1}} & \text{diag}(\exp({\sf s}_1(\widetilde{\Param}_{m_1})))
    \end{bmatrix}, \qquad
    \frac{\partial f_2(f_1(\widetilde{\Param}_m))}{\partial (f_1(\widetilde{\Param}_m))} = \begin{bmatrix} \text{diag}(\exp({\sf s}_2({\sf Z}_{m_2})))%
    & \frac{\partial {\sf Z}_{m_1}}{\partial {\sf Z}_{m_2}} \\
    {0}  & 
    \mathbb{I}_{N_{\param_{m,2}}} 
    \end{bmatrix}, \nonumber
\end{align} 
and have respective determinants $\exp(\sum_{j=1}^{N_{\param_{m,2}}}{\sf s}_1(\widetilde{\Param}_{m_1})_j)$  and $\exp(\sum_{j=1}^{N_{\param_{m,1}}}{\sf s}_2({\sf Z}_{m_2})_j)$.

Multiple such complete transformations can be composed together for greater expressiveness. For example, adopting NFs with $n_{\text{trans}}=3$ sets of complete transformations entails
$f(\widetilde{\Param}_m;\phi_{\Param_m}) = (f_2 \circ f_{1})^{T3} \circ (f_2 \circ f_{1})^{T2} \circ (f_2 \circ f_{1})^{T1}(\widetilde{\Param}_m)$ and $g({{\sf Z}_m};\phi_{\Param_m})=(g_1 \circ g_{2})^{T3} \circ (g_1 \circ g_{2})^{T2} \circ (g_1 \circ g_{2})^{T1} ({{\sf Z}_m})$.

To incorporate the $\info_k$-dependence into the approximate posteriors $q(\param_m|\info_k; \phi_{\Param_m})$, the ${\sf s}$ and ${\sf t}$ functions are designed to additionally take $\info_k$ as input, leading to a form of conditional INNs (cINNs)~\cite{Padmanabha_21_cINN}. Similar to the GMM setup, $\info_k$ is first fed into a feature network whose output has the same dimension as $\info_k$. The architectures of the feature network, and the ${\sf s}_1$, ${\sf t}_1$, ${\sf s}_2$, ${\sf t}_2$  networks in NFs are provided in \cref{tab:arch_NFs_feature,tab:arch_NFs_s1_t1,tab:arch_NFs_s2_t2}. Mirroring the GMM net, a nugget of $10^{-27}$ is added to \cref{eq:composing} to prevent numerical underflow.

\begin{table}[htbp]
    \centering
    \caption{Architecture for the feature net within the NFs.
    The first value under the `Dimension' column is used for the source location problem in \cref{sec:source} and the CES problem in \cref{sec:ces};
    the second value is used for the SIR problem in \cref{sec:sir}. }
    \begin{tabular}{cccc}
    \toprule
    Layer & Description & Dimension & Activation \\
    \midrule 
    Input & $\info_k$ & $k(N_{\design}+N_y)$ & - \\
    Hidden 1 & Dense & 256 / 128  & ReLU \\
    Hidden 2 & Dense & 256 / 128 & ReLU \\
    Hidden 3 & Dense & 256 / None & ReLU \\
    Output & Dense
    & $k(N_{\design}+N_y)$ & - \\
    \bottomrule
    \end{tabular}
    \label{tab:arch_NFs_feature}
\end{table}

\begin{table}[htbp]
    \centering
    \caption{Architecture for the ${\sf s}_1$ and ${\sf t}_1$ nets within the NFs. The first value under the `Dimension' column is used for the source location problem in \cref{sec:source} and the CES problem in \cref{sec:ces};
    the second value is used for the SIR problem in \cref{sec:sir}.}
    \begin{tabular}{cccc}
    \toprule
    Layer & Description & Dimension & Activation \\
    \midrule 
    Input & Feature($\info_k$) + $\param_1$ & $k(N_{\design}+N_y) + N_{\Param_{m,1}}$  & - \\
    Hidden 1 & Dense & 256 / 128  & ReLU \\
    Hidden 2 & Dense & 256 / 128  & ReLU \\
    Hidden 3 & Dense & 256 / 128  & ReLU \\
    Output & Dense
    & $N_{\Param_{m,2}}$ & - \\
    \bottomrule
    \end{tabular}
    \label{tab:arch_NFs_s1_t1}
\end{table}

\begin{table}[htbp]
    \centering
    \caption{Architecture for the ${\sf s}_2$ and ${\sf t}_2$ nets within the NFs. The first value under the `Dimension' column is used for the source location problem in \cref{sec:source} and the CES problem in \cref{sec:ces};
    the second value is used for the SIR problem in \cref{sec:sir}.}
    \begin{tabular}{cccc}
    \toprule
    Layer & Description & Dimension & Activation \\
    \midrule 
    Input & Feature($\info_k$) + $\tilde{\param}_2$ & $k(N_{\design}+N_y) + N_{\Param_{m,2}}$  & - \\
    Hidden 1 & Dense & 256 / 128 & ReLU \\
    Hidden 2 & Dense & 256 / 128 & ReLU \\
    Hidden 3 & Dense & 256 / 128 & ReLU \\
    Output & Dense
    & $N_{\Param_{m,1}}$ & - \\
    \bottomrule
    \end{tabular}
    \label{tab:arch_NFs_s2_t2}
\end{table}

\subsection{Neural network architectures for actor and critic}
\label{app:actor_critic}
The architectures for the {actor} (policy) and {critic} (action-value function) networks from~\cite{shen2021bayesian} are adopted in this work.
The actor $\mu_{k,w}$ is a mapping from $\info_k$ to design $\design_k$. Instead of learning separate actor networks for each stage $k$, we combine them into a single actor. The overall input to the actor network takes the form
\begin{align}
    \info^{\text{actor}}_k = [e_k, \Tilde{\info}_k], \nonumber
\end{align}
where $e_k$ is an 0-indexed one-hot encoding vector of size $N$ 
that represents the current experiment stage:
\begin{align}
    e_k=[0,\dots,0,\underbrace{1}_{k\rm{th}},0,\dots,0], \nonumber
\end{align}
and $\tilde{\info}_k$ is a vector of fixed size $(N-1)(N_{\design}+N_y)$ obtained by extending $\info_k$ with zero-padding:
\begin{align}
    \tilde{\info}_k = [ \overbrace{\design_0}^{N_{\design}},\dots,\design_{k-1},\underbrace{0,\dots,0}_{N_{\design}(N-1-k)},\overbrace{y_0}^{N_y},\dots,y_{k-1},\underbrace{0,\dots,0}_{N_y(N-1-k)} ]. \nonumber
\end{align}
The total dimension of $\info^{\text{actor}}_k$ is $N+(N-1)(N_{\design}+N_y)$. 
Similarly, the overall input to the critic network is
\begin{align}
    \info^{\text{critic}}_k = [\info^{\text{actor}}_k, \design_k], \nonumber
\end{align}
with total dimension $N+(N-1)(N_{\design}+N_y)+N_{\design}$. The output of the critic network is a scalar. 
We note that without the presence of $e_k$, it would not be possible for the actor or the critic to distinguish between %
whether the state is at stage $k$, or at a later stage but with $\design_k$ and $y_k$ actually being zero (i.e., whether zero values are padding or actual results). The architectures for the actor and critic networks are presented in \cref{tab:arch_actor,tab:arch_critic}, respectively, with `Linear mapping' in \cref{tab:arch_actor} indicating the mapping of the output value to be within the design bounds. The same actor and critic architectures are used across all numerical cases.

\begin{table}[htbp]
    \centering
    \caption{Architecture for the actor network.}
    \begin{tabular}{cccc}
    \toprule
    Layer & Description & Dimension & Activation \\
    \midrule 
    Input & $\info^{\text{actor}}_k$ & $N+(N-1)(N_{\design}+N_y)$ & - \\
    Hidden 1 & Dense & 256 & ReLU \\
    Hidden 2 & Dense & 256 & ReLU \\
    Hidden 3 & Dense & 256 & ReLU \\
    Hidden 4 & Dense & $N_{\design}$ & Sigmoid \\
    Output & Identity & $N_{\design}$ & Linear mapping \\
    \bottomrule
    \end{tabular}
    \label{tab:arch_actor}
\end{table}

\begin{table}[htbp]
    \centering
    \caption{Architecture for the critic network.}
    \begin{tabular}{cccc}
    \toprule
    Layer & Description & Dimension & Activation \\
    \midrule 
    Input & $\info^{\text{critic}}_k$ & $N+(N-1)(N_{\design}+N_y)+N_{\design}$ & - \\
    Hidden 1 & Dense & 256 & ReLU \\
    Hidden 2 & Dense & 256 & ReLU \\
    Hidden 3 & Dense & 256 & ReLU \\
    Output & Dense & 1 & - \\
    \bottomrule
    \end{tabular}
    \label{tab:arch_critic}
\end{table}

\subsection{More about the critic}
\label{app:critic}

While we introduced the critic $\tilde{Q}_k^{\pi_w}(\info_k,\design_k; \phi)$ in \cref{e:Q1,e:Q2,e:Q3} for the variational one-point reward terms $\tilde{r}_k$ and $\tilde{r}_N$ from \cref{sec:vsOED}, a corresponding critic can be formed for the one-point reward terms $\check{r}_k$ and $\check{r}_N$ from \cref{sec:one_point_utility}:
\begin{align}
\check{Q}_k^{\policy_w}(\info_k,\design_k)&= \EE_{M_{0},\Param_{m,0},\Nuis_{m,0},Z_{m,0}|\info_k}\Bigg[\EE_{Y_{k:N-1}|w,s_0,\info_k,\design_k,M_{0},\Param_{m,0},\Nuis_{m,0}}\bigg[\check{r}_k(\info_k,\design_k,{Y_k}) 
\nonumber\\
&\hspace{14em}+ \sum_{t=k+1}^{N-1} \check{r}_t{(\Info_t,\mu_{t,w}(\Info_t),Y_t)} + \check{r}_N({\Info_N})\bigg]\Bigg]
\nonumber\\
&=\EE_{M_{0},\Param_{m,0},\Nuis_{m,0},Z_{m,0}|\info_k}\Bigg[\EE_{Y_{k}|w,s_0,\info_k,\design_k,M_{0},\Param_{m,0},\Nuis_{m,0}}\bigg[ \check{r}_k(\info_k,\design_k,{Y_k}) \nonumber\\
&\hspace{18.5em}+ \check{Q}^{\policy_w}_{k+1}({\Info_{k+1},\mu_{k+1,w}(\Info_{k+1})})\bigg]\Bigg],
\nonumber\\
\check{Q}_{N}^{\policy_w}(\info_N,\cdot) &= \EE_{M_{0},\Param_{m,0},\Nuis_{m,0},Z_{m,0}|\info_N} \bigg[\check{r}_N(\info_N)\bigg], \nonumber
\end{align}
for $k=0,\ldots,N-1$ and subject to
{$\Info_{k+1}=\{\Info_k, \design_k, Y_k\}$}. 

When using the TIG formulation, these critics become
\begin{align}
    \check{Q}_{T,k}^{\pi_w}(\info_k,\design_k) &= \EE_{M_{0},\Param_{m,0},\Nuis_{m,0},Z_{m,0}|\info_k}\Bigg[\EE_{Y_{k:N-1}|w,s_0,\info_k,\design_k,M_{0},\Param_{m,0},\Nuis_{m,0}} \Bigg[ \nonumber \\
    &\hspace{3em} \alpha_M \log \frac{P(M_0|\Info_N)} {P(M_0)}  + \alpha_\Param \log \frac{p(\Param_{m,0}|\Info_N)}{p(\Param_{m,0})} + \alpha_Z \log \frac{p(Z_{m,0}|\Info_N)}{p(Z_{m,0})} \Bigg]\Bigg] \nonumber \\
    \check{Q}_{T,N}^{\policy_w}(\info_N,\cdot) &= \EE_{M_{0},\Param_{m,0},\Nuis_{m,0},Z_{m,0}|\info_N} \Bigg[\alpha_M \log \frac{P(M_0|\info_N)} {P(M_0)}  \nonumber\\
    & \hspace{3em}+ \alpha_\Param \log \frac{p(\Param_{m,0}|\info_N)}{p(\Param_{m,0})} + \alpha_Z \log \frac{p(Z_{m,0}|\info_N)}{p(Z_{m,0})}\Bigg], \nonumber
\end{align}
for $k=0,\ldots,N-1$ and subect to
{$\Info_{k+1}=\{\Info_k, \design_k, Y_k\}$}, and
\begin{align}
    \tilde{Q}_{T,k}^{\pi_w}(\info_k,\design_k;\phi) &= \EE_{M_{0},\Param_{m,0},\Nuis_{m,0},Z_{m,0}|\info_k}\Bigg[\EE_{Y_{k:N-1}|w,s_0,\info_k,\design_k,M_{0},\Param_{m,0},\Nuis_{m,0}} \Bigg[ \nonumber \\
    &\hspace{3em} \alpha_M \log \frac{q(M_0|\Info_N;\phi_M)}{P(M_0)}  + \alpha_\Param \log \frac{q(\Param_{m,0}|\Info_N;\phi_{\Param_m})}{p(\Param_{m,0})} 
    + \alpha_Z \log \frac{q(Z_{m,0}|\Info_N;\phi_{Z_m})}{p(Z_{m,0})} \Bigg]\Bigg], \nonumber\\
    \tilde{Q}_{T,N}^{\policy_w}(\info_N,\cdot; \phi) &= \EE_{M_{0},\Param_{m,0},\Nuis_{m,0},Z_{m,0}|\info_N} \Bigg[\alpha_M \log \frac{q(M_0|\info_N;\phi_M)}{P(M_0)}  \nonumber\\
    & \hspace{3em}+\alpha_\Param \log \frac{q(\Param_{m,0}|\info_N;\phi_{\Param_m})}{p(\Param_{m,0})} + \alpha_Z \log \frac{q(Z_{m,0}|\info_N;\phi_{Z_m})}{p(Z_{m,0})}\Bigg], \nonumber
\end{align}
for $k=0,\ldots,N-1$ and subject to
{$\Info_{k+1}=\{\Info_k, \design_k, Y_k\}$}. 

Similarly, when using the IIG formulation, these critics become
\begin{align}
    \check{Q}_{I,k}^{\pi_w}(\info_k,\design_k) &= \EE_{M_{0},\Param_{m,0},\Nuis_{m,0},Z_{m,0}|\info_k}\Bigg[\EE_{Y_{k:N-1}|w,s_0,\info_k,\design_k,M_{0},\Param_{m,0},\Nuis_{m,0}} \Bigg[ \nonumber \\
    &\hspace{3em} \alpha_M \log \frac{P(M_0|\Info_{k+1})} {P(M_0|\info_k)}  + \alpha_\Param \log \frac{p(\Param_{m,0}|\Info_{k+1})}{p(\Param_{m,0}|\info_k)} + \alpha_Z \log \frac{p(Z_{m,0}|\Info_{k+1})}{p(Z_{m,0}|\info_k)}
    \nonumber\\ 
    &\hspace{3em}+ \sum_{t=k+1}^{N-1}
    \alpha_M \log \frac{P(M_0|\Info_{t+1})} {P(M_0|\Info_t)}  + \alpha_\Param \log \frac{p(\Param_{m,0}|\Info_{t+1})}{p(\Param_{m,0}|\Info_t)} + \alpha_Z \log \frac{p(Z_{m,0}|\Info_{t+1})}{p(Z_{m,0}|\Info_t)} \Bigg]\Bigg], \nonumber \\
    &= \EE_{M_{0},\Param_{m,0},\Nuis_{m,0},Z_{m,0}|\info_k}\Bigg[\EE_{Y_{k:N-1}|w,s_0,\info_k,\design_k,M_{0},\Param_{m,0},\Nuis_{m,0}} \Bigg[ \nonumber \\
    &\hspace{3em} \alpha_M \log \frac{P(M_0|\Info_{N})} {P(M_0|\info_k)}  + \alpha_\Param \log \frac{p(\Param_{m,0}|\Info_{N})}{p(\Param_{m,0}|\info_k)} + \alpha_Z \log \frac{p(Z_{m,0}|\Info_{N})}{p(Z_{m,0}|\info_k)}
    \Bigg]\Bigg], \nonumber \\
    \check{Q}_{I,N}^{\pi_w}(\info_k,\cdot) &= 0, \nonumber
\end{align}
for $k=0,\ldots,N-1$ and subject to
{$\Info_{k+1}=\{\Info_k, \design_k, Y_k\}$}, and
\begin{align}
    \tilde{Q}_{I,k}^{\pi_w}(\info_k,\design_k;\phi) &= \EE_{M_{0},\Param_{m,0},\Nuis_{m,0},Z_{m,0}|\info_k}\Bigg[\EE_{Y_{k:N-1}|w,s_0,\info_k,\design_k,M_{0},\Param_{m,0},\Nuis_{m,0}}\Bigg[ \nonumber \\
    &\hspace{-3em} \alpha_M \log \frac{q(M_0|\Info_{k+1};\phi_M)}{q(M_0|\info_k;\phi_M)}  + \alpha_\Param \log \frac{q(\Param_{m,0}|\Info_{k+1};\phi_{\Param_m})}{q(\Param_{m,0}|\info_k;\phi_{\Param_{m,0}})} 
    + \alpha_Z \log \frac{q(Z_{m,0}|\Info_{k+1};\phi_{Z_m})}{q(Z_{m,0}|\info_k;\phi_{Z_m})}  \nonumber\\
    &\hspace{-3em}+ \sum_{t=k+1}^{N-1}
    \alpha_M \log \frac{q(M_0|\Info_{t+1};\phi_{M})} {q(M_0|\Info_t;\phi_{M})}  + \alpha_\Param \log \frac{q(\Param_{m,0}|\Info_{t+1};\phi_{\Param_m})}{q(\Param_{m,0}|\Info_t;\phi_{\Param_m})} + \alpha_Z \log \frac{q(Z_{m,0}|\Info_{t+1};\phi_{Z_m})}{q(Z_{m,0}|\Info_t;\phi_{Z_m})} \Bigg]\Bigg], \nonumber \\
    &= \EE_{M_{0},\Param_{m,0},\Nuis_{m,0},Z_{m,0}|\info_k}\Bigg[\EE_{Y_{k:N-1}|w,s_0,\info_k,\design_k,M_{0},\Param_{m,0},\Nuis_{m,0}}\Bigg[ \nonumber \\
    &\hspace{-3em} \alpha_M \log \frac{q(M_0|\Info_{N};\phi_M)}{q(M_0|\info_k;\phi_M)}  + \alpha_\Param \log \frac{q(\Param_{m,0}|\Info_{N};\phi_{\Param_m})}{q(\Param_{m,0}|\info_k;\phi_{\Param_{m,0}})} 
    + \alpha_Z \log \frac{q(Z_{m,0}|\Info_{N};\phi_{Z_m})}{q(Z_{m,0}|\info_k;\phi_{Z_m})}  
    \Bigg]\Bigg], \nonumber \\
    \tilde{Q}_{I,N}^{\pi_w}(\info_k,\cdot;\phi) &= 0, \nonumber
\end{align}
for $k=0,\ldots,N-1$ and subject to
{$\Info_{k+1}=\{\Info_k, \design_k, Y_k\}$}. 

\begin{remark}
The differences between the one-point-TIG and one-point-IIG critics are
\begin{align}
    \check{Q}_{T,k}^{\pi_w}(\info_k,\design_k) - \check{Q}_{I,k}^{\pi_w}(\info_k,\design_k) &= \EE_{M_{0},\Param_{m,0},\Nuis_{m,0},Z_{m,0}|\info_k}\Bigg[\EE_{Y_{k:N-1}|w,s_0,\info_k,\design_k,M_{0},\Param_{m,0},\Nuis_{m,0}} \Bigg[ \nonumber \\
    &\hspace{3em} \alpha_M \log \frac{P(M_0|\info_k)} {P(M_0)}  + \alpha_\Param \log \frac{p(\Param_{m,0}|\info_k)}{p(\Param_{m,0})} + \alpha_Z \log \frac{p(Z_{m,0}|\info_k)}{p(Z_{m,0})} \Bigg]\Bigg] \nonumber \\
    &= \EE_{M_0,\Param_{m,0},\Nuis_{m,0},Z_{m,0}|\info_k} \Bigg[ \nonumber \\
    &\hspace{3em} \alpha_M \log \frac{P(M_0|\info_k)} {P(M_0)}  + \alpha_\Param \log \frac{p(\Param_{m,0}|\info_k)}{p(\Param_{m,0})} + \alpha_Z \log \frac{p(Z_{m,0}|\info_k)}{p(Z_{m,0})} \Bigg], \nonumber\\
    \check{Q}_{T,N}^{\pi_w}(\info_N,\cdot) - \check{Q}_{I,N}^{\pi_w}(\info_N,\cdot) &= \EE_{M_{0},\Param_{m,0},\Nuis_{m,0},Z_{m,0}|\info_N} \Bigg[  \nonumber\\
    & \hspace{3em} \alpha_M \log \frac{P(M_0|\info_N)} {P(M_0)}+ \alpha_\Param \log \frac{p(\Param_{m,0}|\info_N)}{p(\Param_{m,0})} + \alpha_Z \log \frac{p(Z_{m,0}|\info_N)}{p(Z_{m,0})}\Bigg], \nonumber
\end{align}
for $k=0,\ldots,N-1$ and subject to
{$\Info_{k+1}=\{\Info_k, \design_k, Y_k\}$}. 
Notably, all difference expressions are 
independent of $\design_k$. When using the one-point reward formulations, the same policy gradient expression as \cref{eq:policy_gradient} will emerge but with  $\nabla_{\design_k} \check{Q}_{k}^{\pi_w}(\info_k,\design_k)$. Since the difference of the above critics do not depend on $\design_k$, $\nabla_{\design_k} \check{Q}_{k}^{\pi_w}(\info_k,\design_k)$ will be identical for one-point-TIG and one-point-IIG, and hence their policy gradients will also be identical.
\end{remark}

\begin{remark} The differences between the one-point-TIG and variational one-point-TIG critics
are
\begin{align}
   & \check{Q}_{T,k}^{\pi_w}(\info_k,\design_k) - \tilde{Q}_{T,k}^{\pi_w}(\info_k,\design_k;\phi) \nonumber \\
   &= \EE_{M_{0},\Param_{m,0},\Nuis_{m,0},Z_{m,0}|\info_k}\Bigg[\EE_{Y_{k:N-1}|w,s_0,\info_k,\design_k,M_{0},\Param_{m,0},\Nuis_{m,0}} \Bigg[ \nonumber \\
    &\hspace{3em} \alpha_M \log \frac{P(M_0|\Info_N)} {q(M_0|\Info_N;\phi_M)}  + \alpha_\Param \log \frac{p(\Param_{m,0}|\Info_N)}{q(\Param_{m,0}|\Info_N;\phi_{\Param_m})} + \alpha_Z \log \frac{p(Z_{m,0}|\Info_N)}{q(Z_{m,0}|\Info_N;\phi_{Z_m})} \Bigg]\Bigg] \nonumber \\
    &= \EE_{\Info_N|w,s_0,\info_k,\design_k} \Bigg[ \alpha_M \DKL\(\, P_{M_0|\Info_N}\,||\,q_{M_0|\Info_N; \phi_M} \,\) \nonumber \\
    &\hspace{3em} + \EE_{M_0|\Info_N} \Bigg[ \alpha_\Param \DKL\(\, p_{\Param_{m,0}|\Info_N}\,||\,q_{\Param_{m,0}|\Info_N;\phi_{\Param_m}} \,\) + \alpha_Z \DKL\(\, p_{Z_{m,0}|\Info_N}\,||\,q_{Z_{m,0}|\Info_N;\phi_{Z_{m}}} \,\) \Bigg] \Bigg], \nonumber \\
    & \check{Q}_{T,N}^{\pi_w}(\info_N,\cdot) - \tilde{Q}_{T,N}^{\pi_w}(\info_N,\cdot;\phi) \nonumber \\
   &= \EE_{M_{0},\Param_{m,0},\Nuis_{m,0},Z_{m,0}|\info_N}\Bigg[ \nonumber \\
    &\hspace{3em} \alpha_M \log \frac{P(M_0|\info_N)} {q(M_0|\info_N;\phi_M)}  + \alpha_\Param \log \frac{p(\Param_{m,0}|\info_N)}{q(\Param_{m,0}|\info_N;\phi_{\Param_m})} + \alpha_Z \log \frac{p(Z_{m,0}|\info_N)}{q(Z_{m,0}|\info_N;\phi_{Z_m})} \Bigg] \nonumber \\
    &= \alpha_M \DKL\(\, P_{M_0|\info_N}\,||\,q_{M_0|\info_N; \phi_M} \,\) \nonumber \\
    &\hspace{3em} + \EE_{M_0|\info_N} \Bigg[ \alpha_\Param \DKL\(\, p_{\Param_{m,0}|\info_N}\,||\,q_{\Param_{m,0}|\info_N;\phi_{\Param_m}} \,\) + \alpha_Z \DKL\(\, p_{Z_{m,0}|\info_N}\,||\,q_{Z_{m,0}|\info_N; \phi_{Z_{m}}} \,\) \Bigg], \nonumber 
\end{align}
for $k=0,\ldots,N-1$ and subject to
{$\Info_{k+1}=\{\Info_k, \design_k, Y_k\}$}. 
Notably, all difference expressions are expectations of weighted sum of KL divergence terms with non-negative weights. The difference expressions are zero if and only if the variational posterior approximations $q(\cdot|\Info_N;\phi_{(\cdot)})$ are equal to the true posteriors $p(\cdot|\Info_N)$ (except the trivial case when $\alpha_M=\alpha_{\Param}=\alpha_Z=0$). Hence, $\tilde{Q}_{T,k}^{\pi_w}(\info_k,\design_k;\phi)$ forms a lower bound to $\check{Q}_{T,k}^{\pi_w}(\info_k,\design_k)$ for all $k$, and learning an accurate variational posterior approximation would help reduce the error in the critic.
\end{remark}

\begin{remark} The differences between the one-point-IIG and variational one-point-IIG critics
are
\begin{align}
   & \check{Q}_{I,k}^{\pi_w}(\info_k,\design_k) - \tilde{Q}_{I,k}^{\pi_w}(\info_k,\design_k;\phi) \nonumber \\
   &= \EE_{M_{0},\Param_{m,0},\Nuis_{m,0},Z_{m,0}|\info_k}\Bigg[\EE_{Y_{k:N-1}|w,s_0,\info_k,\design_k,M_{0},\Param_{m,0},\Nuis_{m,0}} \Bigg[ \nonumber \\
    &\hspace{3em} \alpha_M \log \frac{P(M_0|\Info_N)} {q(M_0|\Info_N;\phi_M)}  + \alpha_\Param \log \frac{p(\Param_{m,0}|\Info_N)}{q(\Param_{m,0}|\Info_N;\phi_{\Param_m})} + \alpha_Z \log \frac{p(Z_{m,0}|\Info_N)}{q(Z_{m,0}|\Info_N; \phi_{Z_m})} \nonumber \\
    &\hspace{3em} - \alpha_M \log \frac{P(M_0|\info_k)} {q(M_0|\info_k;\phi_M)}  - \alpha_\Param \log \frac{p(\Param_{m,0}|\info_k)}{q(\Param_{m,0}|\info_k;\phi_{\Param_m})} - \alpha_Z \log \frac{p(Z_{m,0}|\info_k)}{q(Z_{m,0}|\info_k;\phi_{Z_m})} \Bigg]\Bigg] \nonumber \\
    &= \EE_{\Info_N|w,s_0,\info_k,\design_k} \Bigg[ \alpha_M \DKL\( \, P_{M_0|\Info_N}\,||\,q_{M_0|\Info_N;\phi_M} \, \) \nonumber \\
    &\hspace{3em} + \EE_{M_0|\Info_N} \Bigg[ \alpha_\Param \DKL \( \, p_{\Param_{m,0}|\Info_N}\,||\,q_{\Param_{m,0}|\Info_N;\phi_{\Param_m}} \, \) + \alpha_Z \DKL\(\, p_{Z_{m,0}|\Info_N}\,||\,q_{Z_{m,0}|\Info_N;\phi_{Z_{m}}} \,\)  \Bigg] \nonumber \\
    &\hspace{5.5em} - \alpha_M \DKL\(\, P_{M_0|\info_k}\,||\,q_{M_0|\info_k;\phi_M} \,\) \nonumber \\
    &\hspace{3em}- \EE_{M_0|\info_k} \Bigg[ \alpha_\Param \DKL\(\, p_{\Param_{m,0}|\info_k}\,||\,q_{\Param_{m,0}|\info_k;\phi_{\Param_m}} \,\) + \alpha_Z \DKL\( \, p_{Z_{m,0}|\info_k}\,||\,q_{Z_{m,0}|\info_k;\phi_{Z_m}} \, \) \Bigg], \nonumber\\
    & \check{Q}_{I,N}^{\pi_w}(\info_N,\cdot) - \tilde{Q}_{I,N}^{\pi_w}(\info_N,\cdot;\phi) = 0, \nonumber
\end{align}
for $k=0,\ldots,N-1$ and subject to
{$\Info_{k+1}=\{\Info_k, \design_k, Y_k\}$}. 
Applying triangle inequality, these difference expressions become bounded by  
\begin{align}
    &\abs{\check{Q}_{I,k}^{\pi_w}(\info_k,\design_k) - \tilde{Q}_{I,k}^{\pi_w}(\info_k,\design_k;\phi)} \nonumber \\
    &\leq \EE_{\Info_N|w,s_0,\info_k,\design_k} \Bigg[ \alpha_M \DKL\(\, P_{M_0|\Info_N}\,||\,q_{M_0|\Info_N;\phi_M} \,\) \nonumber \\
    &\hspace{3em} + \EE_{M_0|\Info_N} \Bigg[ \alpha_\Param \DKL\(\, p_{\Param_{m,0}|\Info_N}\,||\,q_{\Param_{m,0}|\Info_N;\phi_{\Param_m}} \,\) + \alpha_Z \DKL\(\, p_{Z_{m,0}|\Info_N}\,||\,q_{Z_{m,0}|\Info_N;\phi_{Z_m}} \,\) \Bigg] \Bigg] \nonumber \\
    &\hspace{5.5em} + \alpha_M \DKL\(\, P_{M_0|\info_k}\,||\,q_{M_0|\info_k;\phi_M} \,\) \nonumber \\
    &\hspace{3em} + \EE_{M_0|\info_k} \Bigg[ \alpha_\Param \DKL\(\, p_{\Param_{m,0}|\info_k}\,||\,q_{\Param_{m,0}|\info_k;\phi_{\Param_m}} \,\) + \alpha_Z \DKL\(\, p_{Z_{m,0}|\info_k}\,||\,q_{Z_{m,0}|\info_k;\phi_{Z_m}} \,\) \Bigg], \nonumber\\
    & \abs{\check{Q}_{I,N}^{\pi_w}(\info_N,\cdot) - \tilde{Q}_{I,N}^{\pi_w}(\info_N,\cdot;\phi)} = 0, \nonumber
\end{align}
for $k=0,\ldots,N-1$ and subject to
{$\Info_{k+1}=\{\Info_k, \design_k, Y_k\}$}. 
Therefore, the error of the critic is contributed from both the error in the final variational posterior $q(\cdot|\Info_N;\phi_{(\cdot)})$ and the errors in the intermediate variational posteriors $q(\cdot|\Info_k;\phi_{(\cdot)})$. The critic errors become zero if 
all the variational posteriors equal their corresponding true posteriors (except the trivial case when $\alpha_M=\alpha_{\Param}=\alpha_Z=0$).
\end{remark}

\subsubsection{Hyperparameter tuning}
\label{app:hyperparameter_tuning}

Our main strategy for hyperparameter tuning is to start with a relatively large hyperparameter value and gradually decrease it.

For optimizing the GMM and NFs approximate posteriors for model indicator and PoIs, we start with the initial learning rate of $10^{-3}$ and an exponential learning rate decay rate $0.9999$. 
For optimizing the critic network, we use an initial learning rate of $10^{-3}$ and a learning rate decay rate of $0.9999$ across all numerical cases. Both posterior approximation and critic network optimizations are updated 5 steps (i.e., applying gradient ascent 5 times) within each outer iteration. Making too many update steps within each outer iteration may result in overestimation of the value function and adversely affect the policy search \cite{hasselt2010double}. 

For optimizing the actor network, we use a learning rate decay rate of $0.9999$. However, the choice of the initial learning rate is more problem-dependent. Typically, we start with an initial learning rate of $10^{-3}$ and gradually decrease it to $5\times 10^{-4}$ or $2\times 10^{-4}$ if divergence occurs. For the IIG formulation, an initial learning rate of $10^{-3}$ works well. For TIG, a smaller learning rate is generally required. This is likely due to the slower propagation of critic values from $N$ to the earlier $k$ stages under TIG, and a large learning rate may induce divergence in the early iterations of training.

For other hyperparameters including the number of updates $n_{\text{update}}$, number of sample trajectories $n_{\text{traj}}$, batch size $n_{\text{batch}}$, and replay buffer size $n_{\text{buffer}}$, a number of combinations are tested to identify the optimal setting. Their values are specified for each numerical case in \cref{app:exp_details}.

\section{Numerical experiment details} 
\label{app:exp_details}

\subsection{Prior contrastive estimator}
\label{app:reward_one_point}

The PCE for batch (non-sequential) OED is introduced in \cite{foster2020unified}:
\begin{align}
  U^{\text{PCE}}(\design) =  \mathbb{E}_{Y_0 \vert \Param_0,\design} \mathbb{E}_{\Param_0} \mathbb{E}_ {\Param_{1:L}} \left [ \log \frac{p(Y_0 \vert \Param_0,\design)} {\frac{1}{L+1} \sum_{j=1}^{L+1} p(Y_0 \vert \Param_{j},\design)} \right ],
\nonumber
\end{align}
where the expectation is over $\Param_0, Y_0|\design \sim p(\Param, Y|\design)$
and $\Param_{1:L} \stackrel{\text{iid}}{\sim}
p(\Param)$, and the subscripts here in the non-sequential setting represent multi-sample indexing.  
One can show that $U^{\text{PCE}}(\design)$ is a lower bound to the mutual information between $Y$ and $\Param$, $\mathcal{I}(Y ; \Param|\design)$, (i.e., the EIG of $\Param$)---that is, $U^{\text{PCE}}(\design) \leq \mathcal{I}(Y ; \Param|\design)$---for any $L>0$, and the bound becomes tight as $L \to \infty$~\citep[Theorem~1]{foster2020unified}. 
However, the expectation in $U^{\text{PCE}}$ is generally intractable to evaluate, and $U^{\text{PCE}}$ needs to be estimated numerically, for example through MC:
\begin{align}
  U^{\text{PCE}}(\design) \approx \frac{1}{n_{\text{out}}} \sum_{i=1}^{n_{\text{out}}} \log \frac{p(y_0^{(i)} \vert \param_0^{(i)},\design)} {\frac{1}{L+1} \sum_{j=1}^{L+1} p(y_0^{(i)} \vert \param_{j}^{(i)},\design)},
\nonumber
\end{align}
where $(y_0^{(i)}, \param_0^{(i)}, \param_{1:L}^{(i)})$ are $n_{\text{out}}$ independent and identically distributed (i.i.d.) realizations of $(Y_0, \Param_0,
\Param_{1:L})$.

For sequential OED, the PCE becomes~\cite{foster2021deep}:
\begin{align}
  U^{\text{sPCE}}(\policy) = \EE_{Y_{0:N-1}|\policy, \Param_0} \EE_{\Param_0} \EE_{\Param_{1:L}} \left[\log
    \frac{\pdf(Y_{0:N-1}|\Param_0,\design_{0:N-1})}{\frac{1}{L+1}\sum_{j=1}^{L+1}
      \pdf(Y_{0:N-1}|\Param_{j},\design_{0:N-1})} \right]
   \nonumber
\end{align}
where $\Param_0$ is the data-generating parameter for
$\design_{0:N-1},{Y_{0:N-1}}$ with
$\design_{k}=\mu_k({\Info_k})$ following the given policy $\policy$; the
subscripts for $Y$ and $\design$ refer to the experiment (stage)
index, while those for $\Param$ refer to the multi-sample index.
One can show that $U^{\text{sPCE}}(\policy) \leq \mathcal{I}(Y_{0:N-1};\Param\vert \policy)$ for any $\policy$ and $L>0$, and the bound becomes tight as $L \to \infty$ \citep[Theorem~1]{foster2021deep}. Similarly, $U^{\text{sPCE}}$ can be estimated using MC:
\begin{align}
  U^{\text{sPCE}}(\policy) \approx \frac{1}{n_{\text{out}}} \sum_{i=1}^{n_{\text{out}}} \log
    \frac{\pdf(y_{0:N-1}^{(i)}|\param_0^{(i)},\design_{0:N-1})}{\frac{1}{L+1}\sum_{j=1}^{L+1}
      \pdf(y_{0:N-1}^{(i)}|\param_{j}^{(i)},\design_{0:N-1})},
   \nonumber
\end{align}
where $(y_{0:N-1}^{(i)}, \param_0^{(i)}, \param_{1:L}^{(i)})$ are $n_{\text{out}}$ i.i.d. realizations of $(Y_{0:N-1}, \Param_0,
\Param_{1:L})$.
We use this estimator for evaluating policies from `OED for PoIs' in this paper when applicable.

From the expressions above, we can see that PCE cannot be used when nuisance parameters $\Nuis$ are present since the term $\pdf(y_{0:N-1}|\param_0,\design_{0:N-1})$ 
would be intractable as it needs to marginalize out $\Nuis$.
Similarly, PCE cannot be used for `OED for QoIs' since in that scenario, the role of $\Param$ is replaced by $Z$ and $\pdf(y_{0:N-1}|\param_0,\design_{0:N-1})$ becomes $\pdf(y_{0:N-1}|z_0,\design_{0:N-1})$, which is intractable to evaluate. Lastly, PCE cannot be used for implicit likelihood cases since $\pdf(y_{0:N-1}|\param_0,\design_{0:N-1})$ would not be accessible. 

We can write an analogous PCE expression for model indicator by replacing $\Param$ with $M$ and use the exact expression for the `marginal-likelihood' term in the denominator:
\begin{align}
  U^{\text{sPCE},M}(\policy) &= \EE_{Y_{0:N-1}|\policy, M_0} \EE_{M_0} \left[\log
    \frac{\pdf(Y_{0:N-1}|M_0,\design_{0:N-1})}{\frac{1}{|\mathcal{M}_m|}\sum_{j=1}^{|\mathcal{M}_m|}
      P(m_j) \pdf(Y_{0:N-1}|m_{j},\design_{0:N-1})} \right]
   \nonumber\\
 &\approx \frac{1}{n_{\text{out}}} \sum_{i=1}^{n_{\text{out}}} \log
    \frac{\pdf(y_{0:N-1}^{(i)}|m_0^{(i)},\design_{0:N-1})}{\frac{1}{|\mathcal{M}_m|}\sum_{j=1}^{|\mathcal{M}_m|}
      P(m_j)\pdf(y_{0:N-1}^{(i)}|m_{j},\design_{0:N-1})},
   \nonumber
\end{align}
where $M_0$ is the data-generating parameter for
$\design_{0:N-1},{Y_{0:N-1}}$ with
$\design_{k}=\mu_k({\Info_k})$ following the given policy $\policy$, and $(y_{0:N-1}^{(i)}, m_0^{(i)})$ are $n_{\text{out}}$ i.i.d. realizations of $(Y_{0:N-1}, M_0)$.
The `likelihood' term can be estimated, for example, via
\begin{align}
    \pdf(y_{0:N-1}|m,\design_{0:N-1}) &= \iint p(\param_m,\nuis_m|m) \, p(y_{0:N-1}|m,\param_m,\nuis_m,\design_{0:N-1}) \,\text{d}\param_m \,\text{d}\nuis_m \nonumber \\
    &\approx \frac{1}{L} \sum_{l=1}^{L} p(y_{0:N-1}|m,\param_{m}^{(l)},\nuis_{m}^{(l)},\design_{0:N-1}), \nonumber
\end{align}
where $\param_{m}^{(l)},\nuis_{m}^{(l)} \sim p(\param_m,\nuis_m|m)$. We use this estimator for evaluating policies from `OED for model indicator' in this paper.

\subsection{Case 1: source location finding}
\label{app:source}

\paragraph{Hyperparameters} \Cref{tab:hyper_source_uni,tab:hyper_source_multi} present the hyperparameter settings for Case 1a and 1b, respectively. In both cases, for the `Linear mapping' in the GMM net, we map the PoI GMM mean to $[-6, 6]$, PoI GMM standard deviation to $[10^{-5}, 1]$, QoI GMM mean to $[-6, 6]$, and QoI GMM standard deviation to $[10^{-5}, 2]$. The truncated normal distribution is not used.

\begin{table}[htbp]
    \centering
    \caption{Case 1a. Hyperparameter settings. In the table, `lr' stands for `learning rate'.}
    \begin{tabular}{ccccc}
    \toprule
     & vsOED-G-T & vsOED-G-I & vsOED-N-T & vsOED-N-I\\
    \midrule 
     $n_{\text{updates}}$ & $10001$ & $10001$ & $10001$ & $10001$ \\
     $n_{\text{traj}}$ & $1000$ & $1000$ & $1000$ & $1000$ \\
     $n_{\text{batch}}$ & $10000$ & $10000$ & $10000$ & $10000$ \\
     $a_{\phi,0}$ & $10^{-3}$ & $10^{-3}$ & $10^{-3}$ & $10^{-3}$ \\
     $a_{\phi}$ decay & $0.9999$ & $0.9999$ & $0.9999$ & $0.9999$ \\
     \# of $\phi$ updates per $l$-iteration & $5$ & $5$ & $5$ & $5$  \\
     $n_{\text{mixture}}$ & $8$ & $8$ & N/A & N/A \\
     $n_{\text{trans}}$ & N/A & N/A & $4$ & $4$ \\
     $a_{w,0}$ & $5\times 10^{-4}$ & $10^{-3}$ & $10^{-3}$  & $ 10^{-3}$  \\
     $a_{w}$ decay & $0.9999$ & $0.9999$ & $0.9999$ & $0.9999$ \\
     Initial critic lr  & $10^{-3}$ & $10^{-3}$ & $10^{-3}$& $10^{-3}$ \\
     Critic lr decay & $0.9999$ & $0.9999$ & $0.9999$ & $0.9999$\\
     \# of $\nu$ updates per $l$-iteration & $5$ & $5$ & $5$ & $5$  \\
     Max buffer size & $10^{6}$ & $10^{6}$ & $10^{6}$ & $10^{6}$ \\
     $\gamma$ & $1$ & $0.9$ &  $1$ & $0.9$ \\
     Initial $\sigma_{k,\text{explore}}$     & $0.5$ & $0.5$ & $0.5$  & $0.5$ \\
     $\sigma_{k,\text{explore}}$ decay & $0.9999$ & $0.9999$ & $0.9999$ & $0.9999$ \\
     Target network lr & $0.1$ & $0.1$ & $0.1$ & $0.1$ \\
    \bottomrule
    \end{tabular}
    \label{tab:hyper_source_uni}
\end{table}

\begin{table}[htbp]
    \centering
    \caption{Case 1b. Hyperparameter settings. In the table, `lr' stands for `learning rate'.}
    \begin{tabular}{ccc}
    \toprule
     & vsOED-G-T & vsOED-G-I\\
    \midrule 
     $n_{\text{update}}$ & $10001$ & $10001$ \\
     $n_{\text{traj}}$ & $1000$ & $1000$ \\
     $n_{\text{batch}}$ & $10000$ & $10000$ \\
     $a_{\phi,0}$ & $10^{-3}$ & $10^{-3}$  \\
     $a_{\phi}$ decay & $0.9999$ & $0.9999$  \\
     \# of $\phi$ updates per $l$-iteration & $5$ & $5$  \\
     $n_{\text{mixture}}$ & $8$ & $8$  \\
     $a_{w,0}$ & $2\times10^{-4}$ & $10^{-3}$  \\
     $a_{w}$ decay & $0.9999$ & $0.9999$  \\
     Initial critic lr & $10^{-3}$ & $10^{-3}$  \\
     Critic lr decay & $0.9999$ & $0.9999$  \\
     \# of $\nu$ updates per $l$-iteration & $5$ & $5$ \\
     Max buffer size & $10^{6}$ & $10^{6}$  \\
     $\gamma$ & $1$ & $0.9$  \\
     Initial $\sigma_{k,\text{explore}}$ & $0.5$ & $0.5$  \\
     $\sigma_{k,\text{explore}}$ decay & $0.9999$ & $0.9999$  \\
     Target network lr & $0.1$ & $0.1$ \\
    \bottomrule
    \end{tabular}
    \label{tab:hyper_source_multi}
\end{table}

\begin{figure}[htbp]
  \centering
  \subfloat[OED for PoIs]{\label{fig:source_uni_poi_EU_vs_update}\includegraphics[width=0.49\linewidth]{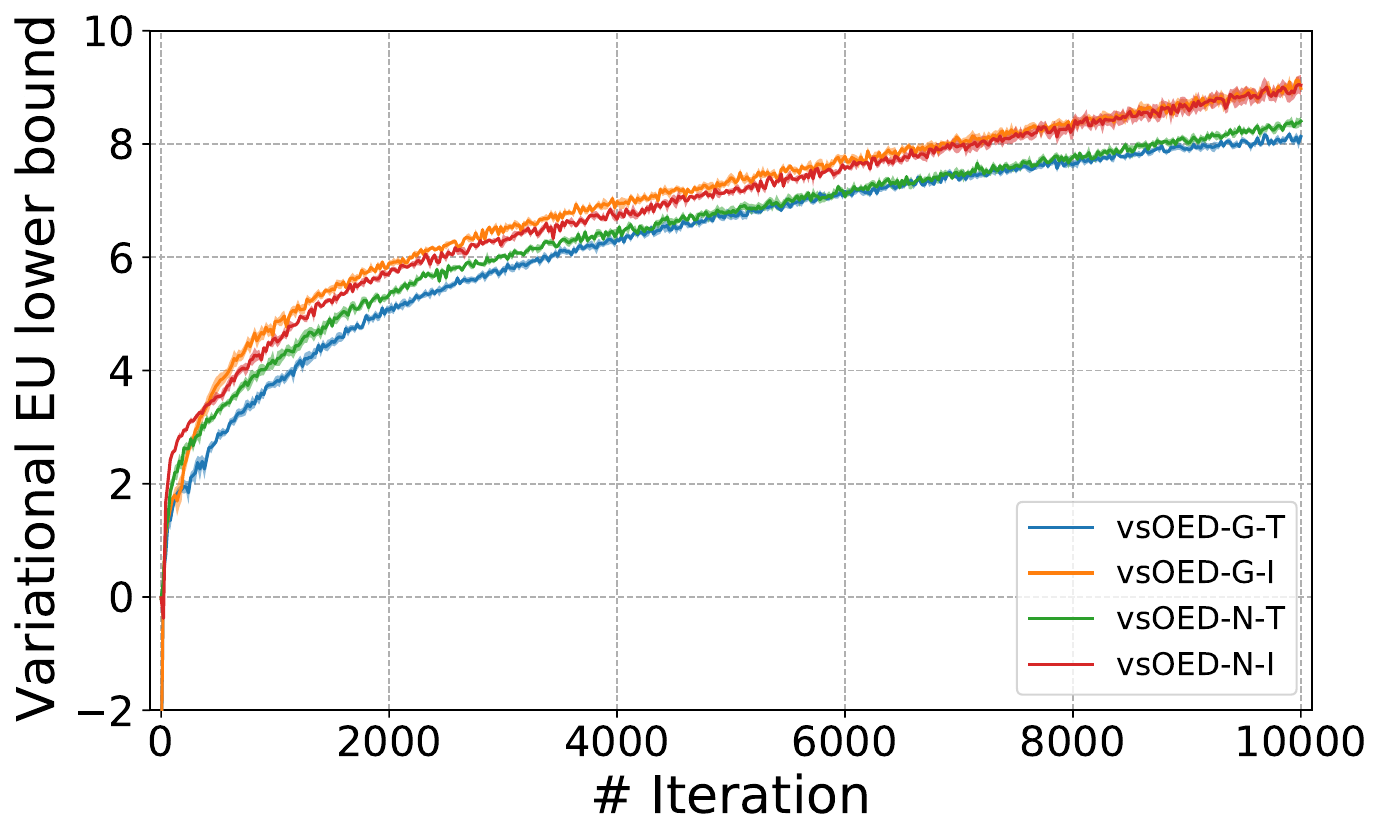}}
  \subfloat[OED for QoIs]{\label{fig:source_uni_goal_EU_vs_update}\includegraphics[width=0.49\linewidth]{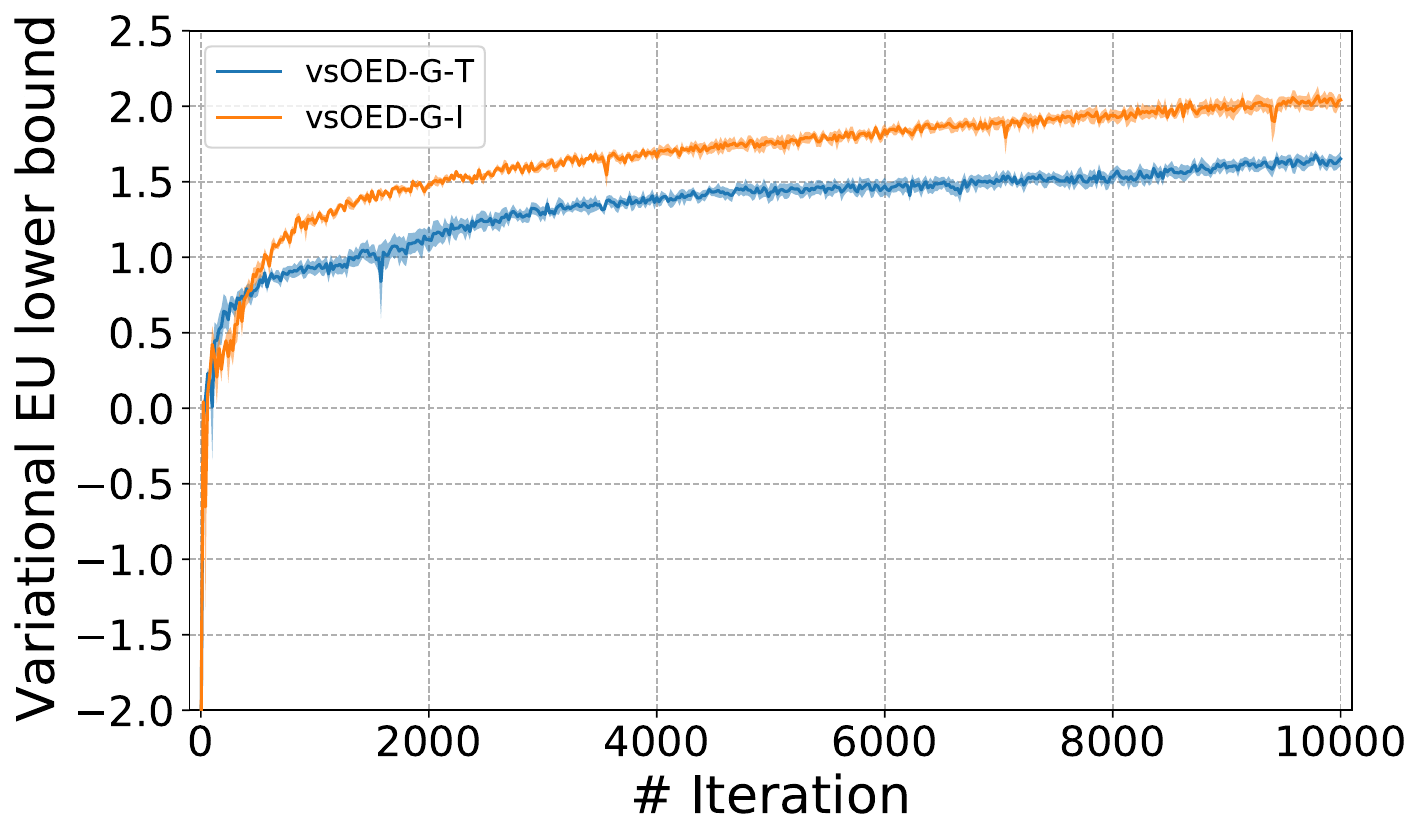}}
  \caption{Case 1a. Training history of average $\tilde{U}$ over four training replicates for $N=30$. The shaded regions represent standard error. 
  }
  \label{fig:source_EU_vs_update}
\end{figure}

\begin{figure}[htbp]
  \centering
  \includegraphics[width=0.6\linewidth]{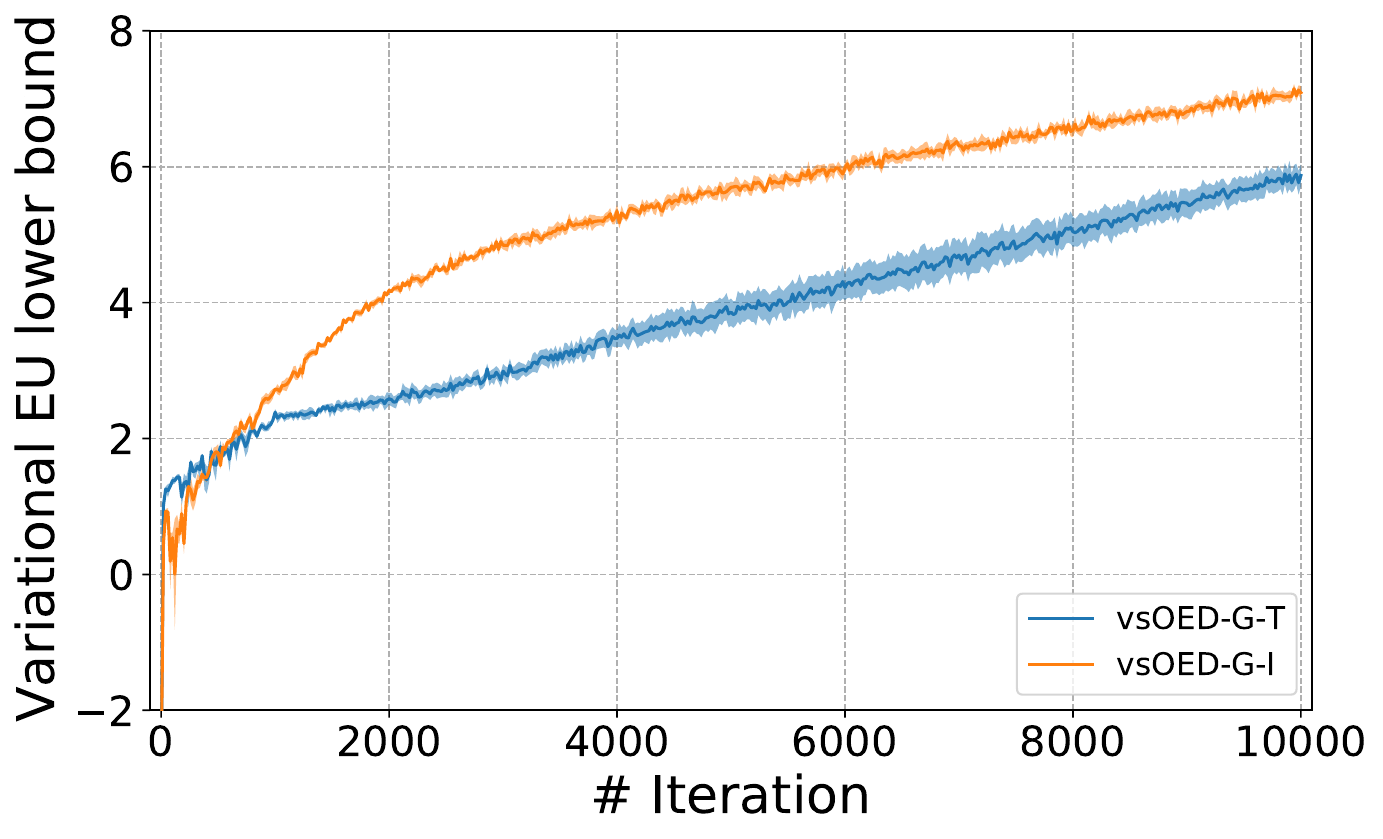}
  \caption{Case 1b. Training history of average $\tilde{U}$ over four training replicates for `OED for PoIs' and $N=30$. The shaded regions represent standard error.
  }
  \label{fig:source_multi_poi_EU_vs_update}
\end{figure}

\paragraph{Training stability} 
For Case 1b, we only illustrate training stability results for `OED for PoIs' for brevity.
\Cref{fig:source_EU_vs_update,fig:source_multi_poi_EU_vs_update}
present the training history of average $\tilde{U}$ over four training replicates for Case 1a and 1b, respectively, for $N=30$. The shaded regions represent the standard error. 
\Cref{tab:source_replicates,tab:source_multi_poi_replicates} present the expected utility and standard error ($\pm$) for four training replicates for `OED for PoIs', evaluated using PCE, for $N=30$. The last columns display the average of the four replicates along with its standard error. 
The results indicate a good level of training robustness.

\begin{table}[htbp]
    \centering
    \caption{Case 1a. Expected utility and standard error ($\pm$) for four training replicates for `OED for PoIs', evaluated using PCE, for $N=30$. }
    \begin{tabular}{ccccc|c}
    \toprule
     & Replicate 1 & Replicate 2 & Replicate 3 & Replicate 4 & Average \\
    \midrule 
    vsOED-G-T & $11.26 \pm 0.05$ & $11.09 \pm 0.05$ & $11.24 \pm 0.05$ & $11.02 \pm 0.04$ &  $11.15\pm 0.05$ \\
    vsOED-G-I & $12.50 \pm 0.04$ & $12.31 \pm 0.04$ & $12.58 \pm 0.04$ & $11.90 \pm 0.04$  & $12.32 \pm 0.13$ \\
    vsOED-N-T & $11.24 \pm 0.05$ & $11.67 \pm 0.05$ & $11.15 \pm 0.05$ & $11.59 \pm 0.05$  & $11.42 \pm 0.11$ \\
    vsOED-N-I & $12.39 \pm 0.04$ & $12.34 \pm 0.04$ & $12.16 \pm 0.04$ & $12.54 \pm 0.04$ & $12.36 \pm 0.07$ \\
    \bottomrule
    \end{tabular}
    \label{tab:source_replicates}
\end{table}

\begin{table}[htbp]
    \centering
    \caption{Case 1b. Expected utility and standard error ($\pm$) for four training replicates for `OED for PoIs', evaluated using PCE, for $N=30$. }
    \begin{tabular}{ccccc|c}
    \toprule
     & Replicate 1 & Replicate 2 & Replicate 3 & Replicate 4 & Average \\
    \midrule 
    vsOED-G-T & $9.67 \pm 0.06$ & $9.26 \pm 0.06$ & $9.06 \pm 0.06$ & $8.83 \pm 0.06$ & $9.21 \pm 0.15$ \\
    vsOED-G-I & $10.57 \pm 0.05$ & $10.09 \pm 0.05$ & $10.46 \pm 0.05$ & $10.43 \pm 0.05$ & $10.39 \pm 0.09$ \\
    \bottomrule
    \end{tabular}
    \label{tab:source_multi_poi_replicates}
\end{table}

\subsection{Case 2: constant elasticity of substitution}
\label{app:ces_hyperparam}

\paragraph{Hyperparameters} \Cref{tab:hyper_ces} presents the hyperparameter settings for this case. For the `Linear mapping' in the GMM net, we map the PoI GMM mean to $[-1, 2]$ for $\rho$ and $\beta$ and to $[-17, 19]$ for $\log{u}$, and PoI GMM standard deviation to $[10^{-5}, 3]$ for all variables. The truncated normal distribution is used on $\rho$ and $\beta$ with support $[0, 1]$.

\begin{table}[htbp]
    \centering
    \caption{Case 2. Hyperparameter settings. In the table, `lr' stands for `learning rate'.}
    \begin{tabular}{ccccc}
    \toprule
     & vsOED-G-T & vsOED-N-T \\
    \midrule 
     $n_{\text{update}}$ & $10001$ & $10001$ \\
     $n_{\text{traj}}$ & $1000$ & $1000$ \\
     $n_{\text{batch}}$ & $10000$ & $10000$ \\
     $a_{\phi,0}$ & $10^{-3}$ & $10^{-3}$\\
     $a_{\phi}$ decay & $0.9999$ & $0.9999$ \\
     \# of $\phi$ updates per $l$-iteration & $5$ & $5$ \\
     $n_{\text{mixture}}$ & $8$ & N/A \\
     $n_{\text{trans}}$ & N/A &  $4$ \\
     $a_{w,0}$ & $10^{-3}$ &  $10^{-3}$  \\
     $a_{w}$ decay & $0.9999$ & $0.9999$  \\
     Initial critic lr & $10^{-3}$ & $10^{-3}$ \\
     Critic lr decay & $0.9999$ &  $0.9999$  \\
     \# of $\nu$ updates per $l$-iteration & $5$ & $5$ \\
     Max buffer size & $10^{6}$ & $10^{6}$ \\
     $\gamma$ & $1$ & $1$ \\
     Initial $\sigma_{k,\text{explore}}$ & $5$ & $5$  \\
     $\sigma_{k,\text{explore}}$ decay & $0.9998$ & $0.9998$  \\
     Target network lr & $0.1$ &  $0.1$ \\
    \bottomrule
    \end{tabular}
    \label{tab:hyper_ces}
\end{table}

\begin{figure}[htbp]
  \centering
  \includegraphics[width=0.6\linewidth]{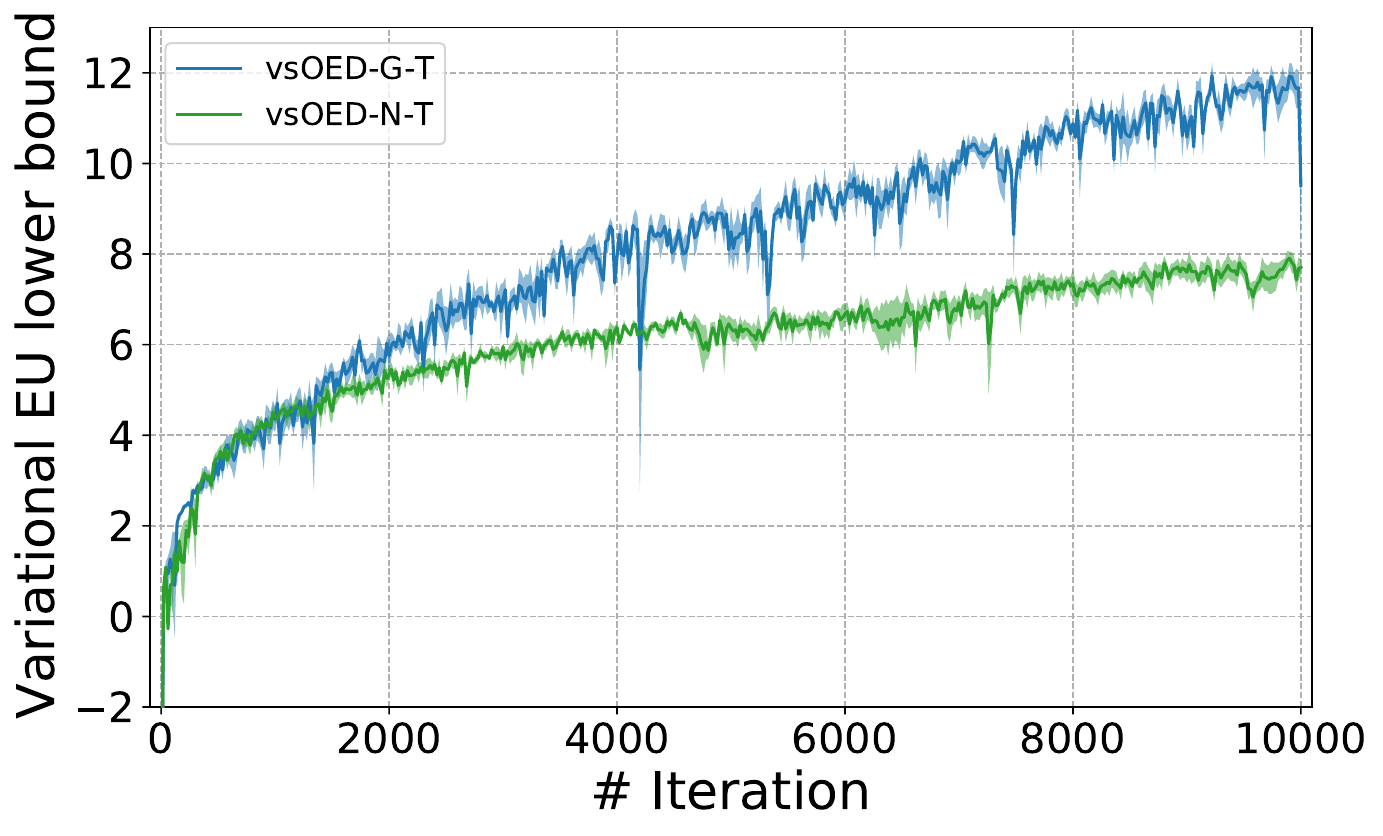}
  \caption{Case 2. Training history of average $\tilde{U}$ over four training replicates for $N=10$. The shaded regions represent the standard error.
  } \label{fig:ces_EU_vs_update}
\end{figure}

\paragraph{Training stability} 
\Cref{fig:ces_EU_vs_update} presents the training history of average $\tilde{U}$ over four training replicates for $N=10$, with the shaded regions representing the standard error. 
\Cref{tab:ces_replicates} presents the expected utility and standard error ($\pm$) for four training replicates, evaluated using PCE, for $N=10$. The last column displays the average of the four replicates along with its standard error. 
The results exhibits slightly more variation compared to Case 1, but overall remains robust.

\begin{table}[htbp]
    \centering
    \caption{Case 2. Expected utility and standard error ($\pm$) for four training replicates, evaluated using PCE, for $N=10$.}
    \begin{tabular}{ccccc|c}
    \toprule
     & Replicate 1 & Replicate 2 & Replicate 3 & Replicate 4 & Average \\
    \midrule 
    vsOED-G-T & $11.79 \pm 0.07$ & $12.34 \pm 0.06$ & $12.29 \pm 0.06$ & $11.13 \pm 0.08$ & $11.89 \pm 0.25$\\
    vsOED-N-T & $8.40 \pm 0.10$ & $9.51 \pm 0.09$ & $8.91 \pm 0.09$ & $10.30 \pm 0.08$ & $9.28 \pm 0.35$\\
    \bottomrule
    \end{tabular}
   \label{tab:ces_replicates}
\end{table}

\subsection{Case 3: SIR model for disease spread}
\label{app:sir_hyperparam}

\paragraph{Hyperparameters} 
\Cref{tab:hyper_sir} presents the hyperparameter settings for this case. For the `Linear mapping' in the GMM net, we map the PoI GMM mean to $[-6, 4]$, and PoI GMM standard deviation to $[10^{-5}, 0.5]$. The truncated normal distribution is not used.

\begin{table}[htbp]
    \centering
    \caption{Case 3. Hyperparameter settings. In the table, `lr' stands for `learning rate'.}
    \begin{tabular}{ccccc}
    \toprule
     & vsOED-G-T & vsOED-N-T \\
    \midrule 
     $n_{\text{update}}$ & $10001$ & $10001$ \\
     $n_{\text{traj}}$ & $1000$ & $1000$ \\
     $n_{\text{batch}}$ & $10000$ & $10000$ \\
     $a_{\phi,0}$ & $5\times 10^{-4}$ & $10^{-3}$ \\
     $a_{\phi}$ decay & $0.9999$ &$0.9999$\\
     \# of $\phi$ updates per $l$-iteration & $5$ & $5$\\
     $n_{\text{mixture}}$ & $8$ & N/A \\
     $n_{\text{trans}}$ & N/A & $4$ \\
     $a_{w,0}$ & $5\times 10^{-4}$ & $5\times 10^{-4}$ \\
     $a_{w}$ decay & $0.9999$ &  $0.9999$  \\
     Initial critic lr & $10^{-3}$ &  $10^{-3}$ \\
     Critic lr decay & $0.9999$ &  $0.9999$ \\
     \# of $\nu$ updates per $l$-iteration & $5$ & $5$ \\
     Max buffer size & $10^{6}$ & $10^{6}$ \\
     $\gamma$ & $1$ & $1$\\
     Initial $\sigma_{k,\text{explore}}$ & $5$ & $5$  \\
     $\sigma_{k,\text{explore}}$ decay & $0.9999$ & $0.9999$ \\
     Target network lr & $0.1$ & $0.1$ \\
    \bottomrule
    \end{tabular}
    \label{tab:hyper_sir}
\end{table}

\begin{figure}[htbp]
  \centering
  \includegraphics[width=0.6\linewidth]{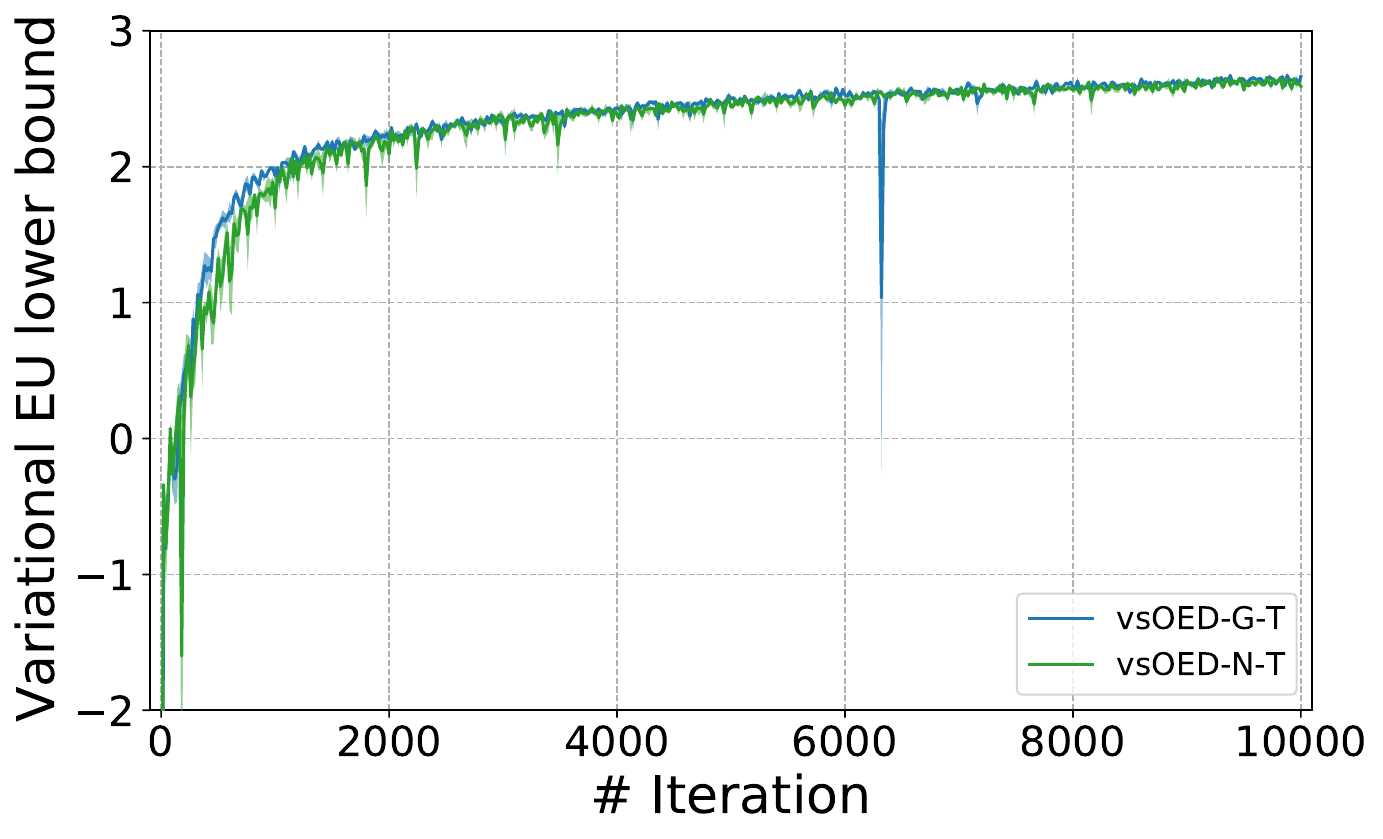}
  \caption{Case 3. Training history of average $\tilde{U}$ over four training replicates for $N=10$. The shaded regions represent the standard error.
  }
  \label{fig:sir_EU_vs_update}
\end{figure}

\paragraph{Training stability} 
\Cref{fig:sir_EU_vs_update} presents the training history of average $\tilde{U}$ over four training replicates for $N=10$, with the shaded regions representing the standard error. 
\Cref{tab:sir_replicates} presents the expected utility and standard error ($\pm$) for four training replicates, evaluated using PCE, for $N=10$. The last column displays the average of the four replicates along with its standard error. 
The training appears highly stable in this case with consistent performance across different random seeds, except for a dip in the training history for vsOED-G-T.

\begin{table}[htbp]
    \centering
    \caption{Case 3. Expected utility and standard error ($\pm$) for four training replicates, evaluated using PCE, for $N=10$. }
    \begin{tabular}{ccccc|c}
    \toprule
     & Replicate 1 & Replicate 2 & Replicate 3 & Replicate 4 & Average \\
    \midrule 
    vsOED-G-T & $4.091 \pm 0.002$ & $4.093 \pm 0.002$ & $4.090 \pm 0.001$ & $4.092 \pm 0.001$ & $4.092 \pm 0.001$\\
    vsOED-N-T & $4.097 \pm 0.002$ & $4.100 \pm 0.002$ & $4.091 \pm 0.002$ & $4.106 \pm 0.002$ & $4.099 \pm 0.003$ \\
    \bottomrule
    \end{tabular}
    \label{tab:sir_replicates}
\end{table}

\subsection{Case 4: convection-diffusion-reaction}
\label{app:convec_diff_hyperparam}

\paragraph{Hyperparameters} \Cref{tab:hyper_conv_diff} presents the hyperparameter settings for this case. 
this case. For the `Linear mapping' in the GMM net, we map the PoI GMM mean to $[-1, 2]$, PoI GMM standard deviation to $[10^{-5}, 1]$, QoI GMM mean to $[-15, 3]$, and QoI standard deviation net to $[10^{-5}, 4]$.
The truncated normal distribution is used on all PoIs with support $[0, 1]$.

\begin{table}[htbp]
    \centering
    \caption{Case 4. Hyperparameter settings. In the table, `lr' stands for `learning rate'.}
    \begin{tabular}{cc}
    \toprule
     & vsOED-G-T\\
    \midrule 
     $n_{\text{update}}$ & $10001$ \\
     $n_{\text{traj}}$ & $1000$ \\
     $n_{\text{batch}}$ & $10000$ \\
     $a_{\phi,0}$ & $10^{-3}$ \\
     $a_{\phi}$ decay & $0.9999$ \\
     \# of $\phi$ updates per $l$-iteration & $5$ \\
     $n_{\text{mixture}}$ & $8$ \\
     $a_{w,0}$ & $5 \times 10^{-4}$ \\
     $a_{w}$ decay & $0.9999$ \\
     Initial critic lr & $10^{-3}$ \\
     Critic lr decay & $0.9999$ \\
     \# of $\nu$ updates per $l$-iteration & $5$ \\
     Max buffer size & $10^{6}$ \\
     $\gamma$ & $1$ \\
     Initial $\sigma_{k,\text{explore}}$ & $0.05$ \\
     $\sigma_{k,\text{explore}}$ decay & $0.9999$ \\
     Target network lr & $0.1$ \\
    \bottomrule
    \end{tabular}
    \label{tab:hyper_conv_diff}
\end{table}

\begin{figure}[htbp]
  \centering
  \includegraphics[width=0.6\linewidth]{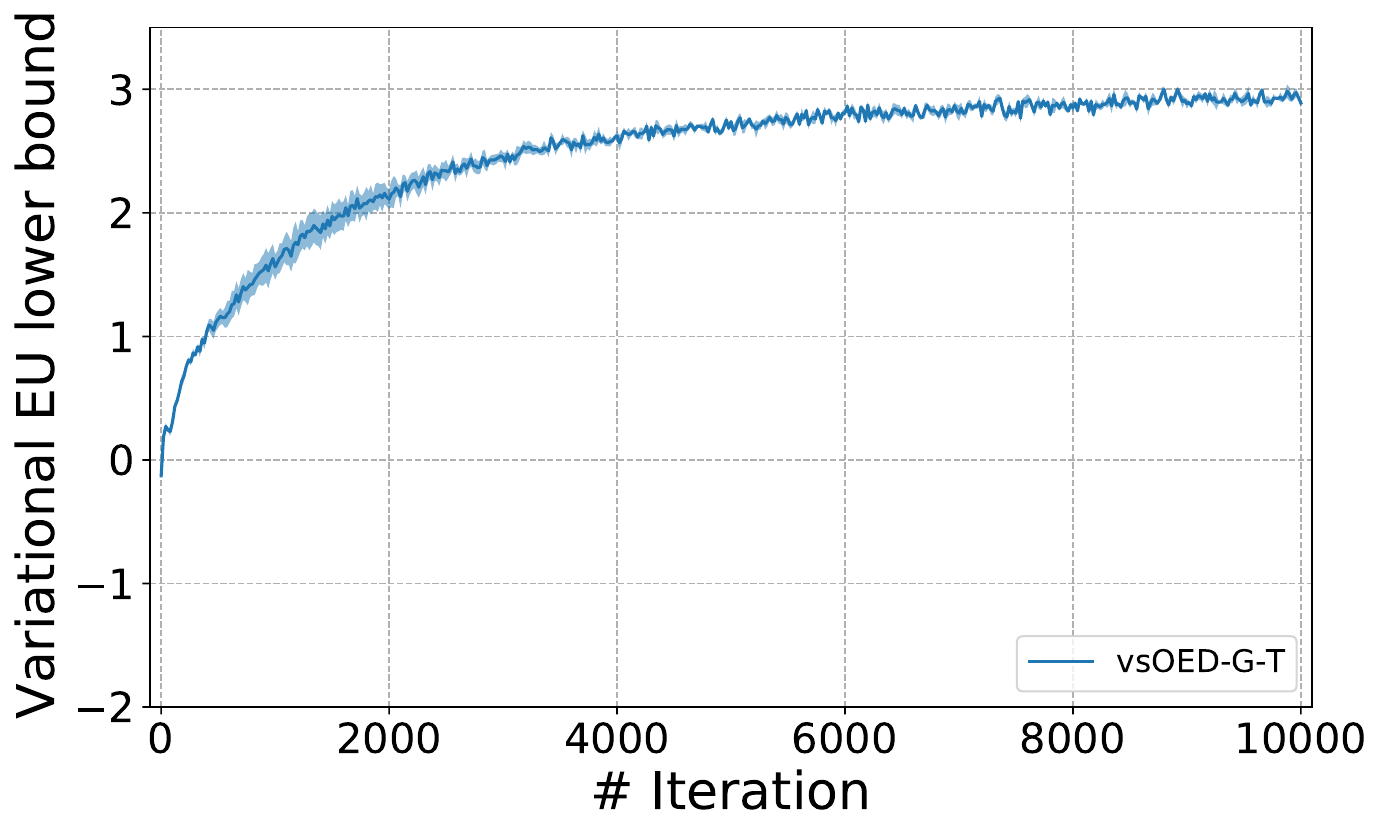}
  \caption{Case 4. Training history of average $\tilde{U}$ over four training replicates for $N=10$. The shaded regions represent the standard error.
  }
  \label{fig:conv_diff_po\Info_EU_vs_update}
\end{figure}

\paragraph{Training stability} 
We only illustrate training stability results for `OED for PoIs' for brevity.
\Cref{fig:conv_diff_po\Info_EU_vs_update} presents the training history of average $\tilde{U}$ (but with the prior term omitted due to the presence of nuisance parameters, per \cref{app:omit_prior}) over four training replicates, with the shaded regions representing standard error. 
\Cref{tab:conv_diff_po\Info_replicates} presents the prior-omitted $\tilde{U}$ and standard error ($\pm$) for four training replicates for $N=10$.
The last column displays the average of the four replicates along with its standard error. 
These results indicate vsOED to have excellent training robustness.

\begin{table}[htbp]
    \centering
    \caption{Case 4. Prior-omitted $\tilde{U}$ and standard error ($\pm$) for four training replicates and $N=10$.}
    \begin{tabular}{ccccc|c}
    \toprule
     & Replicate 1 & Replicate 2 & Replicate 3 & Replicate 4 & Average \\
    \midrule 
    vsOED-G-T & $2.998 \pm 0.002$ & $3.057 \pm 0.002$ & $2.857 \pm 0.002$ & $3.039 \pm 0.002$ & $2.99 \pm 0.04$ \\
    \bottomrule
    \end{tabular}
    \label{tab:conv_diff_po\Info_replicates}
\end{table}

\paragraph{Neural network architectures for surrogate models} 
The architecture for the NN-based surrogate models of $G$ and $\varphi$ 
are provided in \cref{tab:arch_conv_diff_forward,tab:arch_conv_diff_goal}.

\begin{table}[htbp]
    \centering
    \caption{Case 4. Architecture for the surrogate model of $G$. }
    \begin{tabular}{cccc}
    \toprule
    Layer & Description & Dimension & Activation \\
    \midrule 
    Input & $[\param_m, \nuis_m, x, y]$ & $2m+4$  & - \\
    Hidden 1 & Dense & 256 & ReLU \\
    Hidden 2 & Dense & 256 & ReLU \\
    Hidden 3 & Dense & 256 & ReLU \\
    Hidden 4 & Dense & 256 & ReLU \\
    Output & Dense & 1 & - \\
    \bottomrule
    \end{tabular}
    \label{tab:arch_conv_diff_forward}
\end{table}

\begin{table}[htbp]
    \centering
    \caption{Case 4. Architecture for the surrogate model of $\varphi$. }
    \begin{tabular}{cccc}
    \toprule
    Layer & Description & Dimension & Activation \\
    \midrule 
    Input & $[\param_m, \nuis_m]$ & $2m+2$& - \\
    Hidden 1 & Dense & 256 & ReLU \\
    Hidden 2 & Dense & 256 & ReLU \\
    Hidden 3 & Dense & 256 & ReLU \\
    Output & Dense & 1 & - \\
    \bottomrule
    \end{tabular}
    \label{tab:arch_conv_diff_goal}
\end{table}

\end{document}